# A Comprehensive Review on Computer Vision Analysis of Aerial Data

[1]Vivek Tetarwal, [2]Sandeep Kumar*
[1, 2] Central Research Laboratory, Bharat Electronics Limited-Ghaziabad, India
*Corresponding author mailid: sann.kaushik@gmail.com

**Abstract:** With the emergence of new technologies in the field of airborne platforms and imaging sensors, aerial data analysis is becoming very popular, capitalizing on its advantages over land data. This paper presents a comprehensive review of the computer vision tasks within the domain of aerial data analysis. While addressing fundamental aspects such as object detection and tracking, the primary focus is on pivotal tasks like change detection, object segmentation, and scene-level analysis. The paper provides the comparison of various hyper parameters employed across diverse architectures and tasks. A substantial section is dedicated to an in-depth discussion on libraries, their categorization, and their relevance to different domain expertise. The paper encompasses aerial datasets, the architectural nuances adopted, and the evaluation metrics associated with all the tasks in aerial data analysis. Applications of computer vision tasks in aerial data across different domains are explored, with case studies providing further insights. The paper thoroughly examines the challenges inherent in aerial data analysis, offering practical solutions. Additionally, unresolved issues of significance are identified, paving the way for future research directions in the field of aerial data analysis.

## 1. Introduction

In recent years, the rapid advancement of UAVs technology has generated an unprecedented volume of aerial data. This surge in data availability, coupled with the growing need for efficient and accurate analysis, has propelled computer vision techniques to the forefront of aerial data research [1]. The application of computer vision in analyzing aerial data has far-reaching implications across various domains, including agriculture, environmental monitoring, disaster management, urban planning, and defense etc. [2], [3]. The integration of computer vision methodologies with aerial data has unlocked new possibilities for extracting valuable insights from complex and expansive datasets.

Traditional manual methods of analyzing aerial data are not only time-consuming but also prone to errors, limiting their reliability. Moreover, the scalability of traditional manual methods is also limited [4]. As the volume of aerial data continues to grow exponentially, it becomes impractical to keep pace with the sheer quantity of information. This limitation hampers the timely extraction of meaningful insights from the data, especially in applications where rapid decision-making is crucial, such as disaster response and surveillance etc.

In contrast to that, computer vision offers a paradigm shift in the analysis of aerial data. By leveraging algorithms and computational models, it performs the fundamental steps of aerial data analysis like object detection, classification, segmentation, and tracking more efficiently and accurately [4]. Computer vision algorithms are designed to automatically identify and locate objects or patterns within images. This not only accelerates the analysis process but also reduces the dependence on human intervention. In addition to that, it can efficiently process vast amounts of aerial data in a relatively short time frame, making it feasible to analyze large datasets that would be impractical in case of traditional methods [5]. In nutshell, the transition from traditional manual methods to computer vision in the analysis of aerial data represents a significant advancement. It not only addresses the limitations of manual processes in terms of time and error but also introduces automation, scalability, and precision, thereby revolutionizing the way we extract information from the ever-expanding troves of aerial imagery.

Here, we aim to foster a deeper understanding of the synergies between computer vision and aerial data analysis. We provided the fundamentals of computer vision, along with its associated tasks and their categorization. The significance of aerial datasets and challenges associated with it is highlighted. Further various surveys related to analysis of aerial data are discussed and the corresponding research gaps in the studies are identified. A detailed exploration of the distinctions between our work and existing surveys is followed by an examination of the motivation and contributions of our study, emphasizing the timely necessity of this research.

### 1.1. Computer vision

Computer vision is an integral part of artificial intelligence that gives computers the ability to comprehend visual information, akin to human capabilities. Its analogy to human vision is shown in Figure 1. It enables a machine to perceive its 3D surroundings by means of 2D representations viz. images and videos [4], [6].

Computer vision empowers computers or more generically machines to see and interpret visual information by means of various sophisticated algorithms and models. Its primary goal is to equip machines with the ability to recognize objects, understand scenes, and make decisions based on visual input [7]. The algorithms and models for computer vision are developed to gain high-level understanding from digital images or videos. It goes beyond the basic image processing tasks to understand the content of images. Image processing focuses on altering the visual representation of images, computer vision is more concerned with the interpretation and understanding of those images [5].

Within the domain of artificial intelligence, the integration of computer vision and image processing has propelled machines to interpret and comprehend the visual world, mirroring the nuanced capabilities inherent in human vision.

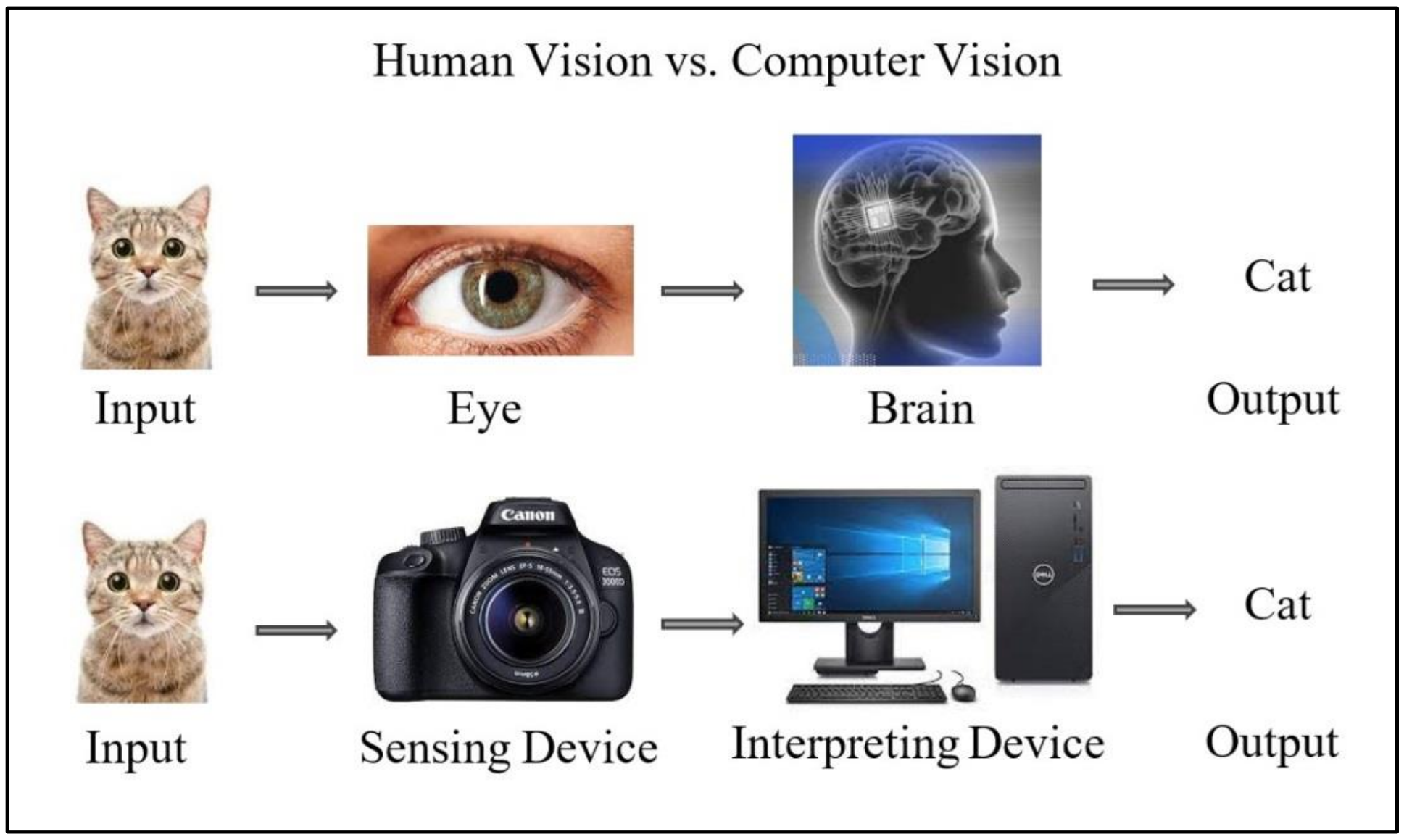

Figure 1: Human vision vs. computer vision

The roots of computer vision and image processing trace back to the 1950s when Frank Rosenblatt developed the perceptron, laying the foundation for early image recognition [8]. The subsequent decades witnessed the evolution of theories and models, with David Marr's work in the 1970s contributing valuable insights into visual processing. Image processing techniques, such as edge detection, gained prominence during this period [9], [10].

The 1980s saw a synergy between computer vision and image processing, with researchers exploring ways to integrate these fields [11]. The active vision approach proposed in [12] emphasized the control of a camera's viewpoint to improve the perception quality of the sensed visual data. During the 1990s, neural networks and ML algorithms were integrated, leading to significant contribution in DL-based image classification into the domain of computer vision [13]. In the 2010s, a huge shift in the field of computer vision was witnessed with increase in popularity of DL algorithms [14]. This period witnessed the convergence of computer vision and image processing in real-world applications marking a pivotal moment in the journey of visual understanding. Based on the area of interest in the image/video, these tasks can be categorized in two major classes namely objects level tasks and scene level tasks, which is illustrated in Figure 2.

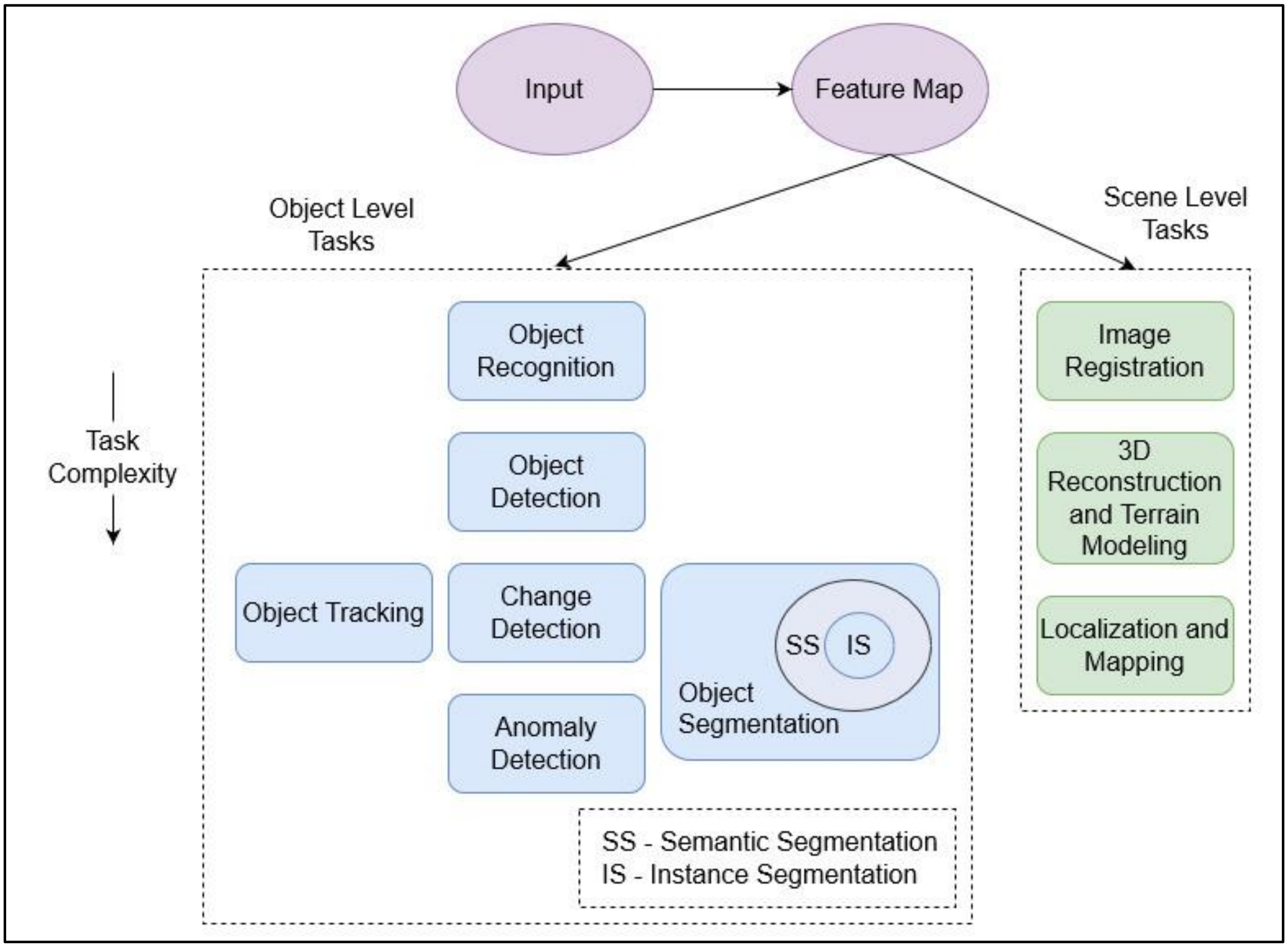

Figure 2: Complexity based organisation of computer vision tasks

### *1.1.1. Object level tasks*

Object level tasks focus on objects in a visual scenario, that involves the analysis and understanding of various entities or instances which belong to one or more classes within a given context. Such tasks encompass object recognition, object detection, object tracking, change detection, anomaly detection, and object segmentation etc. Object recognition is the process of identifying and categorizing objects into predetermined classes or categories [15]. When dealing with a single object in an image or video, a singular label is assigned, leading to the term image classification. However, in scenarios involving two or more objects, each entity is individually identified and categorized, constituting the broader process known as object recognition [16]. In object detection, the procedure of recognizing and precisely locating an object within an image or video is performed by creating a bounding box around it. The primary goal of object detection is to furnish details about both the identity i.e., what and spatial location i.e., where of an object in the given visual content [17]. The process of tracking the movement of objects, such as vehicles or persons, across multiple frames of aerial video or image sequences is referred to as object tracking [18]. It involves systematic analysis of consecutive frames to trace the trajectory of specific objects over time to maintain their identity and location, as they move through the scene, facilitating a coherent understanding of their dynamics. The process of identifying alterations in imagery over different time instances is known as change detection [19]. This technique involves comparing images of a scene at distinct time periods and highlighting the areas of change. Change detection provides temporal insights into landscape evolution and supports decision-making in urban planning and environmental management. It is instrumental in monitoring urban development, tracking environmental changes, video surveillance and analysing the impact of disasters. The process of systematic identification of abnormal patterns or objects within visual data, contrasting them with established norms is referred to as anomaly detection [20]. This sophisticated process is essential in surveillance contexts, as it plays a pivotal role in pinpointing potential threats or irregularities that may go unnoticed by human observers. Object segmentation is the process of delineating object boundaries at pixel level in visual data [21]. This process is further categorized into two types namely semantic segmentation and instance segmentation. In semantic segmentation, we assign semantic labels or class labels to each pixel in an image that can be used for mapping land cover types and analyzing diverse terrains by categorizing pixels into specific land cover types etc. [22]. In instance segmentation, the semantic labels or class labels are assigned to each pixel in an image with distinction between individual instances of the same class. It plays a key role in scenarios where precise counting and tracking of individual objects are essential, such as in surveillance or industrial applications [23]. In each of these tasks, a common subtask involves the identification of one or more objects, followed by processes such as localization, movement tracking, comparison with historical instances for change detection, extrapolation of future states for anomaly detection, or delineation of object boundaries.

### *1.1.2. Scene level tasks*

Scene level tasks focus on a particular scene or the entire image/frame scenario in the visual data, involving the comprehensive analysis of context, composition, or environmental characteristics. Tasks such as image registration, 3D reconstruction and terrain modeling, as well as localization and mapping are some scene level tasks. Image registration is a critical process in computer vision that involves aligning images of a scene captured from different sources and time points. This alignment is achieved by establishing the correspondences between points in two images, typically a reference image and a sensed image. The reference image serves as a baseline, and for each point in this image, the corresponding point in the sensed image is identified [24]. This spatial alignment is essential for tasks such as comparing images, integrating data from diverse sources, creating mosaics, and analyzing single or multi-source videos. In 3D reconstruction and terrain modeling, using DEMs and SfM techniques, a detailed representation of the Earth's surface and objects is created [25]. 3D reconstructions can be enhanced by using precise elevation data, contributed by LiDAR technology. The process of localization & mapping is used for creating maps of an area, while simultaneously tracking the device's location. In aerial data analysis, it enables drones or UAVs to navigate and map their surroundings in real-time with the help of EO sensors [26]. In the scene-level tasks, the emphasis is on capturing the entire scene, encompassing activities like aligning images, creating 3D terrain models, mapping and locating objects within an environment.

In any computer vision task, the first step involves the processing of input data and analyzing it to extract the meaningful features. These features constitute a pivotal element in the execution of these tasks. Subsequently, these extracted features are employed in training the models which are tailored for various tasks. In terms of relative complexity, the organization of these tasks is shown in Figure 2, with complexity increasing from top to bottom [27].

Typically, these tasks are carried out sequentially in accordance with their increasing complexities. For example, in the case of object level tasks, object detection precedes object tracking [28], and object detection comes before change detection [29]. However, the order of task execution based on complexity is not consistent. In [30], change detection is performed before object detection; in [31], [32], object segmentation precedes object recognition; and in [33], [34] object segmentation is performed before object detection. Additionally, in certain instances, both object level and scene level tasks may be required based on the application scenarios [35]. This flexible approach accommodates diverse requirements, ensuring optimal outcomes across a range of applications.

### 1.2. Aerial Datasets

An aerial data in the context of computer vision and remote sensing refers to a collection of images or video data that has been captured from aerial platforms, such as satellites, drones or UAVs, or manned aircrafts. Satellite-based aerial datasets consist of images captured by orbiting satellites that can cover large geographical areas [36]. They are often used for global-scale applications like climate monitoring, land cover classification, and disaster management. UAVs, commonly known as drones, have opened up new frontiers in various fields and have become a valuable tool for capturing data in a wide range of emerging applications. The datasets captured by UAVs are typically used for a variety of purposes, including land cover classification, object detection, scene understanding, environmental monitoring, urban planning, and much more [37]. UAVs or drones are increasingly equipped with video cameras to capture aerial video datasets which are further used for tasks like tracking moving objects, surveillance, and disaster response. The videos also provide a more extensive view of an area and are used in applications like mapping and large-scale monitoring [38]. Aerial data captured by UAVs can be classified into different categories such as optical photography and videography, thermal imagery and hyperspectral imagery [39], [40], [41].

Analyzing the aerial data provides a wealth of applications across diverse domains, but introduces unique challenges, which are not encountered in ground-level visual data analysis [42]. These challenges are rooted in several factors intrinsic to the aerial perspective. Some common challenges associated with aerial data analysis include variability in scale and resolution, unpredictable orientations, presence of cluttered backgrounds and unstructured patterns etc., that emphasize the need of specialized handling of the data for its efficient interpretation and utilization. Successfully navigating through these challenges is crucial for unlocking the full potential of aerial data and harnessing the wealth of information it offers across numerous fields. These challenges are explained in detail in Section 5.

### 1.3. Related Surveys and Research Gaps

A lot of surveys related to the analysis of aerial data have been carried out in the past. In [42], an overview of the current landscape of aerial data has been provided. The survey encompasses a broad spectrum of works dedicated to tasks like detection, tracking, identification, and action recognition etc. The exploration extends to diverse aerial datasets and their associated architectures, with a specific focus on surveillance applications. However, the study falls short in terms of providing details on the tasks of change detection, object segmentation and scene level tasks. Also, the details of evaluation metrics, hyper parameters, and computational costs have not been explored in [42]. In [43], the authors have studied the works related to object detection task only. Although the authors have discussed various aerial datasets and the techniques for object classification however the thorough details about the architectures used in aerial data analysis has not been provided. Along with that the evaluation metrics, computation and hyper parameters details have not been provided in [43]. Likewise, in [36] and [39], the attention is given not only to object detection but also to object tracking works. The papers furnish information on diverse datasets and the architectures employed. Although the details regarding computational costs associated with various methods has been provided in [39], the study lacks specification regarding the applications of their findings and the specific hyper parameters used. The authors of [36] have provided the evaluation metrics details and applications of the aerial data analysis in vehicle surveillance but the details of computation cost and hyper parameters have not been provided. An overview of these studies in terms of important parameters is shown in Table 1 which is summarized as follows.

While some surveys, like those documented in [36], [42] and [44], have centred around surveillance applications, others, including [39], [43], [45], [46], [47] lack application specificity. Importantly, with the exception of [39] in which computation cost is considered based on the experimental environment and backbone model, contingent on the computing power of NVIDIA's GPU, these surveys have not extensively covered essential parameters such as computation cost, model hyper parameters, and libraries details, which are integral components of a model. These parameters play a significant role in the creation of a model and need to be focused upon to provide an insight of the configurations that have proved to be optimum in the past and may also direct the future studies in aerial data analysis. Based on these facts, we have found out the following gaps in the already existing surveys related to the analysis of the aerial data.

- The existing surveys in the field of aerial data analysis have predominantly focused on the specific tasks of object detection and tracking. Change detection, object segmentation and scene level tasks are not being well explored in the existing surveys.
- Parameters such as model hyper parameters, and libraries used, which are integral component of all computer vision methods used in aerial data analysis have not been explored in the related surveys.
- The existing surveys have not provided the comprehensive review of the aerial datasets, architectures used and evaluation metrics of computer vision in aerial data analysis.
- Computer vision task-oriented applications in aerial data used in different domains have not been explored till now. Further, the case study of different applications is also not present in the related surveys.

Table 1: Analysis of past surveys based on different parameters

| S. NO. | References | Survey Parameters | | | | | | | |
|---|---|---|---|---|---|---|---|---|---|
| | | **Main Task discussed** | **Aerial Data Type** | **Dataset** | **Architecture & Libraries** | **Evaluation Metrics** | **Applications Case study** | **Hyper parameters** | **Computation Cost** |
| 1. | [42] | Object detection & tracking, identification, action recognition | NC | C | Architectures- C, Libraries-NC | NC | NC | NC | NC |
| 2. | [43] | Object detection | NC | C | Architectures- C, Libraries-NC | C | NC | NC | NC |
| 3. | [39] | Object detection & Tracking | Image and Video | C | Architectures- C, Libraries-NC | C | NC | NC | C |
| 4. | [36] | Object detection and tracking | Image and Video | C | Architectures- C, Libraries-NC | C | NC | NC | NC |
| 5. | [45] | Object detection | Images | C | Architectures- C, Libraries-NC | NC | NC | NC | NC |
| 6. | [44] | Object tracking | Videos | C | Architectures- C, Libraries-NC | C | NC | NC | NC |
| 7. | [18] | Object detection | Images | C | Architectures- C, Libraries-NC | C | NC | NC | NC |
| 8. | [46] | Object detection | Images | C | Architectures– C Libraries-NC | C | NC | NC | NC |
| 9. | [47] | Object tracking | Image and Videos | C | Architectures– C Libraries-NC | C | NC | NC | NC |
| 10. | [48] | Change detection | Images | C | Architectures- C, Libraries-NC | C | NC | NC | NC |
| 11. | [49] | Object segmentation | NC | NC | Architectures- C, Libraries-NC | C | NC | NC | NC |
| 12. | [50] | Image localization | NC | NC | Architectures- C, Libraries-NC | NC | NC | NC | NC |

Considered-C, Not considered-NC

### 1.4. Motivation and Contribution

In an attempt to fill these gaps, the current study is motivated by the need to provide a more comprehensive understanding of advancements in aerial data analysis. We aim to extend the scope beyond the fundamental computer vision tasks like object detection and tracking, and giving thorough reviews of the works on change detection, object segmentation and scene level tasks related to aerial data. We aim to extend the scope and offer insights into the key parameters like computation cost, model hyper parameters, and libraries used. Applications and related case studies that have proven optimal in the past, guiding future studies in the broader field of computer vision tasks within aerial data analysis. This comprehensive approach aims to drive advancements and innovations beyond the current focus on object detection and tracking. Taking into considerations all these factors, the main contributions of the surveys are:

- We have provided comprehensive reviews of all computer vision tasks in aerial data analysis. Along with the fundamental tasks of object detection and tracking, we have predominantly focused on change detection, object segmentation and scene level tasks in this work.
- The detail and comparison of various hyper parameters used in different architectures and tasks have been provided here. The detailed discussion on the various libraries, their classification and their usefulness to different domain expertise has been provided.
- The comprehensive review of the aerial dataset, the architectures used and the evaluation metrics of all computer vision tasks in aerial data analysis have been made here.
- Computer vision task-oriented applications in aerial data used in different domains have been discussed. Further, the case studies related to different applications are also discussed in this survey.
- The challenges associated with aerial data analysis are thoroughly reviewed, followed by their practical solutions. Some of the important unresolved problems are discussed and the directions for the future research avenues in the aerial data analysis are given.

This work aims to provide a comprehensive overview of the state-of-the-art in vision analysis of aerial data. By synthesizing existing research and identifying gaps, this survey seeks to guide researchers, practitioners, and policymakers in navigating the dynamic landscape of computer vision applied to aerial data.

For the convenience of the readers and easy referencing, all the acronyms used throughout the work are defined in Table 2.

Table 2: List of acronyms

| Abbreviation | Definition |
|---|---|
| Adagrad | Adaptive Gradient Algorithm |
| Adam | Adaptive Moment Estimation |
| AIDER | Aerial Image Database for Emergency Response |
| AIWR | Aerial Image Water Resource |
| ANDT | Anomaly Detection with Transformer |
| AP | Average Precision |
| AR | Augmented Reality |
| AUC | Area Under the ROC Curve |
| BAIR | Berkeley AI Research |
| BANDON | Building Change Detection with Off-Nadir Aerial Images |
| BRIEF | Binary Robust Independent Elementary Features |
| BVLC | Berkeley Vision and Learning Center |
| CAFFE | Convolutional Architecture for Fast Feature Embedding |
| CCFM | Class-Oriented Context Fusion Module |
| CDNTS | Corner Detection and Nearest Three-point Selection |
| CL | Convolutional Layers |
| CNN | Convolutional Neural Network |
| COCO | Common Objects in Context |
| CRAM | Class-Oriented Region Attention Module |
| CRF | Conditional Random Field |
| CSP | Cross Stage Partial |
| CUDA | Compute Unified Device Architecture |
| CVGLab | Computer Vision and Geometry Laboratory |
| DE-FPN | Densely Embedded FPN |
| DEM | Digital Elevation Model |
| DIRS | Diversity Richness and Scalability |
| DL | Deep Learning |
| DOTA | Dataset for Object Detection in Aerial Images |
| DRoIN | Dual Regions of Interest Network |
| DSM | Digital Surface Model |
| DTM | Digital Terrain Model |
| EPFL | École Polytechnique Fédérale De Lausanne |
| ERA | Event Recognition in Aerial Videos |
| FAIR | Facebook’s AI Research |
| FAST | Features from Accelerated Segment Test |
| FCL | Fully Connected Layers |
| FCN | Fully Convolutional Network |
| FDLID | Flying Deep Learning Integrated Drone |
| FL | Federated Learning |
| FN | False Negative |
| FP | False Positive |
| FPGA | Field-Programmable Gate Array |
| FPN | Feature Pyramid Network |
| FPR | False Positive Rate |
| FPS | Frames Per Second |
| GAN | Generative Adversarial Networks |
| GIS | Geographic Information Systems |
| GLDN | Global-Local Detection Network |
| GLSAN | Global-Local Self-Adaptive Network |
| HBB | Horizontal Bounding Box |
| HPND | Historical Permission Notice and Disclaimer |
| ILSVRC | ImageNet Large Scale Visual Recognition Challenge |
| IoU | Intersection over Union |
| iSAID | Instance Segmentation in Aerial Images Dataset |
| ISBDA | Instance Segmentation in Building Damage Assessment |
| ISR | Intelligence Surveillance and Reconnaissance |
| JAX | Just after Execution |
| LandCover.ai | Land Cover from Aerial Imagery |
| LBAI | Little Birds in Aerial Imagery |
| LiDAR | Light Detection and Ranging |
| LRN | Local Response Normalization |
| LSRN | Local Super-Resolution Network |
| LSTM | Long–Short-Term Memory |
| mAP | Mean AP |
| MDA-Net | Multi-Dimensional Attention Network |
| MDL-RS | Multimodal DL-Remote Sensing |
| MILA | Montreal Institute for Learning Algorithms |
| mIoU | Mean IoU |
| MIT | Massachusetts Institute of Technology |
| ML | Machine Learning |
| MOR | Moving Object Recognition |
| MOT | Multiple Object Tracking |
| mPA | Mean PA |
| MPL2 | Mozilla Public License 2.0 |
| MS CNTK | Microsoft Cognitive Toolkit |
| MSM COCO | Monotonous Scale Match on COCO |
| MTGCD-Net | Multi-Task Guided Change Detection Network |
| MVG | Multiple View Geometry |
| NDFT | Nuisance Disentangled Feature Transform |
| NIR | Near Infra-Red |
| NMS | Non-Maximum Suppression |
| NWPU VHR-10 | Northwestern Polytechnical University Very High-Resolution-10 |
| OAN | Objectness Activation Network |
| OBB | Oriented Bounding Box |
| OBIA | Object-Based Image Analysis |
| OBIA-MDTWS | OBIA- Multilevel Distance Transform Watershed Segmentation |
| OCSVM | One Class Support Vector Machine |
| OpenCV | Open-Source Computer Vision |
| PA | Pixel Accuracy |
| PIL | Python Imaging Library |
| POG | People on Grass |
| RCNN | Region-based CNN |
| ReLU | Rectified Linear Units |
| ResNet | Residual Network |
| RFEB | Receptive Field Expansion Block |
| RMSProp | Root Mean Square Propagation |
| ROC | Receiver Operating Characteristic |
| ROI | Region of Interest |
| RPN | Region Proposal Network |
| RW-UAV | Rotary-Wing UAV |
| S2FL | Shared and Specific Feature Learning |
| SAMFR | Spatial Attention for Multi-Scale Feature Refinement |
| SAR | Synthetic Aperture Radar |
| SARD | Search and Rescue Image Dataset for Person Detection |
| SARSA | Self-Adaptive Region Selecting Algorithm |
| SARUAV | Search and Rescue with UAV |
| SCRDet | Small Cluttered Rotated Objects Detector |
| SENet | Squeeze-and-Excitation Network |
| SfM | Structure-from-Motion |
| SF-Net | Sampling Fusion Network |
| SGD | Stochastic Gradient Descent |
| SHDL | ScatterNet Hybrid Deep Learning |
| SIFT | Scale-Invariant Feature Transform |
| SLAM | Simultaneous Localization and Mapping |
| SM COCO | Scale Match on COCO |
| SNIPER | Scale Normalization for Image Pyramids with Efficient Resampling |
| SOT | Single Object Tracking |
| SRM | Spatial-Refinement Module |
| SSD | Single Shot MultiBox Detector |
| SSH | Single Stage Headless |
| SURF | Speeded-up Robust Features |
| SVI | Street View Images |
| TDA | Training Data Augmentation |
| TN | True Negative |
| TP | True Positive |
| TPR | True Positive Rate |
| UAV | Unmanned Aerial Vehicle |
| UAVDT | UAV Detection and Tracking |
| UMCD | UAV Mosaicking and Change Detection |
| VALID | Virtual Aerial Image Dataset |
| VGGNet | Visual Geometry Group Network |
| ViT | Vision Transformer |
| VR | Virtual Reality |
| WAID | Wildlife Aerial Images from Drone |
| YOLO | You Only Look Once |

The structure of the remaining document is outlined as follows. Section 2 provides the detailed analysis of various computer vision tasks, offering a review of the latest research in these areas. In Section 3, we delve into the principal architectures, datasets, evaluation metrics, libraries and hyper parameters commonly employed for these tasks. The diverse applications of aerial data analysis and the related case studies are detailed in Section 4. Section 5 outlines the challenges and their mitigations in the field. Section 6 pinpoints the open problems and future research avenues in the field and Section 7 concludes the paper. Figure 3 illustrates the paper's detailed organizational structure.

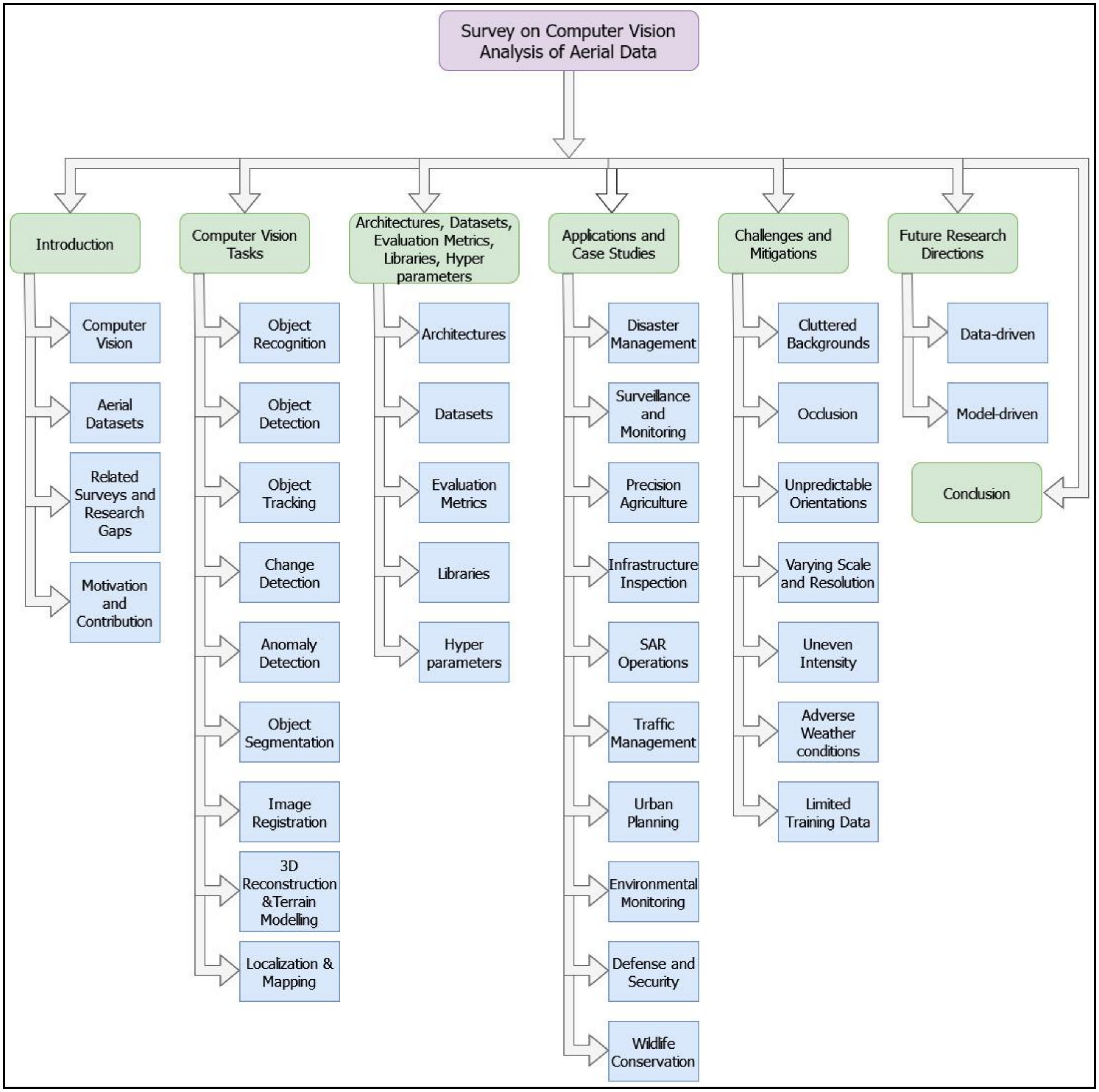

Figure 3: Detailed organisation of the paper

## 2. Computer vision tasks in the analysis of aerial data

As discussed in the previous section, the field of computer vision is characterized by a rich spectrum of tasks, which are broadly classified into two main categories: object level tasks and scene level tasks [51]. The primary objective in object level tasks is to understand and analyze individual objects within visual data, offering a granular perspective. In contrast, scene level tasks are oriented towards interpreting the overall context and structure of visual scenes, providing a holistic understanding beyond individual objects [52].

A comprehensive understanding of a visual environment often demands the simultaneous execution of both object level and scene level tasks. This synergy enables a more holistic interpretation of visual data, as the interplay between individual objects and their broader context is crucial for applications across diverse domains. In the specific realm of aerial data, these tasks have applications across various domains like environmental monitoring, disaster response, agriculture and urban planning [3], [38]. In environmental monitoring, scene level tasks such as terrain modeling assist in understanding landscape changes, while object level tasks like object segmentation may be employed to identify specific flora or fauna for biodiversity studies [53]. In disaster response, the simultaneous execution of object detection and scene level tasks like localization is crucial for assessing the impact of natural disasters and planning effective responses [54], [55]. For agriculture, object recognition can aid in crop monitoring, while 3D reconstruction and mapping contribute to precision agriculture practices. Urban planning benefits from scene level tasks such as localization and mapping for infrastructure analysis, while object level tasks like object detection and segmentation help identify and analyze urban elements like buildings and roads [56]. This multidimensional application of object level and scene level tasks in aerial data analysis highlights their adaptability and significance in extracting valuable insights from large-scale geographic information. The collaborative execution of these tasks is instrumental in addressing diverse challenges and optimizing decision-making processes across various domains. The symbiotic relationship between these tasks is central to advancing the field and addressing the intricacies of real-world visual data analysis.

### 2.1. Object Recognition

Object recognition is the process of identifying and classifying objects within visual data, by assigning predetermined classes or categories to them. Its primary objective is to empower machines to comprehend and interpret visual content, much like humans recognize objects. Assigning a single label to an image with single object is known as image classification, whereas the more extensive task of individually identifying and categorizing multiple objects within an image is termed object recognition [15], [16]. The fundamental nature of object recognition in computer vision lies in its contribution to the development of intelligent systems capable of understanding and interpreting visual information. This capability is pivotal for machines to interact with and interpret their surroundings. Moreover, object recognition serves as a foundational element for more intricate vision tasks, including object detection, segmentation, and tracking [27]. The process involves the utilization of algorithms and models, particularly DL techniques like CNNs [57], [58]. This encompasses training a model on a labeled dataset, allowing it to learn features and patterns associated with various classes. Subsequently, the model produces a probability distribution for potential classes in an input image, selecting the class with the highest probability as the predicted class.

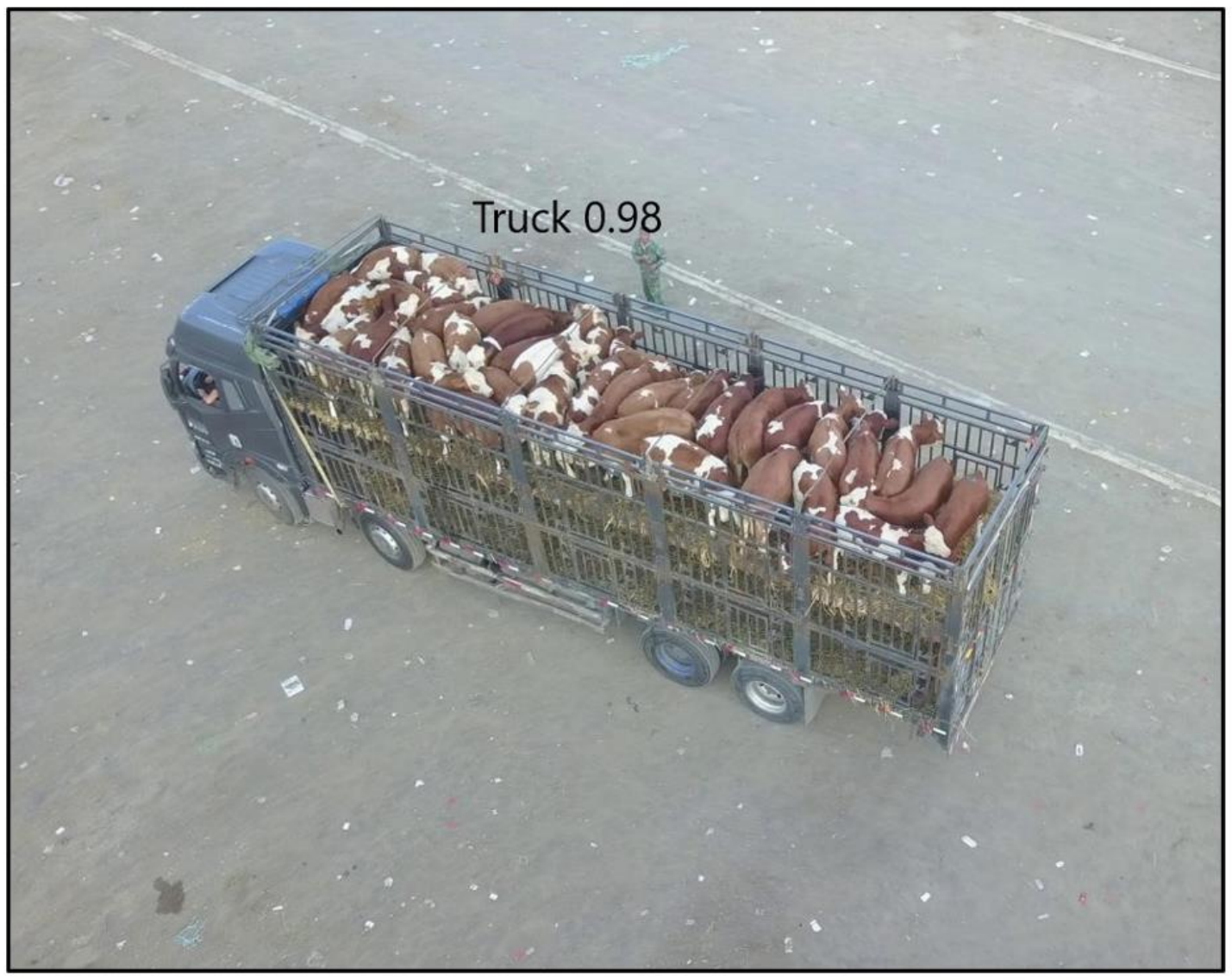

(a)

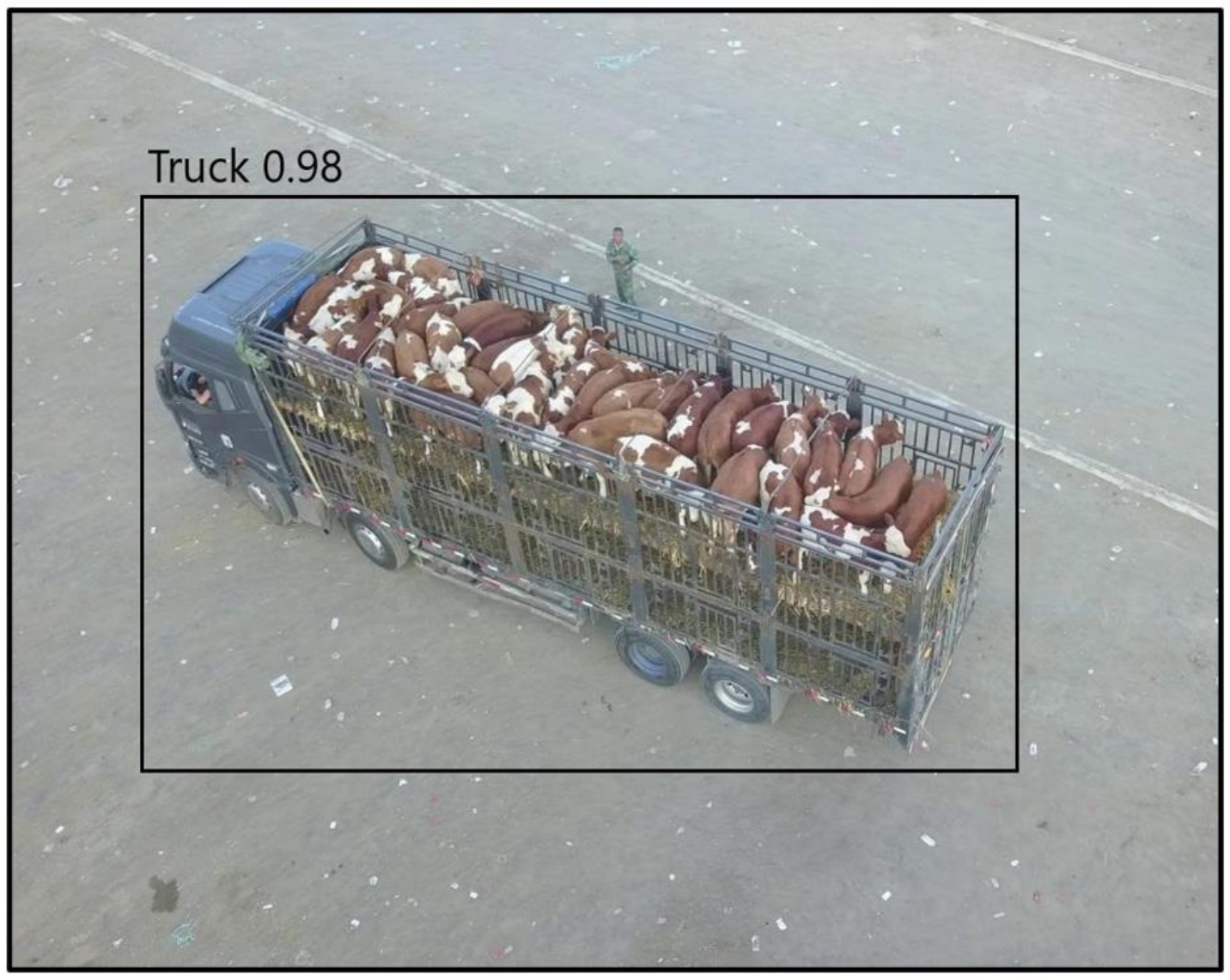

(b)

Figure 4: (a) Identification of object class with confidence score, (b) Object class identification and localization using bounding box [59]

The outcome of an object recognition task can be visually represented, as depicted in Figure 4 (a) where an image of a truck is shown. For a machine to recognize it, the image is fed as input to a trained object recognition model and the model outputs the class probability of this image being a truck as 0.98 i.e., the model is 98% confident that the image is of a truck.

Object recognition extends its applications across diverse domains in the field of aerial data analysis [45]. For urban planning and infrastructure development, it assists in the identification and categorization of elements such as buildings, roads, and other infrastructure components. In disaster response and management, it aids in the identification of damaged structures and hazards, contributing to effective response strategies. Furthermore, in environmental monitoring, it plays a key role in identifying changes in ecosystems, deforestation, and other ecological shifts.

## 2.2. Object Detection

Object detection is the process of identifying and accurately locating an object within an image or video by drawing a bounding box around it [17]. The process involves recognizing the presence of objects, classifying them into predefined categories, and providing accurate bounding box coordinates for their spatial localization. The essence of object detection lies in its fundamental role in empowering computer systems to identify and locate objects within visual data, a crucial capability that mirrors human perception. The process involves the utilization of algorithms and models, particularly DL techniques like CNNs [60], [61]. In this process, an input image or video is fed that is preprocessed using steps such as resizing and normalization. Feature extraction is then performed using CNNs to identify relevant patterns and shapes. Subsequently, region proposal techniques or networks suggest potential areas where objects might be present. These proposed regions undergo ROI pooling to standardize sizes, followed by object classification to determine the probability of each region containing a specific class. Simultaneously, bounding box regression refines the coordinates for accurate localization. Further, NMS is applied to eliminate redundant bounding boxes, and the final output includes detected objects, class labels, and bounding boxes [45]. The outcome of an object detection task can be visually represented, as depicted in Figure 4 (b), that shows the image of a truck. For a machine to detect it, the image is fed as input to a trained object detection model and the model outputs a bounding box around the truck (object) with the class probability of the object being a truck as 0.98 specifying the model is 98% confident that the image is of a truck.

In the specific realm of aerial data, object detection showcase its versatility through applications across various domains like environmental monitoring, wildlife monitoring, agriculture, urban planning, disaster response, and defense and security, contributing to advancements in decision-making and resource optimization [3], [36], [62], [63]. In environmental monitoring, it aids in identifying and locating changes in landscapes and ecosystems, providing valuable insights for ecological research and conservation initiatives [64]. In agriculture, it plays a pivotal role in crop management by identifying crops, assessing their health, and monitoring for potential issues, thereby enabling precision agriculture practices [3]. For urban planning and development, object detection in aerial data assists in the identification and analysis of crucial infrastructure elements, aiding urban planners in making informed decisions about city development and improvements [65]. In disaster response and management, it quickly assesses the impact of natural disasters, facilitating the identification of affected areas and the planning of effective response strategies [66], [67]. Additionally, in defense and security, object detection in aerial data enhances situational awareness by supporting surveillance, border control, and threat detection efforts [36], [63]. The adaptability of object detection in aerial data underscores its significance, extracting valuable insights from large-scale geographic information and contributing to advancements in decision-making processes and resource optimization across various domains.

### 2.3. Object Tracking

The primary objective of object tracking is to achieve a continuous and coherent understanding of the movement of specific objects or multiple objects across consecutive frames in a video or image sequence [44]. In contrast to object detection, which identifies objects in individual frames, object tracking emphasizes the preservation of the identity and location of objects over time. When focusing on a single object, the task is referred to as SOT, while tracking multiple objects simultaneously is denoted as MOT [18]. The visual representation of object tracking outcomes at any particular instance can be depicted as in Figure 5. In Figure 5 (a), a single car is being tracked and its current location is specified by the bounding box shown around the car. As the car moves, the bounding box will also move specifying its new location. Similarly, in Figure 5 (b), multiple objects belonging to different classes, such as car, person etc. are being tracked in a single frame.

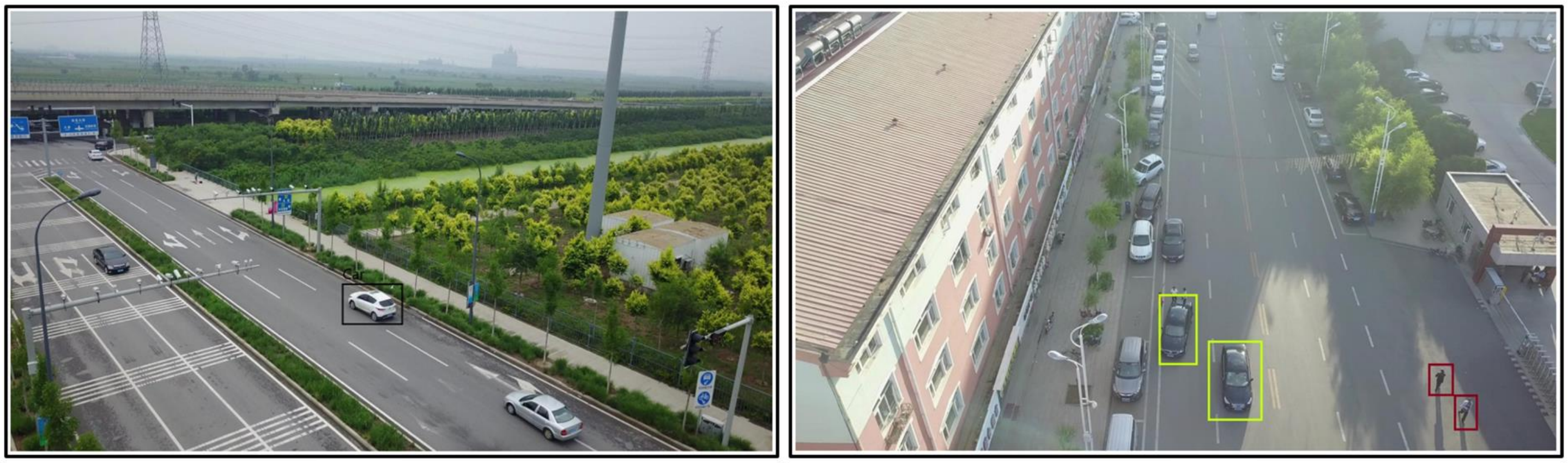

(a) (b)

Figure 5: Tracking of object in aerial data (a) Single object tracking, (b) Multiple object tracking [59]

Tracking can be performed in two modes: online mode and offline mode. In online tracking, frames are processed one by one as they arrive, while in offline tracking the entire video or sequence is available for processing at the beginning [18]. The process provides a coherent understanding of the dynamics of tracked objects by tracing their trajectory over time, maintaining their identity and location as they navigate through the scene. This capability is crucial for comprehending the spatial and temporal evolution of objects within a visual dataset. Object tracking commonly incorporates DL techniques to enhance performance in diverse and dynamic scenarios [68], [69]. The generic process of object tracking begins with the detection of objects in the initial frame of a video or image sequence. After detection, an initial state of the detected objects is established and then using motion prediction algorithms, the object's likely position in the next frame or image is predicted. The predicted position is then linked with the observed position of that object in the current frame through matching and data association, thus facilitating the continuous tracking of the objects across frames. The tracking model is adapted according to the changes in scale, orientation, and appearance, with mechanisms in place to handle occlusions [70], [71]. This iterative process repeats for each subsequent frame, and the tracking algorithm may dynamically adjust parameters based on the evolving characteristics of the scene. The tracking process may end if the object is no longer in the frame or if the tracking algorithm loses confidence. In such cases, the tracker may need to be re-initialized if the object reappears. Post-processing steps, such as noise filtering and trajectory smoothing, may be applied to enhance the accuracy and reliability of the tracking results.

This technique finds significance in traffic monitoring, and surveillance applications [72]. In traffic monitoring, object tracking is crucial for analyzing vehicle movement and behavior, aiding in data collection for traffic flow, congestion patterns, and road safety [73]. Object tracking is a cornerstone in surveillance systems, contributing to monitoring and analyzing movements in diverse environments, from public spaces to critical infrastructure and private settings [44]. Other notable application is in the defense and security domain, where object tracking in aerial data is crucial for surveillance, border control, and threat detection. Tracking the movement of objects or individuals in a given space enhances situational awareness and supports security measures.

2.4. Change Detection

The method involves a meticulous comparison of images acquired at distinct time points, allowing for the precise delineation of areas undergoing change and offering crucial insights into the evolving environmental dynamics. The primary objective of change detection is to reveal temporal variations within a specific scene, providing a temporal perspective that facilitates a nuanced understanding of alterations over time [19]. The essence of this process lies in its capability to track modifications, offering valuable insights into diverse environments. The general process of change detection involves image differencing, where pixel-wise comparisons discern variations [74]. Subsequent techniques, such as thresholding or machine learning algorithms, are then employed to classify and interpret these changes, enabling the detection of subtle alterations in large-scale imagery. This contributes to a comprehensive understanding of evolving environments, serving as a pivotal tool in visual data analysis. In Figure 6 (a) and 6 (b), two images of the same area are shown at different time instances. Figure 6 (c) shows the result of the image differencing at pixel level, highlighting the changes in images over a period of time.

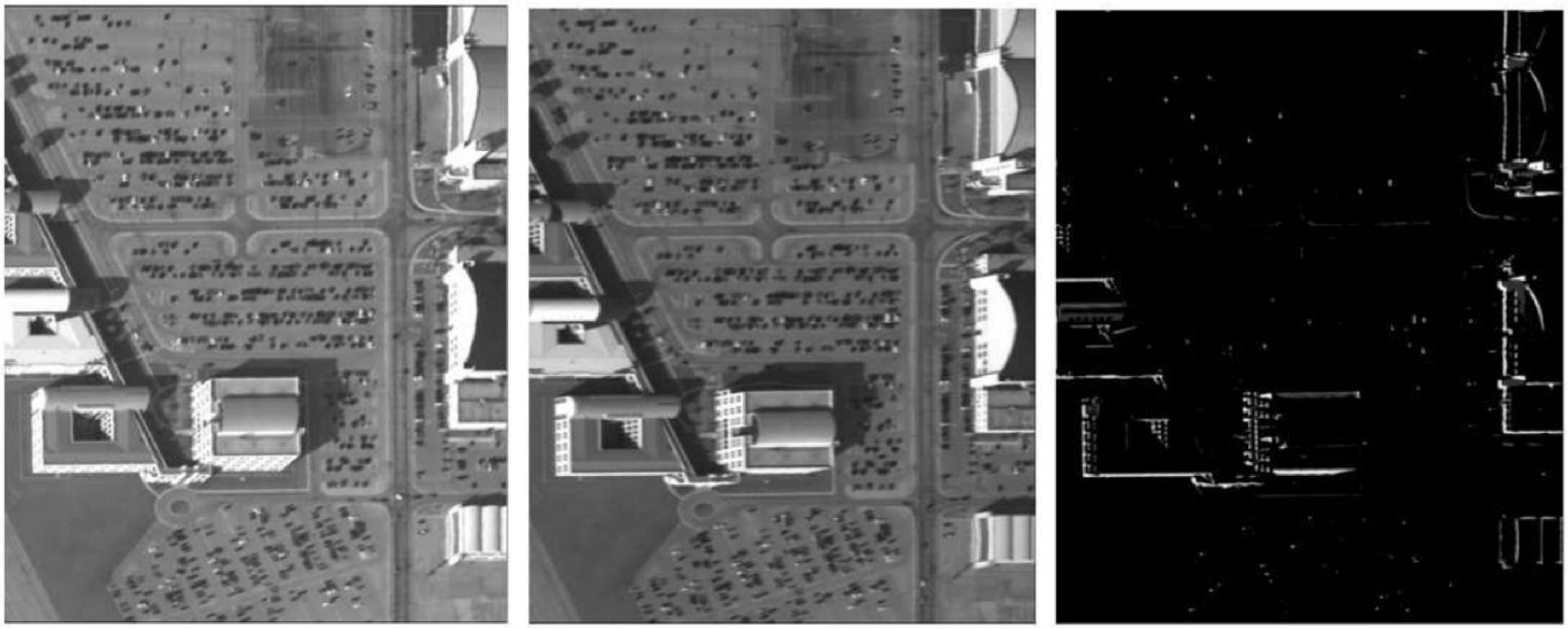

(a) (b) (c)

Figure 6: Change detection in a scene at different time instances, (a) reference image, (b) modified image, (c) the result of change detection [74]

Change detection, with its versatile applications, plays a pivotal role in monitoring urban development by assessing infrastructure growth and aiding urban planners in making informed decisions about land use and development [56], [75], [76]. In the realm of environmental management, it contributes significantly by monitoring vegetation dynamics and assessing the impact of climate change on ecosystems [2]. Moreover, in the domain of video surveillance, change detection serves as a crucial tool for tracking environmental changes over time, enhancing security measures and situational awareness. In disaster response, change detection is instrumental in assessing the aftermath of disasters, facilitating rapid identification of changes in affected areas and supporting effective emergency planning and response strategies. In the specific context of aerial data, change detection proves to be a versatile and indispensable tool with numerous applications [2]. One notable application is in assessing urban sprawl, where it identifies changes in building structures and infrastructure over time, providing valuable insights for urban planning and development. In agriculture, change detection becomes instrumental in monitoring crop health and assessing the effectiveness of irrigation practices, aiding farmers in optimizing their cultivation strategies [3]. Additionally, change detection plays a vital role in disaster response by swiftly identifying changes in affected areas, enabling rapid assessment of the impact and supporting effective emergency planning and response strategies. Past works in change detection have embraced advanced computer vision techniques, such as CNNs, to enhance accuracy and efficiency [77]. These approaches leverage DL to automatically learn and extract features, improving the detection of nuanced changes in visual data. Change detection is a pivotal tool for understanding, managing, and responding to dynamic environments by harnessing the power of visual data analysis.

### 2.5. Anomaly Detection

The primary objective of anomaly detection is to distinguish unusual behavior or events that do not conform to expected patterns and may account to potential threats or abnormalities in different scenarios [78]. The essence of anomaly detection lies in its ability to distinguish unusual behavior or events within visual datasets. Anomaly detection is a pivotal tool with widespread applications. Its ability to identify irregularities within visual data, whether in aerial landscapes or diverse domains, empowers decision-makers with valuable insights for timely intervention and enhanced security [20].

Anomaly detection is achieved through a comprehensive understanding of normal patterns, enabling the identification of deviations that may indicate potential risks or issues [78]. Anomaly detection methods often leverage statistical models, machine learning algorithms, or rule-based systems to establish a baseline of normal behavior and flag instances that significantly deviate from this baseline. By establishing a baseline of expected behavior, anomaly detection methods aim to pinpoint instances that diverge significantly from this norm [79].

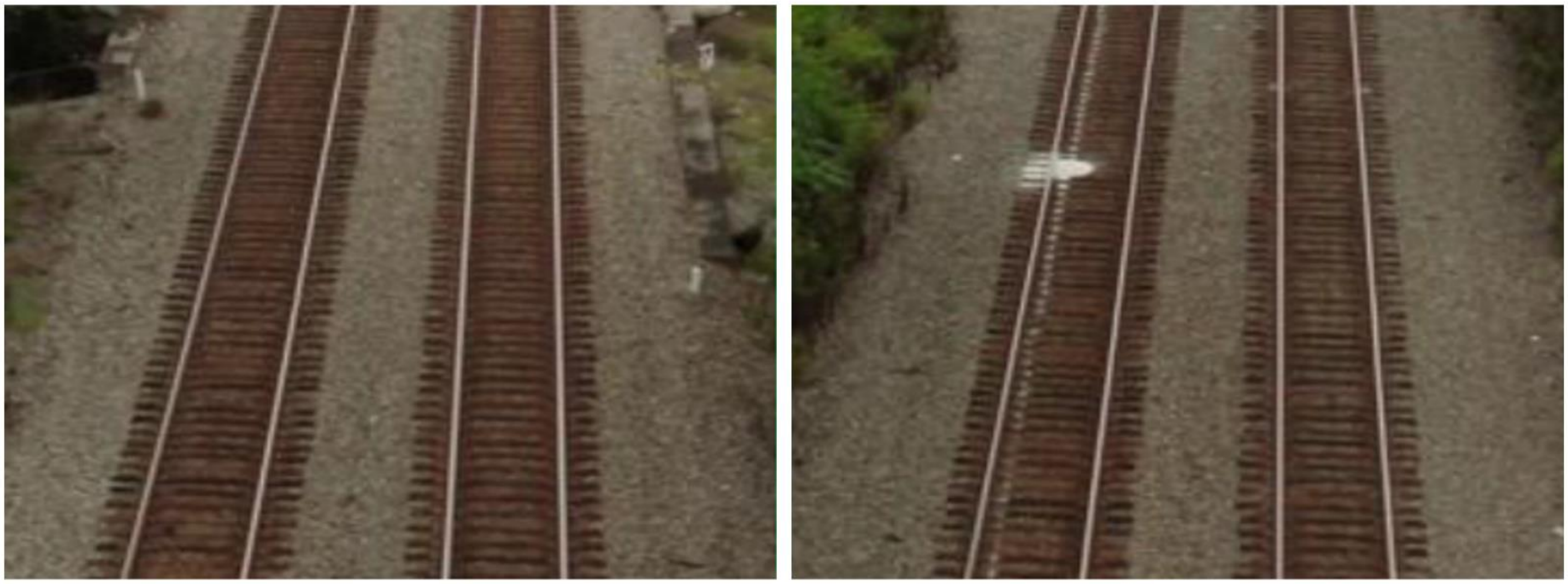

(a) (b)

Figure 7: Identification of irregularities in aerial view of a railway track (a) Normal image (b) Anomalous image [80]

An instance of anomaly detection is depicted in Figure 7. Figure 7 (a) displays a normal frame from a video showing a normal railway track, whereas Figure 7 (b) shows an anomalous frame with some object lying on the track, that may lead to accidents like train derailment.

In the specific context of aerial data, anomaly detection proves to be an invaluable tool for surveillance, monitoring and identifying irregularities or anomalous behavior for identifying irregularities in landscapes [78], [79], [80]. This includes detecting environmental changes, assessing infrastructure damage, or identifying unexpected patterns in terrain. For example, aerial imagery can be analyzed to detect anomalies such as deforestation, land cover changes, or the presence of unauthorized structures. Past works in anomaly detection within aerial data have explored the application of sophisticated algorithms and machine learning techniques [80], [81], [82], [83].

## 2.6. Object Segmentation

Object segmentation is the process of dividing an image into meaningful and distinct regions or segments to identify and delineate individual objects or areas of interest. It involves assigning labels to pixels based on characteristics like color, intensity, or texture, aiming to accurately represent object boundaries at the pixel level [22], [84]. There are two main types of object segmentation, each serving specific purposes, namely semantic segmentation and instance segmentation. Semantic segmentation is the process of partitioning an image into meaningful and coherent segments, with each pixel assigned a semantic label representing a specific class or category [22]. In this process, the objective is to categorize every pixel within the image into predefined classes, such as objects, structures, or background. Instance segmentation involves not only categorizing pixels in an image into semantic classes but also distinguishing between individual instances of objects within the same class [23]. Unlike semantic segmentation, which groups pixels into broad categories, instance segmentation provides a fine-grained understanding by assigning a unique label to each pixel corresponding to a specific object instance. The outcome of object segmentation can be visually represented as depicted in Figure 8. In Figure 8 (a), semantic segmentation is demonstrated, with segments of all instances belonging to one class highlighted similarly; whereas in Figure 8 (b), instance segmentation is demonstrated with each instance segmented separately.

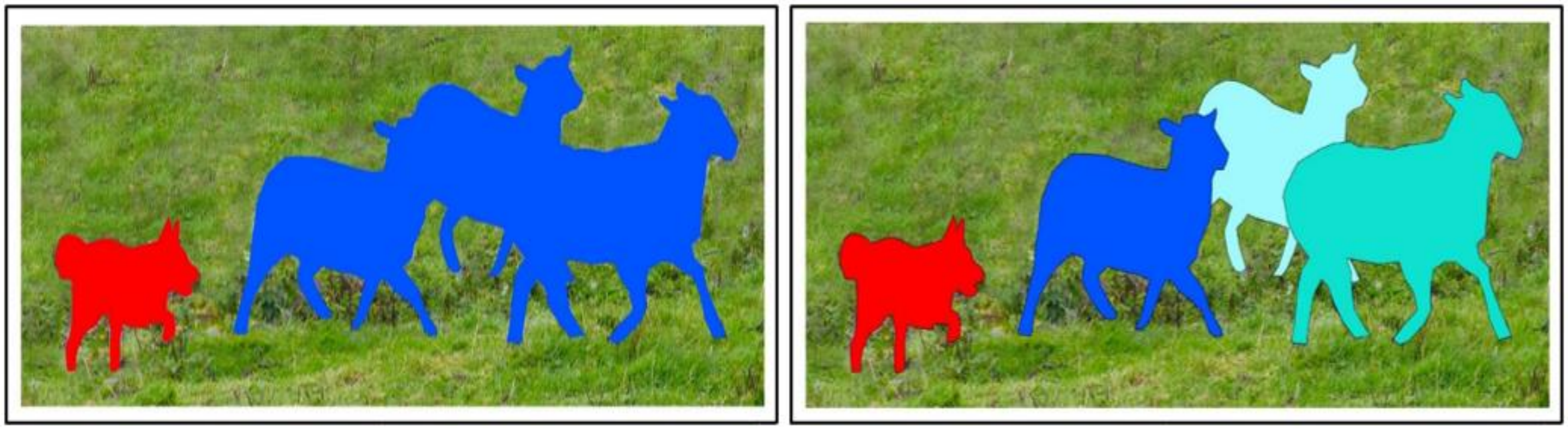

(a) (b)

Figure 8: Pixel level labelling of visual data (a) Semantic Segmentation; (b) Instance Segmentation [85]

It aids in land cover classification, enabling precise mapping and environmental monitoring [86], [87]. Instance segmentation is crucial in scenarios where precise identification, counting, and tracking of distinct objects are essential, such as in surveillance, robotics, and industrial applications. In instance segmentation, the algorithm not only recognizes objects but also separates them into individual entities, allowing for a detailed understanding of the spatial distribution and boundaries of each object instance [88]. In crowded scenes, instance segmentation aids in distinguishing and tracking multiple objects of the same class, contributing to enhanced security and monitoring capabilities [89]. Object segmentation techniques range from basic thresholding and edge detection to more sophisticated methods like clustering, graph-based segmentation, and DL approaches using CNNs. The goal is to accurately separate objects from the background, allowing for precise recognition and analysis [88]. Object segmentation is essential in various applications, including image recognition and AR/VR technologies. Advances in DL have led to highly accurate segmentation models like U-Net [90] and Mask R-CNN [91], contributing to improved performance and efficiency in various tasks. In image recognition, the ability to precisely delineate and isolate objects within an image is crucial for accurate classification and understanding of visual content. Object segmentation allows the systems to not only recognize objects but also to discern their boundaries, enabling a more nuanced interpretation of images [22]. AR heavily relies on object segmentation to seamlessly integrate virtual elements into the real-world environment. Accurate segmentation ensures that virtual objects are precisely placed within the physical space, creating a realistic and immersive AR experience for users. In the realm of aerial data, object segmentation holds particular significance, offering valuable applications and insights for a range of domains. Aerial imagery analysis often involves the identification and isolation of specific objects within large-scale landscapes. This is crucial for tasks such as environmental monitoring, land cover classification, and infrastructure assessment [36].

## 2.7. Image Registration

The primary objective of image registration is to ensure that images from different sources or time points are spatially aligned, allowing for meaningful comparisons and integration. The essence of image registration lies in its ability to overcome disparities in scale, orientation, and position, enabling a coherent synthesis of information from disparate sources [92].

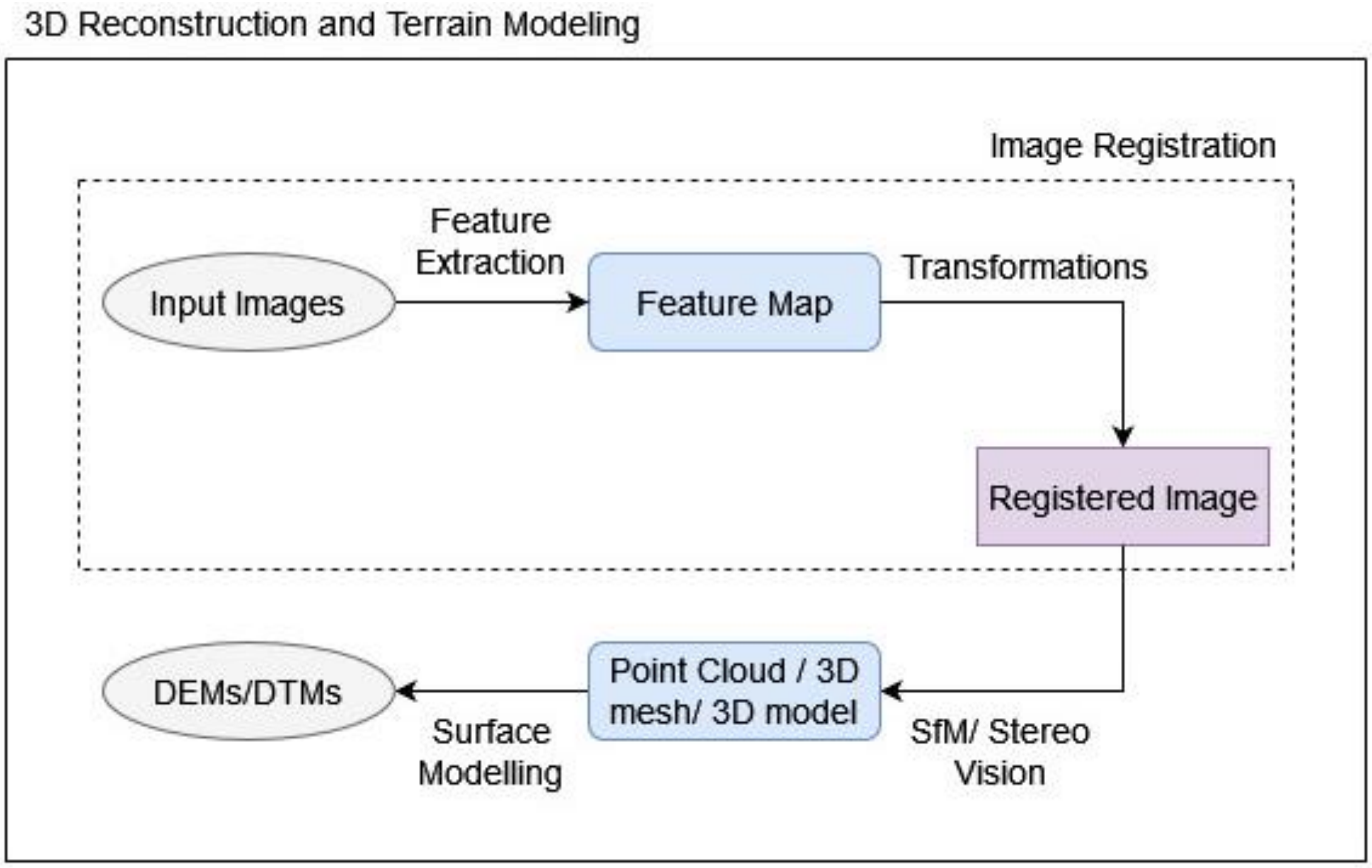

Figure 9: Block diagram of the image registration and terrain modeling process

The process of image registration involves various steps as depicted in Figure 9. It includes establishing correspondences between points in two images, typically a reference image and a sensed image. The reference image serves as a baseline, and for each point in this image, the corresponding point in the sensed image is identified, using several techniques like manual landmark identification, feature-based methods, and intensity-based methods [24]. Feature-based methods involve detecting distinctive features, such as corners or edges, in both images and aligning them, while intensity-based methods use pixel intensities to find similarities. After identifying the similarities, transformations are estimated and applied on the sensed image to align it with the reference image.

The spatial alignment is essential for tasks such as comparing images, integrating data from diverse sources, creating mosaics, and analyzing single or multi-source videos. Image registration plays a pivotal role in applications ranging from remote sensing to surveillance. By facilitating the fusion of information from disparate images, image registration enhances the accuracy and utility of various tasks. In aerial data analysis, image registration finds diverse applications across various domains [93], [94]. In environmental monitoring, registered aerial images enable precise tracking of changes in landscapes over time, such as deforestation, urban expansion, or natural disasters [3], [95]. Urban planning leverages image registration to integrate information from different time points, aiding in monitoring infrastructural developments and land use changes. In precision agriculture, image registration facilitates the alignment of aerial images, allowing for the comparison of crop health over multiple seasons. This aids farmers in identifying areas of concern, optimizing irrigation and fertilization practices, and improving overall crop management [3]. Image registration plays a pivotal role in aligning and integrating information from disparate aerial images, fostering applications in environmental monitoring, urban planning, agriculture, and beyond.

### 2.8. 3D Reconstruction and Terrain Modeling

The objective of 3D reconstruction and terrain modeling is to create accurate and detailed representations of the Earth's surface or objects in three-dimensional space [25]. The process involves extracting spatial information from images to construct a 3D model, providing a comprehensive understanding of the observed environment. The essence of 3D reconstruction and terrain modeling lies in their ability to transform two-dimensional imagery into detailed 3D models, providing valuable insights into terrain features and structures, overcoming the limitations of flat imagery [96].

The process involves training dedicated models to recognize visual features within the images, establishing correspondences between corresponding points in different images i.e., image registration followed by extraction of spatial and depth information from images through SfM or Stereo Vision techniques. This process is shown in Figure 9. After extracting spatial and depth information, point clouds or 3D meshes/models are generated. Surface modeling is applied on the point clouds or 3D models to generate DEMs and DTMs.

3D reconstruction and terrain modeling play pivotal roles in transforming aerial imagery into detailed three-dimensional representations, offering applications across diverse domains and contributing to advancements in environmental analysis, urban planning, and agriculture [97], [98]. These techniques aid in creating detailed 3D models for architectural visualization, infrastructure development, and city management [99]. In environmental monitoring, UAV-based 3D reconstructions enable the analysis of terrain changes, vegetation health, and ecosystem dynamics. In precision agriculture, farmers can use 3D models for crop monitoring, yield estimation, and resource optimization. Specific applications within aerial data include mapping and monitoring construction sites, assessing the impact of natural disasters, and creating digital twins of landscapes for research and planning [3].

### 2.9. Localization and Mapping

Localization and Mapping, more commonly known as SLAM, refers to the process of creating a map of an unknown environment while simultaneously keeping track of the device's location within that environment.

SLAM in aerial data analysis is the process that enables autonomous systems, such as drones or UAVs, to navigate and map their surroundings simultaneously [26]. It involves the integration of sensors, such as cameras, to collect data about the environment as the aerial vehicle moves through it. The primary objective is to generate an accurate and real-time map of the environment while determining the precise location and orientation of the aerial vehicle within that map. The essence of SLAM lies in its ability to provide machines a comprehensive spatial understanding of the environment, allowing for autonomous navigation, obstacle avoidance, and efficient data collection [26].

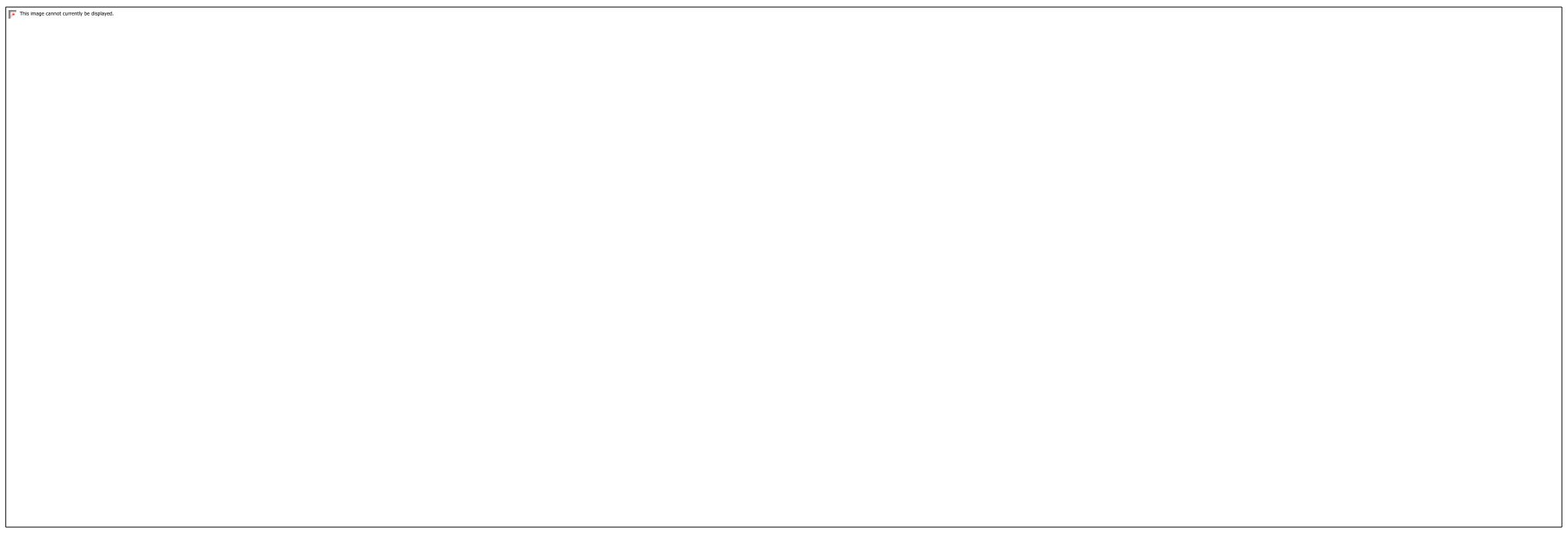

Figure 10: Block diagram of localization and mapping process

The steps involved in the process of SLAM are depicted in Figure 10. It extracts features which are then tracked in the subsequent frames and the device's motion is estimated using data association that involves matching the identified features across frames to establish correspondences. Then loop closures are detected identifying the locations that are revisited and a map is represented and refined during this process, which is then visualized with the device's location in it.

SLAM has wide-ranging applications in aerial data analysis and is pivotal for drones and UAVs to navigate through environments, avoiding obstacles and ensuring precise positioning. It is instrumental in tasks such as environmental monitoring, precision agriculture, infrastructure inspection, and disaster response [3]. For instance, in environmental monitoring, SLAM can be used to create detailed 3D maps of landscapes, enabling the assessment of terrain changes, vegetation health, and ecological dynamics. In precision agriculture, SLAM contributes to mapping and monitoring crop fields, optimizing resource allocation, and improving overall crop management [3]. Past works in SLAM for aerial data analysis have focused on enhancing the robustness and efficiency of algorithms, addressing challenges such as scale variations, dynamic environments, and sensor noise. The continuous advancements in SLAM technology for aerial data analysis contribute to the development of more capable and autonomous UAVs, expanding their applications in various domains.

Table 3 provides a compilation of works related to computer vision tasks discussed earlier. As most of these tasks are interdependent on each other and many of the studies have considered more than one task in their analysis, we've consolidated the studies into a single table, outlining the specific task(s) addressed in each study.

This approach allows for a more cohesive overview, eliminating the need for separate tables dedicated to individual tasks.

Table 3: Task based analysis of various parameters in existing literature for aerial data analysis

| S.No. | Task Performed | Reference | Contribution | Data Type used | Data Collection Details (Method, Tool) | Architecture | Dataset | Libraries used | Evaluation Metrics | Application Area(s) | Hyper parameters configuration |
|---|---|---|---|---|---|---|---|---|---|---|---|
| 1 | Object Recognition and Detection | [57] | Evaluation of CNN training procedure, Exploring the parameter selection to enhance the performance of the models. | Image, Video | Satellite imagery sourced from Google Maps, depicting airplanes stationed on different airfields | YOLO | - | - | Confusion Matrix, Accuracy | - | Learning rate: 0.0001; batch size: 64; momentum: 0.5; decay: 0.00005; no. of iteration: 45000; detection threshold: 0.2 |
| 2 | Object Recognition | [58] | An algorithm with three CNNs was proposed for recognition of marine objects. | Image | Using Google satellite images and images collected from Google Image Search | CNN | - | - | Accuracy | Marine | - |
| 3 | Moving Object Recognition | [100] | A large-scale video dataset namely MOR-UAV was introduced, A deep unified framework MOR-UAVNet for MOR was proposed. | Video | Videos captured from highways, flyovers, traffic intersections etc., in normal and adverse conditions | ResNet | MOR-UAV | Keras, Tensorflow | AP with IoU threshold 0.50 | - | - |
| 4 | Object Classification | [101] | A comparative study on object classification on UAV datasets, using various ML/DL models. | Image | Combination of annotated low-altitude UAV datasets CARPK, Okutama, VEDAI, Birds, and UAVBD | CNN | CARPK, Okutama, VEDAI, Birds, and UAVBD | Scikit-learn; OpenCV; Keras; TensorFlow | Precision, Recall, F1 measure, Accuracy | - | Learning rate: 0.001; epochs: 1000; drop-out rate 0.2; batch size: 32 |
| 5 | Object Recognition | [102] | An enhanced YOLOv4_Tiny model was proposed for UAV images specifically tailored for dense small objects. | Image | - | YOLOv4_Tiny (CSPdarknet53_tiny) | VisDrone | - | mAP, Frames per Second | - | - |
| 6 | Image Recognition and Object Detection | [103] | CNN based models were used for the identification of the shoal of fish for fishing, using transfer learning | Image, Video | UAV is used to capture videos and images in the open sea. | Inception V3, MobileNet V2, and NASNet-A (large) | - | Tensor Flow | Accuracy | Marine | Training steps: 4000; learning rate: 0.01; training batch size: 100; testing batch size: 1; validation batch size: 100 |
| 7 | Image Classification | [104] | Synthetic data was used for training the classification models using CDNTS to work on actual classification problem | Image | Synthetic data was generated using Unity Game Engine (generally used to create 2-D and 3-D games), Real data from internet | AlexNet, SqueezeNet, and VGG16 | - | Tensor Flow, Keras, OpenCV | F-Score, Accuracy, AUC | Surveillance | Optimizers: Adam/Adadelta/RMSprop/SGD; Batch Size: 32/128/256; Learning Rate: 0.1/0.01/0.001; Dropout: 0.5/0.8 |
| 8 | Object Recognition | [105] | The efficiency of ViT along with data augmentation and transfer learning, for image classification of – crop and weed were examined. | Image | Starfury Pilgrim UAV equipped with a Sony ILCE-7R; 36 mega pixel camera was used to capture images of crop fields | ViT-B32 and ViT-B16 | - | Tensor flow 2.4.1, Keras 2.4.3 | precision, recall and F1-Score | Agriculture | Learning rate: 0.0001; batch size: 8; epoch: 100 |
| 9 | Object Detection | [65] | A method for automatic content-based object detection was proposed. | Image | High resolution images chosen from USGS national maps [73] | GoogleNet | UCMerced | Caffe | Accuracy | Land use classification | Learning rate: 0.01 decaying to 50% every 16 epochs. |
| 10 | Object Detection | [106] | An object detection model based on YOLOv3 was proposed to address the problem of small object. | Image | UAV-viewed dataset created using images from UAV123 videos and images captured by the researchers | YOLOv3, ResNet | UAV123 | - | mAP, IoU | Human Detection | - |
| 11 | Object Detection and Object Tracking | [68] | Deep supervision to address vanishing gradient problem in object detection was proposed. LSTM was used for detected object tracking | Image, Video | Frames are taken from UAV123 videos | CNN, RNN | UAV123 | Tensor Flow 1.14 | precision, recall | - | - |
| 12 | Object Recognition and Object Detection | [64] | Various methods (SSD/SSH/YOLO/ TinyFace/U-Net/ Mask R-CNN) were compared to recognize and detect low-resolution small object | Image | Real-life images provided by the Illinois Natural History Survey at the University of Illinois at Urbana-Champaign, LBAI is created. | SSD, SSH, U-Net: VGG-16; YOLOv2: Darknet, TinyFace, Mask R-CNN: ResNet-101 | LBAI | SSD, SSH: Caffe; TinyFace: Matconvent; | precision, recall, F1-Score, MAE | Wildlife Monitoring | SSD: Learning Rate: 0.0000001, batch size 16; SSH: Learning Rate: 0.0004 for easy and 0.004 for hard cases, Momentum: 0.9, decay: 0.0005; YOLO: Learning Rate: 0.0001, decay: 0.0005, batch size 64, subdivisions: 8; TinyFace: Learning Rate: 0.00001 for easy and 0.0001 for hard cases, |
| | | | | | | | | | | | Momentum: 0.9, decay: 0.0005, epochs: 50; U-Net: Learning Rate: 0.001, decay: 0.1 for every 7 epochs, batch size: 2, Optimizer: Adam; Mask R-CNN: Learning Rate: 0.0001, Optimizer: Adam |
| 13 | Object Detection | [107] | A real-time UAV based object detection method using YOLOv5 was presented for detection of Forest fires. | Image, Video | Images from FLAME dataset and FireNet dataset were combined | CSP Darknet | FLAME, FireNet | RoboFlow | precision, recall, F1-Score | Wildfire Detection | Optimizer: SGD; learning rate: 0.00001; batch size: 16; epochs: 350 |
| 14 | Object Detection | [108] | An end-to-end model unifying object clustering and detection was presented, for efficient detection of sparsely distributed small objects. | Image | - | FPN | VisDrone, UAVDT, DOTA | Caffe2, Detectron | AP, $AP_{50}$ (calculated at IoU threshold 0.5 over all categories) and $AP_{75}$ (calculated at IoU threshold 0.75 over all categories) | - | VisDrone, UAVDT: base learning rate: 0.005; total iteration: 140k; learning rate after first 120k iterations: 0.0005; momentum: 0.9; decay: 0.0005 DOTA: learning rate: 0.005; total iterations: 40k; NMS threshold: 0.5 |
| 15 | Object Detection | [109] | An OAN in conjunction with object detectors was used to identify objects within a patch. | Image | - | R50FPN | DOTA | OBBDetection | Accuracy | - | NMS threshold: 0.1 |
| 16 | Image Semantic Segmentation | [110] | An aerial video dataset was presented for semantic segmentation. | Video | Videos were captured using DJI Phantom 3 professional drone at an altitude of 25 meters and resolution of 1280 × 720 at 29 fps. | FCN with VGG-16, U-Net | MANIPAL UAV Aerial Video Dataset | - | mIoU, PA, precision, recall, F1-score | - | Batch size: 10; |
| 17 | Image Instance Segmentation | [88] | An aerial video dataset was presented for post disaster building damage assessment using segmentation and a neural network model MSNet. | Video | Aerial videos recorded post natural disasters like hurricanes and tornados collected from social media platforms. | ResNet-50 FPN | ISBDA | - | AP, $AP_{25}$, $AP_{50}$, IoU | Disaster Management | Epochs: 100; base learning rate: 0.003; learning rate after 10 epochs: 0.001; Optimizer: Mini-batch SGD; batch size: 8; NMS threshold: 0.5 |
| 18 | Image Semantic Segmentation | [111] | A DL-based model using two GANs is proposed for the purpose of semantic segmentation. | Image | - | ResNet-101 | ISPRS | Tensorflow | precision, recall, f -score, accuracy | - | IOptimizer: Adam; Learning rate: 0.0001; epochs: 80 |
| 19 | Image Semantic Segmentation | [86] | A comparative study of various segmentation models based on fully convolutional architectures, namely U-Net, SegNet, FC-DenseNet and DeepLabv3+ was | Image | Images were captured by an RGB camera mounted on Phantom 4 advanced quadcopter in Campo Grande municipality, in the state of Mato Grosso do Sul, Brazil. | U-Net, SegNet, FC-DenseNet, DeepLabv3+ | - | Keras | F1 -score, accuracy, IoU | Forest conservation | Optimizer: Adam; learning rate: 0.0001, epochs: 50 |
| | | | carried out. | | | | | | | | |
| 20 | Image Semantic Segmentation | [87] | An ensemble of SegNet and U-Net models, named SegUnet was proposed for semantic segmentation of buildings. | Image | - | U-Net, SegNet | Massachusetts building dataset | Tensor Flow, Keras | precision, recall, f1 -score, accuracy | Urban Planning | Optimizer: Adam; learning rate: 0.0001 |
| 21 | Image Instance Segmentation | [89] | A deep neural network model named as CableNet based on FCN was proposed for detection of power transmission lines using instance segmentation. | Image | Images are taken during the process of practical inspections of power transmission line | VGG-16 | - | - | Detection: TPR, FPR, Cable-hit ratio Segmentation: pixel level accuracy, instance level accuracy | Infrastructure monitoring | Optimizer: Adam; learning rate: 0.0005; batch size: 4 |
| 22 | Image Instance Segmentation | [112] | A Mask R-CNN model for instance segmentation of artificial and natural water bodies was presented. | Image | Aerial images are collected from Bing map in the northeastern region of Thailand. | ResNet-101 | AIWR | Tensor flow | mAP, IoU | Environmental monitoring | Optimizer: SGD; learning rate: 0.0001; momentum: 0.9; decay: 0.0005; batch size: 4 |
| 23 | Change Detection | [77] | A DL-based architecture with parallel deep CNNs, named ChangeNet was | Image | The images in these datasets resemble drone captured images. | ResNet-50 | VL-CMU-CD, Tsunami | Tensor flow | precision, recall, f -score, AUC and IoU | - | - |
| | | | presented to detect structural changes. | | | | and GSV | | | | |
| 24 | Change Detection | [113] | A change detection model using DRoINs was presented to identify and locate object level change in a pair of images. | Image | A framework was proposed for generation of a different image using GANs and a dataset for Aerial change detection was generated. | ResNet-50 | CDnet, AICD | - | IoU | - | Optimizer: Adam |
| 25 | Change Detection | [114] | A 2-stream Siamese network for change detection in bi-temporal images was presented with an intensity loss function and a constructive loss function. | Image | Two bi-temporal image datasets were considered. In one, the image pair is collected using QuickBird satellite. In another, the image pair was taken from Google Earth. | Siamese Network | - | Caffe | False Alarm Rate, Missed Alarm Rate, Overall Alarm Rate, Kappa | - | Optimizer: SGD; learning rate: 0.001; momentum: 0.9; decay: 0.0005; batch size: 512 |
| 26 | Change Detection | [56] | An aerial dataset, namely BANDON, comprising of off-nadir aerial images of buildings was presented along with a multi-task learning based MTGCD-Net model for building change detection. | Image | Dataset is created by collecting off-nadir aerial images from Google Earth images, Microsoft Virtual Earth images and ArcGIS images. The images having a spatial resolution of 60 cm are collected. | ResNet-50 | BANDON | - | precision, recall, f1 - score, IoU | Urban Planning | Optimizer: SGD; Base learning rate: 0.01; batch size: 16; iterations: 40,000 |
| 27 | Change Detection | [75] | CNN-based models were evaluated using transfer learning for land use | Image | Samples of images are manually extracted from digital orthophotos of the City of Surrey, BC, | AlexNet, GoogLeNet and | UC-Merced | - | Accuracy | Urban Planning | - |
| | | | change analysis. | | Canada. | VGG Net | | | | | |
| 28 | Change Detection | [76] | A symmetric FCN within pyramid pooling called FCN-PP was proposed for end-to-end change detection for landslide mapping. | Image | Zeiss RMK TOP 15 aerial survey camera system was used to capture five pairs of bi-temporal aerial images on five areas in Hong Kong at height 2400 m with resolution 50 cm. | FCN | - | MATLAB, PyTorch | Precision, Recall, F - score, Accuracy | Disaster Monitoring | learning rate: 0.00001; decay: 0.0005; momentum: 0.99; mini-batch size: 4; epochs: 30 |
| 29 | Anomaly Detection, Scene Reconstruction | [79] | Unsupervised Anomaly Detection | Image | Used Parrot Bebop 2 drone for data collection; around Aarhus University DeepTech Experimental Hub in Aarhus (Denmark) at 30-meter altitude under different weather conditions. | Object Detector: MobileNetV2-SSDLite; Anomaly Detector: novel DNN architecture | Object Detector: VisDrone; Anomaly Detector: Collected using drone | - | Scene reconstruction: precision, recall, F1-score; Anomaly detection: accuracy | Surveillance of Critical Infrastructures | Object Detector: Optimizer: RMSProp; Momentum: 0.9; Learning rate: 0.05; Batch Size: 32; Anomaly Detector: Optimizer: Adam; Learning rate: 0:001; beta1: 0:9; beta2: 0:999; Batch size: 64. |
| 30 | Anomaly Detection | [78] | Proposed an anomaly detection approach, combining deep features extracted using a pretrained CNN with an unsupervised classification | Video | A mini-drone Phantom 2 Vision+ was used to capture videos. Videos were recorded with different positions, altitudes and movements of the UAV. | GoogLeNet pretrained on ImageNet | MDVD Dataset | - | Confusion matrix, AUC | Surveillance | Kernel for OCSVM: RBF |
| | | | method, namely OCSVM. | | | | | | | | |
| 31 | Anomaly Detection | [80] | A transformer-based model, ANDT for anomaly detection in aerial videos was presented. | Video | Aerial videos were collected from YouTube and Pexels. | Transformer | AU-AIR-Anomaly | - | AUC | Traffic surveillance | patch size: 16 × 16; number of Transformer layers:2; number of attention heads: 6; MLP size: 4096 |
| 32 | Anomaly Detection | [81] | A GAN- based model with two branches, one each for detection and localization of abnormal or anomalous scenarios in video frames was proposed. | Video | A subset of 20 videos from the change detection dataset UMCD was adapted for anomaly detection. | GANs | UMCD | Pytorch | Anomaly detection: AUC, Anomaly Localization: SSIM | Surveillance | Optimizer: AdamW; learning rate: 0.0002; decay: 0.01; epochs; 200 |
| 33 | Anomaly Detection | [82] | A convolutional variational autoencoder based future frame prediction network was proposed to detect anomalies. | Video | UAV videos were searched using queries for various anomalies from the internet. | - | - | - | Precision, recall, F1 score, AUC, overall accuracy | Surveillance | - |
| 34 | Anomaly Detection | [83] | A DL-based model for extracting spatial features and a RNN for extracting temporal features was proposed for anomaly. | Video | - | VGG-16, LSTM | Mini-drone Video, UMN | - | AUC | Surveillance | Optimizer: Adam; learning rate: 0.00001; dropout rate: 0.5. |
| 35 | Object Detection, | [115] | An aerial video benchmark dataset for traffic anomaly | Video | The videos are recorded using DJI MAVIC MINI 2, at | - | UIT-ADrone | - | Object Detection: mAP; | Traffic Surveillance | - |
| | Anomaly Detection | | detection and object detection was presented. | | 30 fps from 50 to 70 m height, with a resolution of 1920 × 1080 pixels in Ho Chi Minh City, Vietnam. | | | | Anomaly detection: AUC | | |
| 36 | Object Tracking | [73] | A real time object tracking approach for fast moving object tracking in low quality videos were presented. | Video | - | HOG, RGB histogram, motion histogram | OTB2013 | Matlab | AP | Traffic Surveillance | learning rate: 0.20 |

# 3. Architectures, Datasets, Evaluation metrics, Libraries and Hyper parameters

As the application of computer vision into aerial data analysis continues to advance, the foundational elements like architectures, datasets, evaluation metrics, libraries and hyper parameters play a pivotal role in shaping the efficacy and precision of the applied methodologies. Different architectures are used in the analysis of aerial data including CNNs, RNNs, or tailored architectures, the selection profoundly influencing the model's ability to discern objects, detect changes, and perform scene-level analysis [36]. Similarly, the relevance and diversity of datasets are paramount in training robust models for aerial data analysis. A variety of datasets capturing various environmental conditions, terrains, and scenarios enable the development of algorithms capable of addressing the intricacies inherent in real-world aerial imagery [42]. The utilization of specialized libraries streamlines the implementation of complex tasks. Widely adopted libraries offer a wealth of pre-built functions, empowering researchers to focus on the unique challenges posed by aerial data rather than the intricacies of algorithmic implementation [104], [105]. Optimizing hyper parameters is a critical facet in tailoring different models for specific aerial data analysis tasks. Parameters such as learning rates, batch sizes, and model depths demand meticulous tuning to enhance the efficiency and accuracy of algorithms in detecting, classifying, and interpreting aerial features [116], [117]. Robust evaluation metrics are essential for gauging the performance of various models in aerial data analysis. Precision, recall, F1-score, and metrics like IoU provide a quantitative framework to assess the accuracy and efficacy of object detection, segmentation, and other tasks [118]. The subsequent examination of each component will unravel the complexities, challenges, and opportunities that arise in the field of aerial data analysis. In this section, a review of different architectures, datasets, evaluation metrics, libraries and hyper parameters used for aerial data analysis shall be provided.

## 3.1. Architectures for computer vision tasks

Architectures for computer vision tasks refer to the specific frameworks, or models designed to process visual data and derive meaningful insights from it. These architectures are designed to manage the complexity in performing various tasks such as image recognition and classification, object detection, image segmentation, and other visual understanding applications. Prominent examples of such architectures include AlexNet [119], VGGNet [120], and GoogLeNet [121], among others. These architectures predominantly rely on DL-based techniques to efficiently perform various tasks. It's noteworthy that DL-based approaches gained popularity in the past decade, whereas earlier methods involved simpler architectures that relied on manual feature extraction using descriptors like SIFT, SURF, BRIEF, FAST, Hough Transform and others [122]. SIFT identifies features in an image that remain consistent across different scales and orientations. It involves identifying and describing local features, such as corners or blobs, in an image. It is invariant to translation, rotation, and scale changes, making it robust for various tasks. SURF accelerates the computation of feature descriptors by utilizing integral images and approximations [122]. It is used for detecting robust and distinctive image features efficiently.

BRIEF represents image features using binary descriptors instead of producing continuous-valued feature descriptors. It generates binary strings that represent the presence or absence of certain patterns in the image. It is computationally efficient and suitable for real-time applications. FAST is designed for rapid corner detection in images [122]. It identifies corners by comparing pixel intensities in a circular region. It is very fast, making it suitable for applications where real-time performance is crucial. Hough Transform focuses on detecting shapes, particularly lines and circles, within an image. It is robust to noise and can detect shapes even if some part is missing or incomplete. The challenge with these traditional approaches is identifying which features are important in each image. It is done explicitly by manual intervention [122]. It becomes very tedious to extract features as the number of classes increase in the dataset. In DL-based models, the whole process is automatized and carried out faster. Some popular DL-based architectures are explained in the following sub sections with a summary of major path breaking architectures in Table 4.

#### *3.1.1. LeNet-5*

LeNet-5 is a CNN architecture which was designed for handwritten digit recognition consisting of five learnable layers, including three convolutional layers followed by average-pooling, and two fully connected layers [123]. The activation function used is the tanh function. LeNet-5 was instrumental in demonstrating the effectiveness of DL for image classification tasks, particularly in the domain of recognizing hand-written digits for tasks like ZIP code reading. Its architecture and principles laid the foundation for modern vision networks, influencing subsequent advancements in DL and CNNs.

#### *3.1.2. AlexNet*

AlexNet is a deep CNN architecture that won the ILSVRC-2012 for image classification tasks; consisting of five convolutional layers followed by three fully connected layers, utilizing ReLU for activation and incorporating max-pooling for down sampling. It introduced LRN and dropout for regularization, demonstrating a top-5 error rate of 16.4% on ImageNet, outperforming traditional methods [119]. This architecture is used in image classification task for aerial data [104].

#### *3.1.3. GoogLeNet*

GoogleNet, also known as Inception-v1, is a deep CNN architecture designed by Google researchers, that won the ILSVRC in 2014 for image classification [121]. It is a multi-scale architecture that uses inception modules for employing parallel convolutional operations with different filter sizes thus allowing the network to capture features at multiple scales. It also addresses the vanishing gradient problem by utilizing auxiliary classifiers at intermediate layers during training. It achieved a top-5 error rate of approx. 6.67% in ILSVRC-2014. This architecture was further utilized for object detection in aerial data [65].

#### *3.1.4. VGGNet*

VGGNet is a deep CNN architecture introduced by the Visual Geometry Group at Oxford University. It achieved second place in the ILSVRC-2014 for image classification, with a top-5 error rate of approx. 7.3% and first place for detection task [120]. It has a uniform architecture, where all convolutional layers have a small 3x3 filter size and all max-pooling layers have a 2x2 filter size with a stride of 2. The depth of VGGNet can be adjusted by stacking more convolutional layers, leading to variations like VGGNet-16 and VGGNet-19. VGGNet demonstrated strong performance in image classification tasks, emphasizing the importance of depth in CNNs. This architecture was utilized for image classification task in aerial datasets [104].

#### *3.1.5. ResNet*

ResNet is a deep CNN architecture that introduced the innovative concept of residual blocks using skip connections [124]. These skip connections or shortcut connections, bypass one or more layers in between, while making network connections and this forms a residual block. This unique approach mitigates the vanishing gradient problem, enabling the training of exceptionally deep networks. ResNet architectures can have varying depths, such as ResNet-18, ResNet-34, ResNet-50, ResNet-101 or ResNet-152, where the number indicates the number of layers in the network. It has proven effective for various tasks, including image classification, object detection, and other visual recognition tasks. It achieved a top-5 error rate of 3.57% in ILSVRC-2015 for image classification. This architecture was used for moving object recognition in aerial datasets [100].

#### *3.1.6. DenseNet*

DenseNet is a deep CNN architecture that uses the concept of dense connectivity, where each layer receives input from all preceding layers in the network [125]. It organizes layers into dense blocks, where each layer is connected to all previous layers within the block. This architecture addresses challenges related to information flow, vanishing gradients, and parameter efficiency. The dense connectivity strengthens feature propagation and enhances feature reuse, allowing each layer to directly access the feature maps of all previous layers and substantially reduces the number of parameters. This architecture was utilized for semantic segmentation task in aerial data [86].

#### *3.1.7. R-CNN, Fast R-CNN, Faster R-CNN*

These are a series of 2-stage object detection models with R-CNN laying the foundation, and Fast R-CNN, Faster R-CNN significantly improve training speed and detection accuracy. In the first stage, regions are generated with potential bounding box proposals, called region proposals; and in the second stage, CNN is applied for classification and bounding box refinement. In R-CNN [126], region proposals are generated using selective search algorithm and CNN is applied to each proposed region. In Fast R-CNN [127], instead of applying CNN to each region proposal, it is applied to entire image in a single pass, thus it is faster. In Faster R-CNN [128], region proposals are generated using an RPN and it is combined with Fast R-CNN.

#### *3.1.8. SSD*

SSD is a single stage object detection model based on DL that can detect objects in various scales and aspect ratios in a single pass [129]. It utilizes convolutional layers of different sizes to process input images at multiple scales and incorporates multiple feature maps of different resolutions to capture object details at various scales thus, enabling the detection of objects of different sizes. Instead of predicting bounding boxes directly, SSD employs a set of default boxes with different aspect ratios at each spatial location in the feature maps. Predictions are then adjusted based on these default boxes. SSD predicts both the locations and class scores for each default box in a single forward pass. After prediction, NMS is applied to filter out redundant bounding boxes and keep the most confident detections. This model was used for object recognition and detection in aerial scenarios [64].

### *3.1.9. YOLO*

YOLO is a single stage real time object detection architecture that processes the input image in one pass, by dividing it into a grid and then predicts bounding boxes, confidence for the boxes and class probabilities within each grid cell [130]. It uses anchor boxes to refine bounding box predictions for different object scales and shapes. It is faster than other object detection methods processing at 45 frames per second in real-time. This architecture was used for object detection in aerial visual data [57], [102], [106].

### *3.1.10. Mask R-CNN*

Mask R-CNN is an object instance segmentation method that detects objects in an image and generates a segmentation mask to delineate boundaries of each instance of the object [91]. This method is an extension of Faster R-CNN architecture for predicting segmentation masks along with bounding box coordinates and class labels using a parallel dedicated branch for the same. The architecture was used for segmentation in aerial data [112].

### *3.1.11. U-Net*

U-Net is a CNN architecture, deriving its name from its U shape, that consists of a contracting path, a bottleneck, and an expansive path [90]. The contracting path, also known as the encoder, down samples the input image to capture high-level features such as context. The expansive path, or decoder, up samples the features enabling precise localization to generate a segmentation map. It makes use of skip connections that directly link layers from the contracting path to the corresponding layers in the expansive path. Skip connections facilitate the transfer of detailed spatial information, aiding in the precise localization of objects. It uses both convolutional and transposed convolutional layers in the contracting and expansive paths, respectively. U-Net is often employed in conjunction with data augmentation techniques to artificially increase the size of the training dataset, enhancing the model's generalization capability. This architecture was used for segmentation tasks in aerial visual data [86], [110].

### *3.1.12. SegNet*

SegNet is a deep convolutional encoder-decoder architecture consisting of 13 convolutional layers corresponding to convolutional layers in VGG16 network and for each encoder layer there is a corresponding decoder layer [131]. The encoder extracts spatial information and hierarchical features from the input image, and the decoder reconstructs the segmented output. The architecture was used for image segmentation in aerial data [87].

*3.1.13. FCN*

FCN is a deep CNN architecture that is generally used for semantic segmentation tasks. It can process entire image and produce pixel-wise predictions [132]. The deep CNN models for classification like AlexNet, VGGNet, GoogLeNet can be casted to FCNs and their learned representations can be transferred to the segmentation task by fine-tuning. FCNs incorporate skip connections or skip layers, which enable the network to capture both high-level semantic features and fine-grained details. These connections combine features from different network layers to enhance the model's understanding of both global and local context. FCN was used for image segmentation in aerial datasets [89].

*3.1.14. Siamese Network*

Siamese Network is an object tracking architecture based on CNNs [133]. It focuses on tracking arbitrary objects in a video by means of similarity learning, where a score is assigned depending on the similarity of exemplar image to candidate image which signifies if the two images represent the same object. If not, we can then exhaustively test all possible locations and choose the candidate with the maximum similarity to the past appearance of the object.

*3.1.15. ChangeNet*

ChangeNet is a DL-based change detection architecture that detects changes between pairs of images, leveraging transfer learning for efficient feature extraction [77]. The test image and reference image are fed into the DL model for feature extraction. The features extracted from both images are then compared to identify the changes and these changes are labeled or classified based on the analysis of extracted features.

Various other popular DL architectures are employed in performing different computer vision tasks. For instance, OverFeat is utilized for single-pass object classification and detection, employing a sliding window approach to process images [134]. ZFNet, the winner of ILSVRC-2013, provides visual insights into intermediate feature layers, utilizing smaller filter sizes in convolution layers for detailed feature extraction in classification tasks [135]. ResNeXt, an extension of ResNet, incorporates an additional cardinality parameter representing the number of independent paths in each residual block [136]. Additionally, MobileNet [137] and SqueezeNet [138] serve as lightweight CNN architectures specifically designed for mobile and embedded vision applications. SENet [139] introduces a channel-wise attention mechanism for image classification tasks, while EfficientNet [140] focuses on achieving an optimal trade-off between efficiency and accuracy in image classification models. Xception introduces depth-wise separable convolutions with residual connections, reducing the number of parameters and computation cost [141]. Further, U-Net++, an extension of the UNet architecture with a nested design, is applied for semantic segmentation tasks [142]. DeconvNet, also a semantic segmentation architecture, consists of deconvolutional and unpooling layers using the VGG-16 network backbone [143]. DeepLab, designed for semantic segmentation, employs dilated or atrous convolutions to increase the receptive field of the network without down-sampling the input [144]. RegNet, designed for image classification tasks, uses convolutional RNNs as regulator modules to extract complementary features from intermediate layers alongside ResNet [145]. 3D CNNs, applied in action recognition tasks, extract features from both spatial and temporal dimensions [146]. Furthermore, transformer-based vision architectures include ViT, designed for image classification tasks based on a self-attention mechanism, allowing the model to adaptively focus on different parts of the image [147]. SegFormer, a transformer-based framework, utilizes MLP decoders for semantic segmentation tasks [148].

Table 4: Summary of architectures used in different tasks

| S. No. | Architecture | Proposed Year | Task | Layers details | Pooling | Non-Linearity | Dataset (in the base work) | Remarks |
|---|---|---|---|---|---|---|---|---|
| 1. | LeNet or LeNet-5 [123] | 1998 | Character recognition | 3 CLs, 2 FCLs | Average | tanh | MNIST | Simple introductory neural network for recognition tasks. |
| 2. | AlexNet [119] | 2012 | Image Classification, Object Recognition | 5 CLs, 3 FCLs | Max | ReLu | ILSVRC-2010, ILSVRC-2012 (ImageNet) | First deep CNN model to use GPUs for training. |
| 3. | GoogLeNet or Inception v1 [121] | 2014 | Image Classification, Object Recognition; Object Detection | Parallel CLs of different sizes and max pooling layers along with FCLs; 22 layers | Average | ReLu | ILSVRC-2014 (ImageNet) | Uses inception modules to capture features at multiple scales, auxiliary classifiers to address vanishing gradient problem. |
| 4. | VGGNet [120] | 2014 | Image Classification, Object Recognition; Object Detection | 13 & 16 CLs in VGGNet-16 and VGGNet-19 respectively, 3 FCLs | Max | ReLu | ILSVRC-2014 (ImageNet) | Emphasizes the importance of depth in CNNs. |
| 5. | ResNet-18 [124] | 2015 | Image Classification, Object Recognition; Object Detection | 18 CLs, 1 FCL | Average | ReLu | ILSVRC-2015 (ImageNet) | Uses residual learning framework enabling the training of exceptionally deep networks. |
| 6. | DenseNet-121 [125] | 2017 | Image Classification, Object Recognition | 120 CLs, 1FCL | Average | ReLu | CIFAR-10, CIFAR-100, SVHN, and | Uses the concept of dense connectivity. |
| | | | | | | | ImageNet | |

### 3.2. Datasets for Computer Vision Tasks

A dataset is a structured collection of data, organized and presented in a specific format for analysis, interpretation, or further processing. In the context of computer vision, a dataset typically consists of a set of images or videos, each associated with specific annotations or labels indicating the actual output for the given task. This field relies heavily on the availability of diverse and well-annotated datasets to train and evaluate models [149], [150]. These datasets serve as the foundation for developing and testing algorithms across various vision tasks. Each dataset is curated to address specific challenges associated with tasks such as image classification, object detection, semantic segmentation, and more. These datasets play a crucial role in fostering innovation, benchmarking algorithms, and enabling the development of robust and accurate solutions [149]. In this section, a brief review of the commonly used aerial datasets in the visible spectrum for various tasks is carried out and summarized in Table 5.

#### *3.2.1. MOR-UAV*

This dataset includes aerial videos captured by UAVs, comprising 30 videos with 10,948 frames and 89,783 bounding box annotations, recorded at a rate of 30 fps, with resolutions ranging from 1280 × 720 to 1920 × 1080. It addresses challenges like moving objects, occlusion, nighttime, and weather changes. The dataset can be used for applications such as aerial surveillance, search and rescue, event recognition, and urban and rural scene understanding [100].

#### *3.2.2. BirdsEyeView*

BirdsEyeView is an aerial image and video dataset, designed for object classification and detection tasks, consisting of aerial images from UAVs and other datasets. It contains 5000 images and approximately 10,000 annotations, addressing diverse scenarios with different resolutions, angles, and views. The dataset covers six classes, including parking lots, action tests, routine life, outdoor living, harbor, and social party [151].

#### *3.2.3. VisDrone*

VisDrone is a large-scale aerial dataset focusing on object detection in images and videos, single and multi-object tracking in videos, and crowd counting released for vision meets drone challenge. It includes aerial images and videos captured by UAVs. This dataset has many versions depending the year of release. Starting in 2018, VisDrone-Det2018 object detection challenge, a dataset [152] consisting of 8599 images was released; for VisDrone-SOT2018 single-object tracking challenge, a dataset [153] consisting of 132 videos with 106,354 frames was released. For VisDrone-VDT2018 object detection in videos and multi-object tracking challenge, a dataset consisting of 79 video clips with 33, 366 frames and about 1.5 million bounding box annotations was released [154]. The datasets were augmented over years and in 2020, an additional dataset for crowd counting was released for VisDrone-CC2020 challenge with 3360 images [155]. The latest version of the datasets for these challenges is summarized in [149]. The VisDrone dataset contains 10,209 images and 288 video clips with over 2.6 million bounding box annotations. The dataset covers ten classes like pedestrian, person, car, van, bus, truck, motor, bicycle, awning-tricycle, and tricycle; addressing various weather and lighting conditions.

#### *3.2.4. UAVDT*

This dataset involves aerial videos captured by UAVs for object detection and tracking. It includes 100 video sequences with 80,000 frames and 840,000 bounding box annotations. The dataset addresses challenges related to weather conditions, flying altitudes, and camera views, focusing on one class: vehicles [156].

#### *3.2.5. DOTA*

DOTA is designed for object detection and includes aerial images from multiple sensors. It consists of 2806 images with 188,282 oriented bounding box annotations. Diverse scenarios include oriented objects and multiple resolutions, covering fifteen classes such as planes, ships, storage tanks, and more [157]. In [150], an extension of the dataset as DOTA v2.0 was released with 11,268 images and 1,793,658 oriented bounding box annotations with three more classes added.

#### *3.2.6. LandCover.ai*

Specialized in semantic segmentation and change detection, this dataset comprises aerial images chosen from photographs creating a digital orthophoto covering Poland. It consists of 41 orthophoto tiles with polygon shape and polyline annotations. The dataset addresses different scenarios, seasons, and lighting conditions, covering four classes: building, woodland, water, and road [158].

#### *3.2.7. UAVid*

UAVid focuses on semantic segmentation using high-resolution aerial images from UAVs. It includes 300 images with pixel-level, super-pixel level, and polygon-level annotations. The dataset features oblique views and covers eight classes: building, tree, clutter, road, vegetation cover, static car, moving car, and human [159].

#### *3.2.8. iSAID*

iSAID dataset is an aerial image dataset created specifically for instance segmentation tasks. The dataset includes 2806 images with 655,451 polygonal bounding box annotations. Similar to DOTA dataset, it addresses oriented objects and multiple resolutions, covering fifteen classes [160].

#### *3.2.9. UIT-ADrone*

UIT-Adrone dataset is an aerial videos dataset focused on anomaly detection, comprising of 51 videos with 206,194 frames and 63,485 bounding box annotations around abnormal events. The dataset addresses different times of the day and covers two classes: people and vehicles [115].

#### *3.2.10. VALID*

This dataset encompasses object detection, panoptic segmentation, and binocular depth maps using aerial images synthesized in a virtual environment. It includes 6690 images in six virtual scenes with annotations for object detection and pixel-level annotations for panoptic segmentation. There are 30 classes in this dataset and diverse ambient conditions are featured in it [161].

Numerous other aerial datasets, tailored to specific applications in the visible spectrum, are also available. For example, datasets like HERIDAL [162] and SARD [163] focus on search and rescue operations, while FloodNet [164] and AIDER [165] are designed for post-disaster scene understanding and response. SHDL [63] serves as a dataset for surveillance and monitoring, and Agriculture-Vision [166] is dedicated to agricultural pattern analysis. Specifically addressing wildlife detection, the Aerial Elephant Dataset [167] and WAID [168] datasets are relevant. NOMAD [169] caters to emergency response, while Ttpla [170] is designed for the detection of transmission towers and power lines. For vehicular surveillance, datasets such as VSAI [171], EAGLE [172], Vaid [173], AU-AIR [174] and [175] were proposed for experimentation. NITR-Drone [176] is a dataset intended for infrastructure monitoring, and ERA [177] is curated for event recognition tasks. The dataset 3DCD [178] specializes in 3D Change Detection, Drone-Action [179] focuses on human action recognition, and Drone-Anomaly [80] is designed for anomaly detection.

Table 5: Summary of various aerial datasets for different tasks

| S. No. | Dataset | Task | Data Type (Collection Method) | No. of Samples | Annotations | Quality Specifications | Diverse Scenarios or Challenges Addressed | Classes/Categories details | Application Specific |
|---|---|---|---|---|---|---|---|---|---|
| 1 | MOR-UAV [100] | Moving Object Recognition | Aerial videos (UAV) | 30 videos with 10,948 frames | 89,783 bounding box annotations | Resolution: varies from 1280 × 720 to 1920 × 1080 pixels, fps: 30 | Moving objects, occlusion, nighttime, weather changes, camera motion, Changing altitudes, different camera views, and angles. | 2 (Car, Heavy vehicles) | Aerial surveillance, Search and rescue, Event recognition, Scene understanding |
| 2 | BirdsEyeView [151] | Object Classification, Object Detection | Aerial images (UAV), other aerial datasets | 5000 images | Around 10,000 annotations | Resolution: varies from 850 × 480 to 1920 × 1080 pixels | Different resolutions; different angles or views and different densities | 6 (Parking lot, Action test, Routine life, Outdoor living, harbor, social party) | |
| 3 | VisDrone [149] | Object tracking, Crowd counting | Aerial images and videos (UAV) | 10,209 images; 263 video clips with 179,264 frames | More than 2.6 million bounding box annotations | Maximum resolution video clips: 3840 × 2160 pixels Maximum resolution images: 2000 × 1500 pixels | Various weather and lighting conditions | 10 (Pedestrian, Person, Car, Van, Bus, Truck, Motor, Bicycle, Awning-Tricycle, and Tricycle) | |
| 4 | UAVDT [156] | Object detection, Object tracking | Aerial videos (UAV) | 100 video sequences with 80,000 frames | 840,000 bounding box annotations | Resolution: 1080 × 540 pixels, fps: 30 | Weather condition, flying altitude, camera view | 1 (Vehicles) | Vehicle Detection and Tracking |
| 5 | DOTA 2 [150] | Object detection | Aerial images (multiple sensors with multiple resolutions) | 11,268 images | 1,793,658 oriented bounding box annotations | Resolution: 800 × 800 to 4000 × 4000 pixels; 29, 200 × 27, 620 pixels and 7, 360 × 4, 912 pixels | Oriented Objects, multiple resolutions | 18 (plane, ship, storage tank, baseball diamond, tennis court, swimming pool, ground track field, harbor, bridge, large vehicle, small vehicle, helicopter, round-about, soccer ball field basketball court, container crane, airport, helipad) | - |
| 6 | LandCover.ai [158] | Semantic Segmentation, Change Detection | digital orthophoto (aerial photographs) | 41 orthophoto tiles | Polygon shape and polyline annotations | Spatial resolution: 25cm per pixel (9000 × 9500 pixels) or 50 cm per pixel (4200 × 4700 pixels) | Different seasons, time of day, weather and lighting conditions | 4 (building, woodland, water, road) | Landcover and land use monitoring |
| 7 | UAVid [159] | Semantic segmentation | High resolution aerial images from 30 4K video sequences (UAV) | 300 images | Pixel level, super-pixel level, and polygon level annotations | Resolution: 4096 × 2160 or 3840 × 2160 pixels | Oblique View | 8 (Building, Tree, Clutter, Road, Vegetation Cover, Static Car, Moving Car, Human) | - |
| 8 | iSAID [160] | Instance Segmentation | Aerial images; created on DOTA | 2806 images | 655,451 polygonal bounding box annotations | Maximum resolution: 4000 × 4000 pixels | Oriented Objects, multiple resolutions | 15 (plane, ship, storage tank, baseball diamond, tennis court, swimming pool, ground track field, harbor, bridge, large | - |
| | | | | | | | | vehicle, small vehicle, helicopter, round-about, soccer ball field and basketball court) | |
| 9 | UIT-ADrone [115] | Anomaly Detection | Aerial videos (UAV) | 51 videos with 206,194 frames | 63,485 annotations in form of bounding box around each abnormal event | Resolution: 1920 × 1080 pixels fps: 30 | Different times of day | 2 (People, Vehicles) | Traffic Anomaly Detection |
| 10 | VALID [161] | Object Detection, Panoptic Segmentation, binocular depth maps | Aerial images; (synthesized using Unreal Engine 4) | 6690 images in six virtual scenes | HBB and OBB annotations for Object detection, Pixel-level annotations for panoptic segmentation | Resolution: 1024 × 1024 pixels | Ambient conditions (sunny, dusk, night, snow, fog) | 30 (tree, plant, road, pavement, land, water, pool, ice, stone, pier rubble, bridge, sign, small vehicle, large vehicle, building, animal, person, chair, fence, garbage bin, low obstacle, telegraph pole, traffic light, bus stop, lamp, high obstacle, tunnel, ship, plane and harbor) | - |

3.3. Evaluation Metrics for Computer Vision Tasks

Evaluation metrics play an essential role in assessing the performance of any algorithm. These metrics provide quantitative measures to gauge how well the system is performing its intended tasks. The choice of metrics often depends on the specific application and the nature of the problem being addressed. Here are some commonly used evaluation metrics:

*3.3.1. Confusion Matrix*

A confusion matrix is a 2 × 2 matrix that is used to evaluate the performance of a classification algorithm on a set of data for which the true values are known [118]. It consists of four entities and is depicted as shown in Figure 11.

(i) TP: The number of instances correctly predicted as positive.
(ii) TN: The number of instances correctly predicted as negative.
(iii) FP: The number of instances incorrectly predicted as positive.
(iv) FN: The number of instances incorrectly predicted as negative.

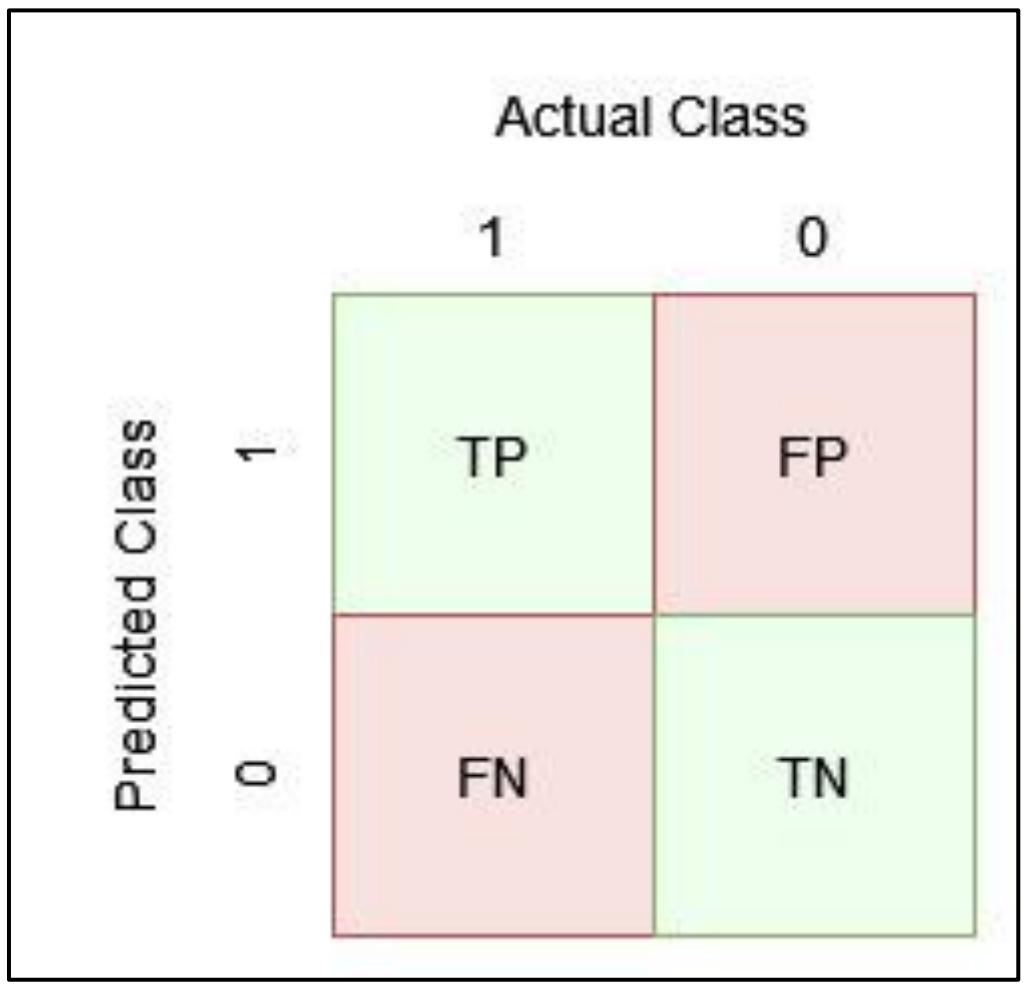

Figure 11: Confusion matrix

These entities are used to calculate other metrics like accuracy, precision and recall and especially used in performance evaluation of a model where a dataset with uneven class distribution is used.

*3.3.2. Accuracy*

It is the proportion of correctly predicted instances out of the total instances and provides a general measure of correctness across all classes. It is used to specify the overall correctness of a classification model [118] and is defined as:

$$Accuracy = \frac{TP + TN}{TP + TN + FP + FN} \tag{1}$$

*3.3.3. Precision*

It is the ratio of correctly predicted positive instances i.e., TP to the total predicted positive instances and is relevant when minimizing instances incorrectly predicted as positive i.e., FP is crucial [118]. It provides a measure of the correctness of a classification model in predicting the target class and is defined as:

$$Precision = \frac{TP}{TP + FP} \tag{2}$$

### 3.3.4. *Recall (Sensitivity or TPR)*

It is the ratio of correctly predicted positive instances i.e., TP to the total actual positive instances. It is considered when minimizing instances incorrectly predicted as negative i.e., FN is critical [118]. It provides a measure of the classification model's capability in classifying all instances of the target or positive class correctly, which is calculated as:

$$Recall = \frac{TP}{TP + FN} \tag{3}$$

### 3.3.5. *FPR*

It is the ratio of incorrectly predicted positive instances i.e., FP to the total actual negative instances. FPR indicates the proportion of actual negatives that were incorrectly classified as positives. It is also known as the Type I error or false alarm rate, and is calculated as:

$$FPR = \frac{FP}{TN + FP} \tag{4}$$

### 3.3.6. *F-Score or F-Measure or F1-Score*

F-Score, alternatively referred to as F-Measure, is the weighted average of both precision and recall with the weight function $\beta$. It proves valuable in achieving a balance between precision and recall. It provides a unified metric that takes into account both FPs (precision) and FNs (recall). When weight function $\beta$ = 1, standard F-Score is obtained and it is called F1-Score [118]. It is used in scenarios where the model's performance needs to be evaluated considering both precision and recall.

$$F - Score = (1 + \beta^2) * \frac{Precision * Recall}{\beta^2 * Precision + Recall} \tag{5}$$

with $\beta$ = 1,

$$F1 - Score = 2 * \frac{Precision * Recall}{Precision + Recall} \tag{6}$$

### 3.3.7. *Top-k Accuracy*

Top-k Accuracy evaluates whether the correct label is one of the top-k predicted labels from the model, where k represents the number of top predictions assessed [180]. This metric is commonly used in classification tasks and is especially relevant in scenarios where the classes are not mutually exclusive, and an object might belong to multiple categories.

$$Top - k\ Accuracy = \frac{Number\ of\ instances\ where\ correct\ label\ is\ in\ Top - k\ predictions}{Total\ number\ of\ instances} \quad (7)$$

*3.3.8. IoU or Jaccard Index*

It is commonly used in the context of evaluating the performance of object detection and image segmentation algorithms. It measures the overlap between the predicted and true bounding boxes or segmentation masks [181], which can be calculated as:

$$IoU = \frac{Area\ of\ Intersection}{Area\ of\ Union} \quad (8)$$

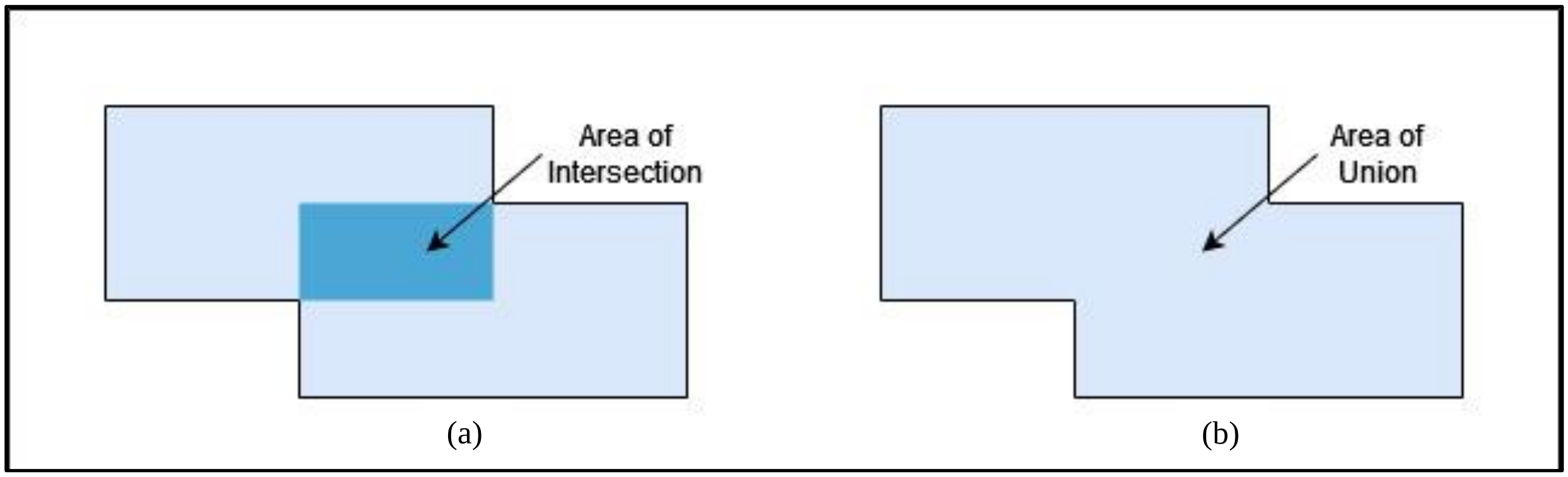

Figure 12: Area of intersection and area of union in the predicted bounding box and the actual bounding box

In Figure 12 (a), the area of intersection of the predicted bounding box and the actual bounding box is shown. In Figure 12 (b), the area combining both the bounding boxes is shown. The objective is to maximize the IoU i.e., the area of intersection of prediction and actual boxes or masks is very large thus implying that the predicted value almost overlaps the actual value. IoU is used when accurately localizing objects is crucial.

*3.3.9. Average Precision (AP)*

AP is a performance measure that assesses a model's precision-recall variation across various confidence thresholds [182] and is calculated by summing precision values at different recall levels and then averaging. It is crucial in tasks like object detection, where achieving high precision while maintaining good recall is essential. In scenarios like object detection, where both precise localization and detecting numerous objects are crucial, AP provides a valuable and comprehensive evaluation metric which is defined as:

$$AP = \frac{1}{n}\sum_{i=1}^{n} P(R_i) * (R_i - R_{i-1}) \quad (9)$$

where, $n$ is the number of recall levels,

$R_i$ is the i-th recall level

$P(R_i)$ is the precision at $R_i$,

$R_{i-1}$ is the recall at the previous level.

AP offers a comprehensive assessment of the model's performance across various IoU thresholds in object detection tasks [183]. When IoU threshold is set at 50% (0.50), a predicted bounding box overlapping with 50% or more of the actual bounding box is considered correct. AP calculated with IoU threshold set at 50% (0.50) is considered as $AP_{50}$. Similarly, AP value calculated using 75% or 0.75 as IoU decision threshold is considered $AP_{75}$ [183].

### *3.3.10. Mean Average Precision (mAP)*

The mAP extends the concept of AP to multi-class object detection scenarios. It calculates the average of AP across all classes, providing a consolidated metric for the overall performance of a model. It is computed by averaging the AP values of individual classes, and is commonly used in tasks where objects belong to different categories [184]. A higher mAP indicates superior performance in terms of precision and recall across diverse object classes. mAP calculated with IoU threshold set at 50% or 0.50 is considered as $mAP_{50}$ [183]. When AP is calculated at IoU thresholds starting from 0.50 to 0.95 and the obtained APs are averaged, then we get mAP@0.5:0.95. This metric is particularly valuable in assessing the holistic effectiveness of object detection models, offering a comprehensive evaluation of their capabilities across a range of categories [184].

$$mAP = \frac{1}{N}\sum_{i=1}^{N} AP_i \tag{10}$$

Where, $N$ is the total number of classes.

### *3.3.11. ROC curve and AUC*

The ROC curve is a graphical representation where FPR is varied on the X-axis and the corresponding change in TPR is shown on the Y-axis. The ROC curve is used to assess the performance of a binary classification model across different decision thresholds i.e., different values used for considering the classification as correct or incorrect [185]. A point on the ROC curve corresponds to a specific threshold, and the curve illustrates how the model's performance varies as the threshold changes. There are certain scenarios, where the ROC curves for different considered methods overlap or cross each other, making it difficult to decide which one performs better. In that scenario, AUC is particularly used to quantify the overall performance of a binary classification model by calculating the area under the ROC curve, with higher values indicating the better performance [185].

### *3.3.12. MSE or MAE*

MSE or MAE metric is used to measure the average squared or absolute difference between the actual and predicted values in a regression task. MSE squares the differences, giving more weight to larger errors whereas MAE treats all errors equally, providing a more straightforward representation of average prediction errors. The choice between MSE and MAE depends on the specific characteristics of the data and the task at hand [186].

$$MSE = \frac{1}{n}\sum_{i=1}^{n} (y_i - \hat{y}_i)^2 \ and \ MAE = \frac{1}{n}\sum_{i=1}^{n} |y_i - \hat{y}_i| \tag{11}$$

Where, $y_i$ is the actual value; $\hat{y}_i$ is the predicted value and $n$ is the number of observations.

These metrics provide a comprehensive set of tools for evaluating the performance of algorithms across different tasks and applications. The selection of specific metrics depends on the goals and requirements of the given problem.

### 3.4. Libraries for Computer Vision Tasks

Libraries are software tools that offer built-in functions, algorithms, and utilities to simplify the development, implementation, and deployment of system solutions. These libraries provide a collection of pre-built functions and algorithms tailored for interpreting and understanding visual information from images and videos for e.g., TensorFlow, Keras, OpenCV used in [104], [105]. They cover a broad spectrum of functionalities, ranging from basic image processing operations to sophisticated machine learning techniques. These techniques include performing tasks like object recognition, segmentation, tracking etc. Based on the functionalities covered by libraries, these can be categorized as low-level and high-level as shown in Figure 13. A summary of the major vision libraries is provided in Table 6.

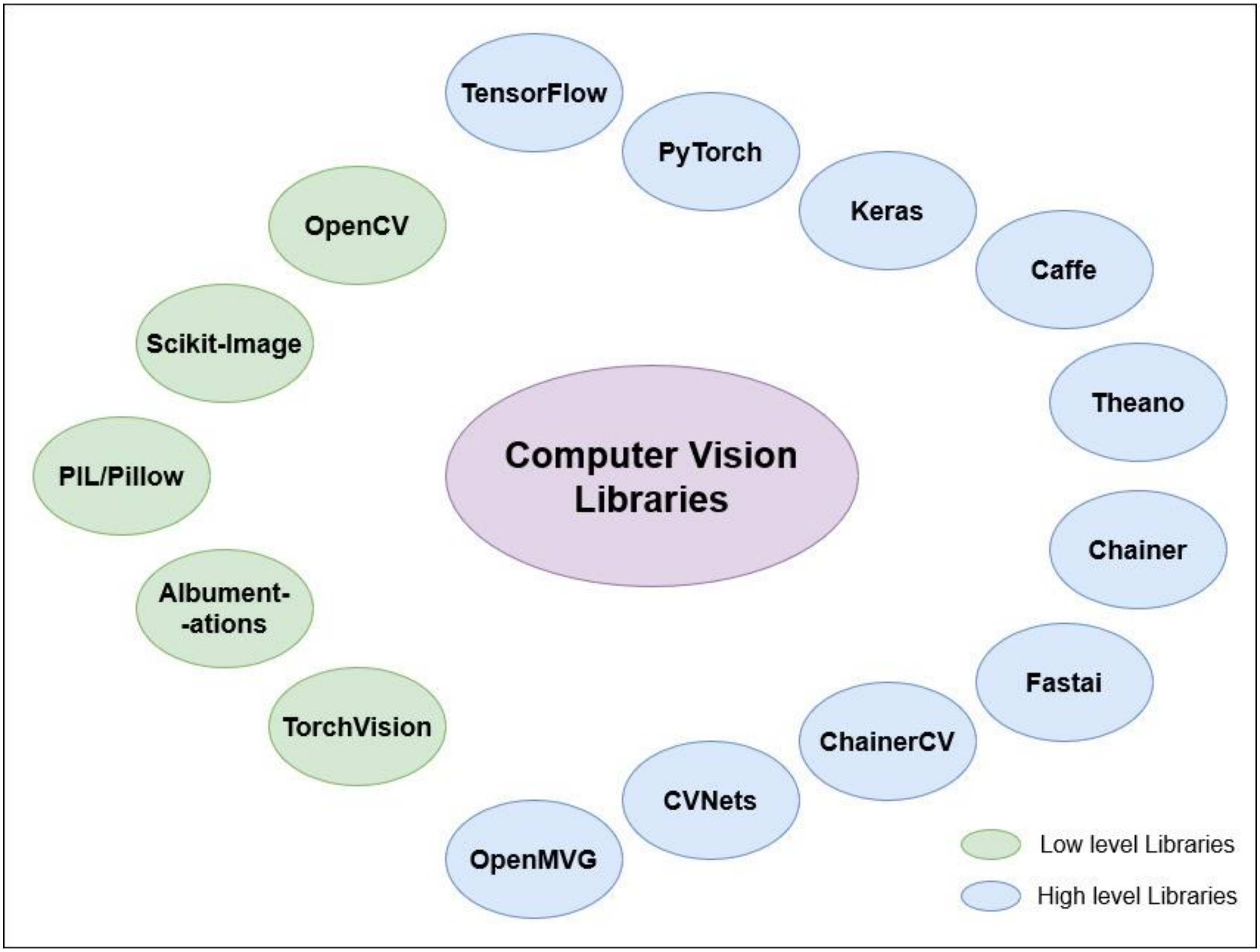

Figure 13: High-level and low-level classification of computer vision libraries

### *3.3.1. Low level libraries*

Libraries offering utilities that deal with image at the primary level are considered low-level libraries. These provide foundational tools and algorithms that enabling more granular control over the processing of visual data. These libraries offer functions for basic image manipulation, pixel-level operations, and mathematical transformations also, along with other operations. They serve as the building blocks for more customized and specialized applications. Some popular low-level libraries are listed below.

(i) OpenCV [187]: It is a BSD licensed library launched in 1999 and first stable release in 2006 by Intel, offering a wide range of functionalities, it equips developers with a comprehensive set of utilities for tasks including image and video processing, feature extraction, object detection, object tracking, and machine learning. Its modular structure provides a flexible framework for diverse applications, encompassing modules such as imgproc and ximgproc for image processing, video for video analysis, videostab for video stabilization, and objdetect and xobjdetect for object detection [188]. Supporting multiple programming languages such as C, C++, Java, and Python, it is versatile, making it suitable for a broad spectrum of use cases in the field of image and video analysis of aerial data.

(ii) Scikit-Image [189]: It is a python library released in 2009, under the permissive modified BSD open-source license, specializes in image processing and computer vision tasks. Within its toolkit, a rich set of utilities designed for tasks like image segmentation, feature extraction, and image filtering are present. This library empowers developers and researchers to efficiently manipulate and analyze images, making it a valuable asset for a wide range of applications in the realms of computer vision and image processing. Its collaborative and innovative nature is fostered by the characteristics like open source, allowing developers to work in concurrency.

(iii) PIL/Pillow [190]: It is a Python library for image manipulation or image enhancement operations such as image inversion, image cropping, image blurring, image filtering etc. It was released in 1995 as an open-source utility and is written in Python and C languages.

(iv) Albumentations [191]: It is a Python library dedicated to image augmentation, offering an open-source solution for enhancing dataset diversity through the application of diverse transformations to existing images. It offers a broad range of transformations like rotation, flip, adjustments in brightness and contrast, scaling, and many other geometric transformations. Albumentations is versatile, efficient, and user-friendly, making it a valuable tool to boost the variability of image datasets.

(v) TorchVision [192]: It is an open-source library available under modified BSD license, written in C++ & Python, and is associated with PyTorch. It offers a variety of image transformations that can be applied during data preprocessing and augmentation. These transformations help to improve the generalization and robustness of the models.

#### *3.3.2. High level libraries*

Libraries that abstract away some of the intricacies of low-level operations and provide more user-friendly interfaces are considered high level libraries. These libraries encapsulate complex algorithms and functionalities into easy-to-use modules, enabling implementation of advanced vision algorithms without delving into the details of the underlying algorithms. High-level libraries are valuable for rapid prototyping, application development, and scenarios where ease of use is a priority. Some popularly used high level libraries are as follows:

(i) TensorFlow [193]: TensorFlow is an open-source machine learning library introduced by Google in November 2015, serves as a versatile platform for creating, training, and deploying machine learning models. Its robust functionality encompasses a high-level neural networks API named Keras, which facilitates streamlined model development. This powerful tool is particularly capable of handling diverse tasks, ranging from image classification to object detection and segmentation. Embraced widely in the field of DL, TensorFlow boasts compatibility with multiple programming languages such as Python, C++, and JavaScript, making it accessible and applicable across a broad spectrum of development environments. It is mainly a product-oriented framework as it adopts a static computational graph approach.

(ii) PyTorch [194]: It is an open-source DL framework developed by the FAIR laboratory in 2016, distinguishes itself with its dynamic computational graph paradigm. In this approach, the computational graph is constructed dynamically during program execution, providing a heightened level of adaptability to varying input sizes and offering a more intuitive debugging experience. Designed with a research-oriented focus, it stands out for its flexibility, user-friendly interface, and dynamic characteristics, making it particularly well-suited for research prototyping.

(iii) Keras [195]: It is an open-source DL library released in 2015, providing a high-level interface for building and training neural networks. It is written in Python language and possesses features such as modularity, extensibility and user-friendliness. It follows a modular design, allowing users to build neural networks layer by layer. This modular approach makes it easy to construct and customize complex network architectures. It provides support for other DL libraries such as TensorFlow, Theano and MS CNTK etc.

(iv) Caffe [196]: It is a BSD-licensed library developed in C++. It can bind with Python and MATLAB to build CNNs and other DL models. It can use GPU computations with the help of CUDA. It includes a model Zoo containing pre-trained models for various tasks, making it convenient to leverage existing models. It was introduced in 2014 and its first stable version was released in 2017.

(v) Theano [197]: Theano is an open-source numerical computation library for Python that was developed by the MILA at the University of Montreal. It was designed to optimize and perform efficient numerical computations, for DL models.

(vi) Chainer [198]: It is an open-source DL framework that supports Python language. It was developed by Preferred Networks, released in 2015, and is known for its flexibility and dynamic computational graph construction. It is based on the paradigm Define-by-Run instead of conventional Define-and-Run used in case of static computation graphs; thus, allowing the construction of dynamic computation graphs on-the-fly, during runtime.

(vii) Fastai [199]: It is an open-source DL library available under the Apache 2 license, released in 2018. It provides access to high-level components for achieving state-of-the-art results in conventional DL applications with speed and simplicity and a suite of low-level components, allowing for the flexible combination of elements to create innovative approaches.

(viii) ChainerCV [200]: It is an open-source library built on top of Chainer [198]. It aims to provide a variety of pre-built tools and functionalities for vision related tasks. It is an extension of the Chainer DL framework that allows for dynamic computation graphs, particularly useful in scenarios where the network architecture needs to change dynamically. It focuses on image classification, object detection and image segmentation tasks.

(ix) CVNets [201]: It is an open-source library, which aids in development of high-performance models in relevance to various vision tasks. It provides independent plug-and-play components that can be used for different visual recognition tasks including classification, detection and segmentation.

(x) OpenMVG [202]: It is an open-source library and software suite, designed for vision applications, particularly in the context of multiple view geometry. The goal of OpenMVG is to provide tools and algorithms for the reconstruction of 3D scenes from multiple images, camera calibration, and SfM. SfM involves estimating the 3D structure of a scene and camera poses from a set of 2D images.

Various other libraries for some specific tasks include DeepFace [203], Face.evoLVe [204] and Lightface [205] for face analytics. PyTorch based Detectron2 [206] and YOLO [130] for object detection while, TASK2VEC [207] for visual classification tasks and Fedcv [208] for FL for various tasks.

This classification into low-level and high-level libraries serves to guide developers and researchers in choosing frameworks that align with their expertise, project requirements, and development goals. Researchers may require fine-grained control over image processing, algorithms, and low-level operations, so they can opt for low-level libraries where specific algorithms can be fine-tuned and experimentations can be performed. For developers, focusing on rapid development and prototyping without delving into intricate details, high-level libraries offer convenient abstractions, saving time and effort.

Table 6: Comparison of various frameworks and libraries

| S. No. | Library/ Framework | License | Initial Release Year | Developed by | Languages | OS/Platform | Dynamic Graphs | Application area |
|---|---|---|---|---|---|---|---|---|
| 1. | OpenCV [187] | BSD, Apache 2 | 2006 | Intel | C, C++, Assembly Language, Java, Python | Cross Platform | No | Image and Video Processing, Computer Vision |
| 2. | Scikit-Image [189] | BSD | 2009 | Stefan van der Walt | Python, Cython, C | Linux, Mac OS X, MS Windows | No | Image Processing |
| 3. | PIL (Pillow) [190] | HPND | 1995 | Secret Labs AB | C, Python | Linux, Mac OS X, MS Windows | No | Image Processing |
| 4. | Albumentations [191] | Apache 2 | 2018 | Kaggle | Python; built on top of OpenCV | Linux, Mac OS, MS Windows | No | Data (Image/Video) Augmentation |
| 5. | Torchvision [192] | Modified BSD | 2016 | FAIR | Python, C++ | Linux, Mac OS, MS Windows | Yes | Image and Video transformations |
| 6. | TensorFlow [193] | Apache 2 | 2015 | Google Brain Team | Python, C++, CUDA | Linux, Mac OS, MS Windows, Android | Yes | ML Library |
| 7. | Keras [195] | Apache 2 | 2015 | François Chollet | Python | Cross-platform | No | Frontend for TensorFlow |
| 8. | Caffe [196] | BSD | 2017 | Yangqing Jia | C++ | Linux, Mac OS, MS Windows | No | DL Library |
| 9. | Theano [197] | BSD-3 | 2007 | MILA | Python, CUDA | Linux, Mac OS, MS Windows | No | ML Library |
| 10. | PyTorch [194] | BSD-3 | 2016 | Meta AI | Python, C++, CUDA | Linux, Mac OS, MS Windows | Yes | ML/DL Library |
| 11. | Chainer [198] | MIT | 2015 | Preferred Networks in partnership with IBM, Intel, Microsoft and Nvidia | Python | Cross-platform | Yes | DL Library |
| 12. | Fastai [199] | Apache 2 | 2018 | fast.ai | Python | Linux, Mac OS, MS Windows | Yes | DL Library, GPU-optimized Computer Vision Library |
| 13. | ChainerCV [200] | MIT | 2017 | Preferred Networks | Python | Cross-platform | Yes | Computer Vision Library |
| 14. | OpenMVG [202] | MPL2 | 2013 | CVGLab at EPFL | C++ | Linux, Mac OS, MS Windows | No | MVG and SfM Library |

### 3.5. Hyper parameters

Hyper parameters serve as top-level configuration settings, established prior to the commencement of the training process. While training a DL model, the model is initialized with some weights that are learnable model parameters, and during the learning process these weights are updated so that the error or loss between the model's predictions and the actual values is minimized [209]. It can be considered as an optimization problem with loss or error quantified using a loss function with an objective of minimizing it. An optimizer is an algorithm, tasked with minimizing the predefined objective function. The selection of an optimizer is an integral part of the hyper parameter tuning process as it affects the model’s convergence speed and its likelihood of getting stuck in local minima [210]. Also, different optimizers have different computational and memory requirements. Hyper parameters are non-learnable and play a pivotal role in governing various aspects of the training dynamics. Factors such as model's learning behavior, convergence speed, memory utilization, and generalization capability are influenced by the chosen hyper parameter values [209]. The model may overfit or underfit if these hyper parameters are not tuned properly. The process of hyper parameter tuning is integral to crafting effective models, aiming to discover the optimal combination that leads to enhanced performance and efficient resource utilization. Hyper parameters encompass a range of settings, including but not limited to learning rate, batch size, number of epochs, dropout rate, and momentum. Out of these hyper parameters, learning rate and momentum are internal to optimizers [210]. For this reason, optimizers significantly impact the model's weight updates, influencing the pace of adaptation during training. Some hyper parameters are explained as follows.

#### *3.4.1. Optimizer or Optimization Algorithm*

An optimizer is an algorithm used to minimize an objective function associated with the error or loss between the model's predictions and the actual values [210]. Some commonly used optimizers are:

(i) Gradient descent: It adjusts model parameters iteratively to minimize loss function by computing the gradient of the entire training dataset.

(ii) Gradient descent with momentum: It enhances the gradient descent optimizer by incorporating momentum, use of past gradients for the current update, accelerating the model’s convergence.

(iii) SGD: It focuses on adjusting the model parameters by using a batch of the training data in each iteration. It accelerates the model’s training.

(iv) RMSProp: It is an adaptive optimizer that adjusts learning rates for each parameter based on the magnitude of their historical gradients.

(v) Adam: It is an optimizer that combines gradient descent with momentum and RMSProp, utilizing both influence of past gradients and adaptive learning rates.

(vi) Adadelta: It is an extension of RMSProp addressing its learning rate decay problem by dynamically adapting learning rates during training.

#### *3.4.2. Learning rate*

It is a hyper parameter that is specific to the optimizer selected, as it is a part of the optimizer algorithm's design. It determines the step size taken during optimization process based on the gradient values. A higher learning rate potentially speeds up the convergence, but a value too high may overshoot the minimum. A low value may cause the optimizer to get stuck in local minima. Different optimizers handle learning rates differently; for example, adaptive optimizers can adjust learning rates for each parameter individually [116].

#### *3.4.3. Learning rate decay*

This hyper parameter specifies the rate at which learning rate is reduced over time during the training process. It is used to avoid the problem of overshooting the minimum, as initially the model uses higher learning rate so that it converges faster and as the training progresses, the model is fine-tuned using lower learning rate [211].

#### *3.4.4. Momentum*

It is used with optimizers like gradient descent with momentum, SGD and its variants [212]. These optimizers use the momentum principle in optimization that involves adding a fraction of previous gradient to the current gradient at each iteration. The momentum hyper parameter controls the influence of past gradients on the current update. A higher momentum value allows the optimization process to remember and build up more information from past gradients, smoothing out the update trajectory.

#### *3.4.5. Batch size*

The batch size specifies the number of training examples utilized in each iteration to update the model's weights. After processing each batch, the model undergoes weight adjustments. A larger batch size accelerates training but demands more memory, while a smaller batch size mitigates the risk of getting stuck in local minima. Choosing an optimal batch size involves considering factors such as the training dataset's size, model complexity, and available computational resources [117].

#### *3.4.6. Number of epochs*

An epoch denotes a full pass through the entire training dataset in the training process. The number of epochs specifies the number of times the model processes the entire dataset, aiming to learn the underlying patterns and relationships. Too few epochs may lead to underfitting, while excessive epochs can result in overfitting. The optimal number of epochs depends on factors such as the chosen optimizer, learning rate, and other hyper parameters. The convergence speed of certain optimizers may necessitate fewer epochs, while others might require more. Implementing early stopping, based on validation performance, is a common practice to prevent overfitting after a specific number of epochs [117].

### *3.4.7. Dropout rate*

Dropout is a regularization technique where randomly selected neurons are ignored or dropped out during training. The dropout rate specifies the probability that a neuron will be dropped out or set to zero during an iteration. The dropout rate is used to reduce overfitting during training. A higher dropout rate increases the regularization effect, making the model more robust but potentially slowing down the training process, whereas, a lower dropout rate may not provide sufficient regularization [213].

All the hyper parameters are external to the model and are usually tuned through experimentation to find the best configuration for a given task.

## 4. Applications and Case Studies

Aerial data analysis within computer vision offers a multitude of applications, transforming decision-making and resource optimization across a broad spectrum of domains. These domains encompass environmental monitoring, agriculture, urban planning, disaster response, defense, and security, among others [2], [3]. Some major applications are as shown in Figure 14. As we discuss the capabilities of aerial data analysis into the multifaceted applications, it will be evident that this technology is a cornerstone in addressing contemporary challenges and steering industries toward more informed, sustainable, and resilient practices [36]. This section will explore these applications in greater detail, highlighting the transformative impact of aerial data analysis across diverse domains. Some of the important case studies in aerial data analysis for various application scenarios have also been discussed here.

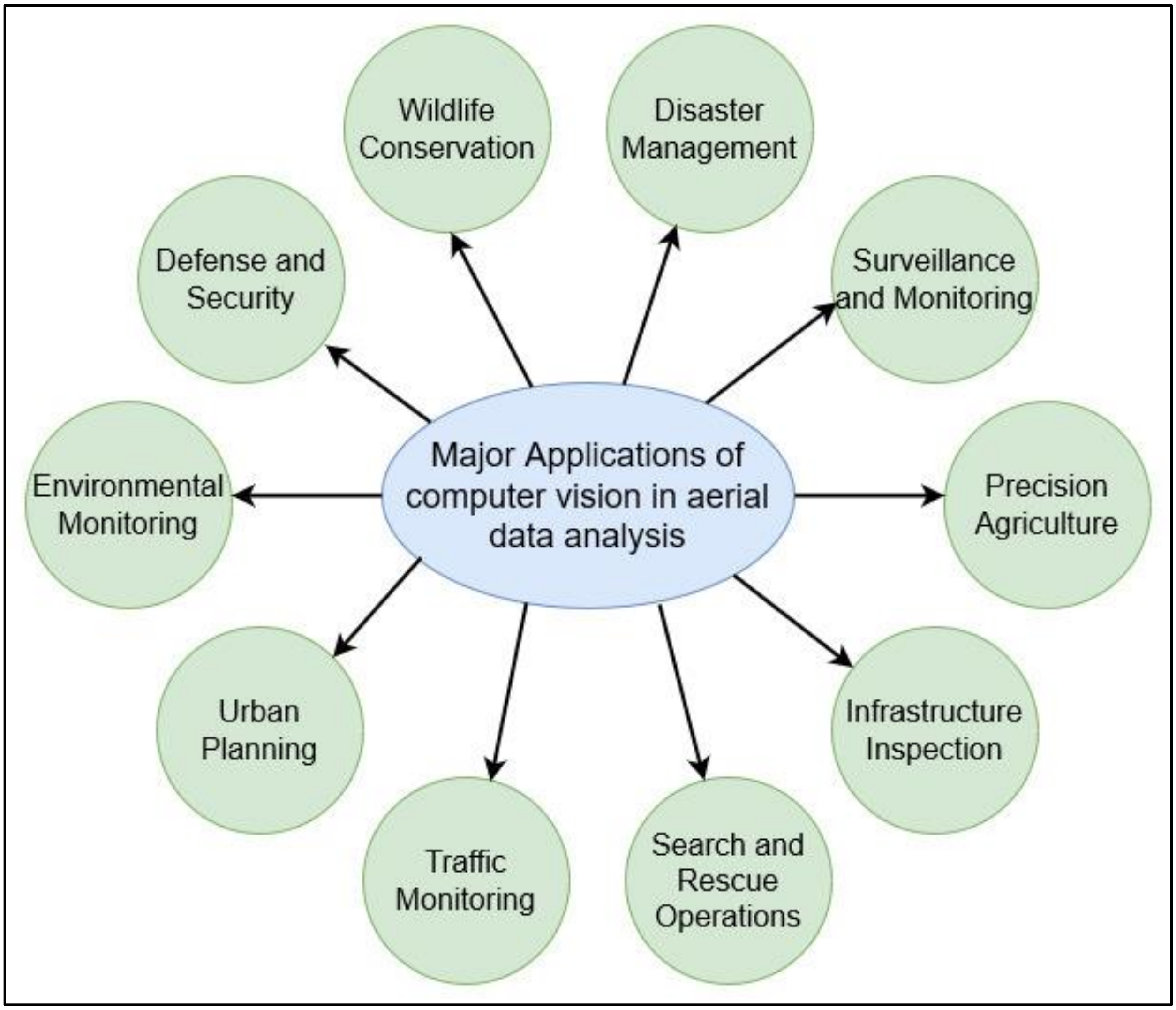

Figure 14: Major applications of computer vision in aerial data analysis

### 4.1. Disaster Management

Disaster management encompasses a multifaceted approach aimed at mitigating the impact of natural or man-made calamities and efficiently coordinating resources for recovery efforts. In this context, the significance of analyzing aerial data through vision technologies becomes paramount [88]. Aerial data offers a comprehensive and dynamic perspective of disaster-stricken areas, allowing for rapid and accurate assessment, as shown in Figure 15 (a), where a drone is used for post disaster monitoring of an area. The role of computer vision in this domain is crucial, as it facilitates the extraction of meaningful insights from large-scale geographic information. Object detection and segmentation algorithms can aid in identifying damaged infrastructure, assessing the extent of destruction, and locating areas that require immediate attention [88]. Change detection algorithms can contribute by highlighting alterations in the landscape caused by the disaster [76]. Terrain modeling can help in understanding the topographical changes, which is vital for planning rescue and relief operations [214]. Additionally, object tracking can provide real-time monitoring of moving entities, aiding in the tracking of debris, vehicles, or displaced individuals during the response phase. By leveraging computer vision technologies to analyze aerial data, disaster response and management teams gain a more nuanced understanding of the situation. This, in turn, enables them to make informed decisions, allocate resources effectively, and coordinate responses with greater precision [88]. The ability to quickly and accurately assess the impact of disasters through aerial data analysis significantly enhances the overall efficiency and effectiveness of disaster response efforts.

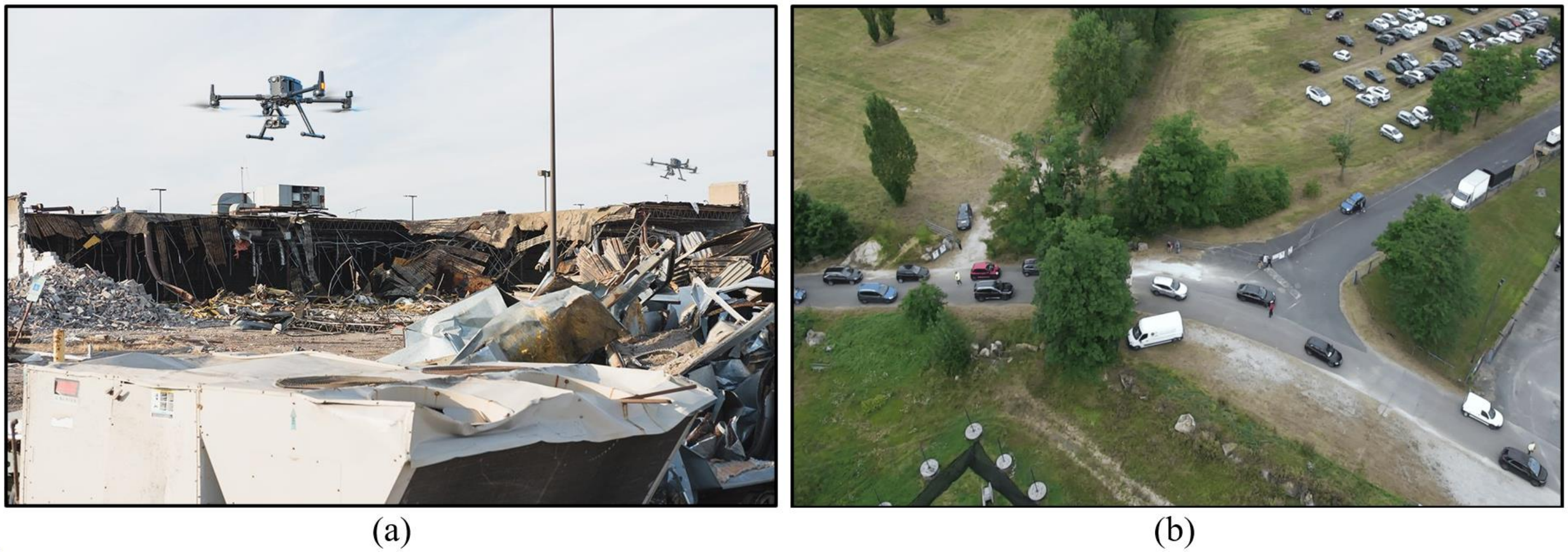

Figure 15: (a) Post disaster monitoring and analysis using a drone [215] (b) Surveillance and monitoring of an area using aerial platform [216]

In [62], the authors presented YOLO based CNN models for classification of critical ground assets including damaged and undamaged building roofs, vehicles, vegetation, debris, and flooded areas post disaster. In [55], the authors proposed a CNN based model for flood detection by means of change detection in the form of damage caused by the flood. In [164], the authors have introduced an aerial dataset named Floodnet, which can be used to understand post flood scenes by the model. The authors of [165] introduced an aerial image database named as AIDER for emergency response and proposed a CNN based solution for real time classification of disastrous events from a UAV. In [54], a CNN based model for building damage identification from post disaster aerial images was proposed using post-earthquake aerial images. Subsequently, the developed model was applied to identify damaged buildings from aerial images taken in September 2019 from Chiba, Japan post typhoon Faxai and the estimated distribution of damage aligns well with the visually interpreted counterpart. In [66], a model based on stacked CNN architecture is introduced for preliminary damage assessment post disaster trained using UAV captured videos post hurricane Dorian. A YOLO based model for detection of collapsed buildings from aerial images post-earthquake was proposed in [67].

### 4.2. Surveillance and Monitoring

Surveillance and monitoring through aerial data analysis represents critical components of modern safety and security strategies, playing a pivotal role in maintaining public safety and safeguarding critical infrastructure [82]. Aerial data offers a unique vantage point for comprehensive surveillance and monitoring. The significance lies in the ability to gather real-time and high-resolution visual information, enabling timely detection, tracking, and response to potential threats or incidents. This could include tracking the movement of people, vehicles, or objects, and detecting anomalies that may indicate security risks [83]. In Figure 15 (b), vehicular movement in an area is being monitored using an aerial platform.

In the context of critical infrastructure protection, such as airports, borders, and key facilities, aerial surveillance provides an added layer of security. The technology enables the detection of unauthorized intrusions, unusual patterns, or potential security breaches etc. [81]. Rapid response to such events becomes possible through the timely analysis of aerial data, contributing to overall safety and security. The analysis of aerial data for surveillance and monitoring purposes offers a comprehensive and efficient means of ensuring public safety, protecting critical infrastructure, and responding effectively to security threats [79]. In [72], a comprehensive drone surveillance system is devised, comprising detection, tracking, and classification modules. The detection module identifies flying objects, the tracking module monitors them, and the classification module distinguishes between drones and other potential distractions. In [217], a systematic review investigates the integration of surveillance drones in smart cities, encompassing the current application status, diverse application areas, proposed models, and distinctive characteristics within the smart city context. In [218], a thorough survey explores the use of networked UAVs for surveillance and monitoring, addressing associated challenges. In [63], the authors presented a real-time drone surveillance system that identifies violent individuals by employing pose estimation of humans in aerial data. The system leverages the SHDL network and introduces a novel dataset specifically designed for detecting violent individuals in aerial scenarios. In [219], a concise survey delves into UAV applications in video surveillance, shedding light on associated challenges. In [220], a novel approach for real-time aerial surveillance, incorporating detection and recognition of multiple moving objects, is proposed. In [221], a DL pipeline is introduced for gait recognition, incorporating SSD, Inception-V3-based CNN, and LSTM models, to detect suspicious actions and abnormal activities of individuals.

### 4.3. Precision Agriculture

Precision agriculture, leveraging aerial data analysis, represents a transformative approach to farming that optimizes resource utilization, enhances crop yield, and minimizes environmental impact [95]. In precision agriculture, aerial data, often obtained through UAVs, offers a detailed and dynamic view of crop health, soil conditions, and overall field performance. Analyzing this data allows farmers to identify variations within their fields, enabling targeted interventions to address specific needs. The real-time or near-real-time nature of aerial data provides farmers with timely information to make accurate and responsive decisions [105]. The role of analyzing aerial data in precision agriculture includes monitoring crop health, assessing the effectiveness of irrigation practices, and identifying areas with nutrient deficiencies or pest infestations [95]. Aerial data also facilitates the creation of prescription maps for variable-rate application of inputs, ensuring that resources like water, fertilizers, and pesticides are applied precisely. For example, in Figure 16 (a), drones are being utilized for spraying pesticides optimally, in a field. The significance of precision agriculture extends beyond immediate economic benefits for farmers. It contributes to sustainability by minimizing environmental impact through targeted resource application. Reduced use of fertilizers and pesticides, coupled with optimized irrigation, leads to more efficient and eco-friendly farming practices [95]. Additionally, precision agriculture supports long-term soil health and biodiversity conservation. By harnessing the power of aerial imagery and advanced analytics, farmers can make informed decisions that enhance productivity, sustainability, and overall agricultural resilience.

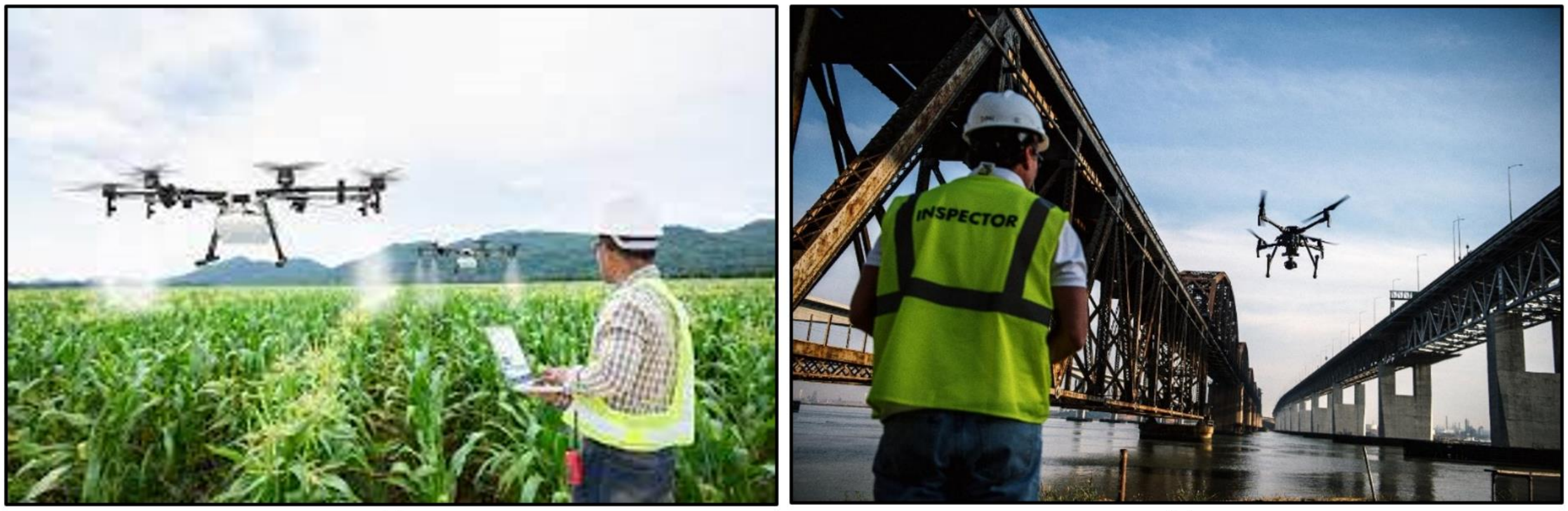

(a) (b)

Figure 16: (a) Pesticide spray in a field using drones [222] (b) Bridge inspection using a drone [229]

Significant studies have been carried out in the domain of agricultural applications using aerial data analysis. In [223], researchers proposed a CNN based classification model for automated weed detection in soybean fields using UAV imagery. The authors in [166], introduced the Agriculture-Vision dataset, a comprehensive collection of high-resolution visible and NIR images, facilitating agricultural semantic segmentation for tasks such as identifying double plants, storm damages, and weed clusters. In [224], researchers harnessed UAV-captured images to enhance the accuracy of remotely sensed irrigated areas. In [225], the authors developed a ML-based method to estimate the plant density of emerging wheat crops, predicting crop yields using UAV images. Additionally, in [226], researchers proposed fine-tuning DL strategies, including Inception-v3, Resnet-50, VGG-16, VGG-19, and Xception, for detecting and classifying pests in soybean plantations, thereby assisting in pest control management. The authors in [227], offers a comprehensive review of UAV or drone technologies, charting their evolution and application in agriculture for monitoring and control purposes. In [228], the authors conducted a comparative study between DL-based algorithms and OBIA algorithms for detection of individual cabbage plants and found that DL-based Mask R-CNN method performs better than the OBIA-MDTWS model for the same.

## 4.4. Infrastructure Inspection

Infrastructure inspection, encompassing critical assets such as bridges, power lines, pipelines, and transportation networks, is essential for maintaining public safety, preventing potential failures, and ensuring the longevity of vital structures. The meaning of infrastructure inspection lies in the systematic examination of physical assets to assess their condition, identify local and global defects, and prioritize maintenance or repair activities [89]. Regular inspections play a crucial role in mitigating these risks, avoiding catastrophic incidents, and optimizing the lifecycle management of infrastructure assets. Aerial platforms capture high-resolution imagery providing a holistic view of infrastructure assets and its analysis enhances the efficiency and accuracy of inspections, allowing for the early detection of structural issues, wear and tear, or potential safety hazards [79]. Moreover, the real-time monitoring capabilities of aerial data contribute to rapid response during emergencies or unforeseen events, as shown in Figure 16 (b), where two bridges are being monitored to assess their health. It empowers asset managers, engineers, and decision-makers with timely and accurate information, fostering a proactive approach to maintenance, reducing risks, and ensuring the resilience and safety of critical infrastructure networks.

There are several works in literature where the analysis of aerial data is used in infrastructure inspection. In [230], researchers presented a comprehensive review of the recent advancements in computer vision techniques concerning the assessment of civil infrastructure condition; delving into the latest research focused on leveraging vision-based methods for evaluating the state of civil infrastructure. In [231], the authors proposed a streamlined system for defects identification during bridge inspection. The system focuses on identification of the defect type, its extent, growth and its 3D localization using different techniques, thus, aiding the bridge inspection process and its health monitoring. In [232], the researchers proposed a vision-based plan for navigation of UAVs with an objective of collecting images for post-earthquake inspection of railway viaducts. The method focuses on identifying and localizing critical structural components for inspection. In [233], the researchers utilized UAV equipped with an optical camera for the inspection and assessment of diverse assets within an airport environment. The challenges involved in conducting 360-degree inspections of bridges using UAVs are documented in the study [234] and the best practices for the same are also outlined. In [235] and [236], different object detection and segmentation models are proposed for addressing the problems related to identification of railway track sleepers or the rectangular support structures laid perpendicular to the rails on a railway track, by using UAV images taken at low altitudes. The authors in [237] presented a vision-based algorithm for detecting railway tracks from drone imagery.

### 4.5. Search and Rescue Operations

The primary goal of search and rescue operations is to save lives and provide assistance to those in need during emergencies or critical situations. Its significance lies in its potential to mitigate the impact of disasters, accidents, or unforeseen events by deploying resources efficiently and effectively [163]. The role of analyzing aerial data in search and rescue involves leveraging high-resolution imagery, to detect and locate individuals in distress. Aerial platforms can cover expansive terrains, identify potential hazards, and provide real-time situational awareness to ground teams. As shown in Figure 17 (a), the people stuck in a mountain region are located and rescued with the help of a drone. This aids in optimizing the deployment of resources, guiding rescue teams to specific locations, and ensuring a swift and targeted response [238]. Furthermore, aerial data analysis facilitates the identification of optimal routes for rescue missions, especially in challenging terrains such as mountains, forests, or disaster-stricken areas. The ability to assess the extent of damage, identify potential hazards, and locate survivors efficiently is crucial for the success of search and rescue operations.

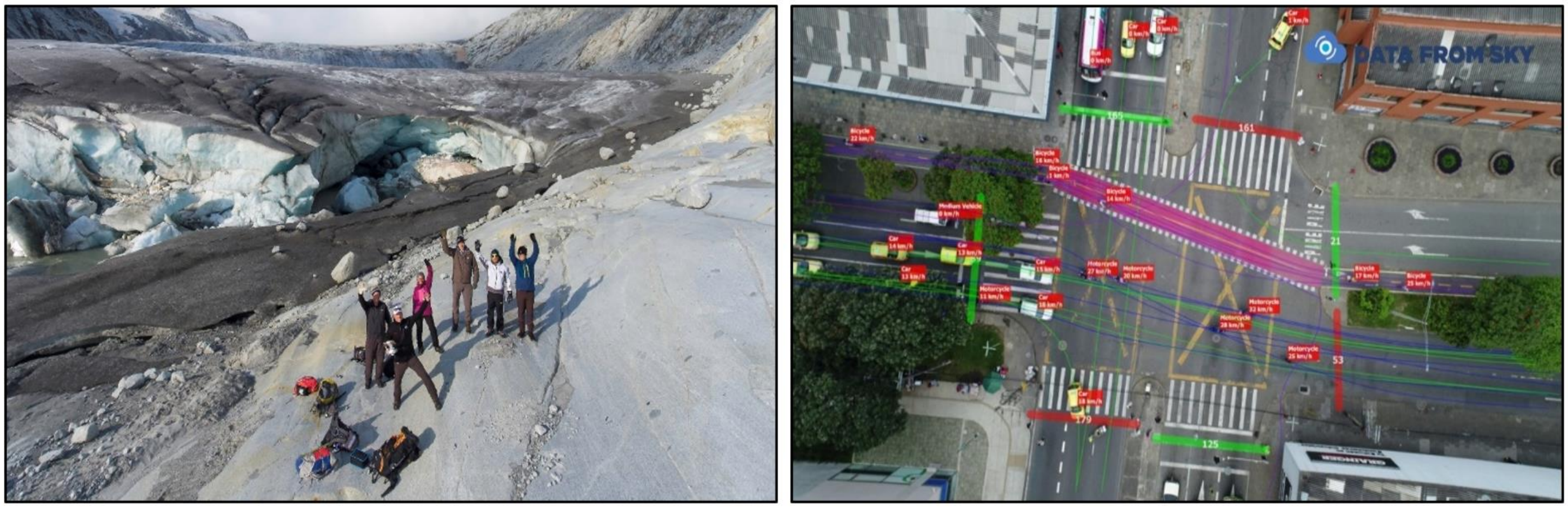

(a) (b)

Figure 17: (a) Search and rescue operations in a mountain region using drone [239] (b) Traffic monitoring at a busy crossroad [246]

The use of aerial data in search and rescue operations significantly enhances the effectiveness and efficiency of response efforts. By providing a comprehensive and timely understanding of the operational environment, aerial data analysis contributes to saving lives, reducing response times, and improving the overall success rate of search and rescue missions. In [162], a novel approach for person detection in UAV images is presented, specifically tailored for search and rescue operations in Mediterranean and Sub-Mediterranean regions. The method leverages pre-trained and fine-tuned CNNs, combined with visual attention algorithms, to enhance the accuracy and efficiency of detection. A dataset namely HERIDAL for the purpose of search and rescue operations is also introduced in this study. In [240], the authors introduced a dataset of UAV images, comprising 2,000 images and 30,000 human instances, designed for human action detection in search and rescue scenarios. Additionally, they proposed a model employing SSD for both human detection and human action recognition. Researchers in [163], presented a DL-based model tailored for search and rescue operations, leveraging the VisDrone benchmark and a custom dataset namely SARD to simulate realistic rescue scenarios. In [241] also, the authors introduced a CNN based detection model designed to identify and locate individuals in distress from aerial data, specifically for search and rescue operations. In [242], the authors presented a YOLOv5 based detection model for detecting persons and casualties in search and rescue operations using

HERIDAL and SARD datasets. A CNN based model using EfficientDet architecture and ensemble learning, for detecting humans in aerial images during search and rescue operations is introduced in [243]. The authors of [244], proposed a CNN based method for supporting the search and rescue operations in case of an avalanche.

### 4.6. Traffic Management

Traffic management involves the systematic control and organization of vehicular and pedestrian movements on roadways to ensure safe, efficient, and smooth transportation. Effective traffic management contributes to improved mobility, reduced travel times, and minimized environmental impact [115]. Aerial platforms provide valuable insights into traffic patterns, congestion points, and overall transportation dynamics [245]. This aerial perspective allows for a comprehensive understanding of the road network and traffic flow, which is instrumental in making informed decisions to optimize traffic management [80]. The role of analyzing aerial data in traffic management involves real-time monitoring and analysis of road conditions. Aerial platforms can capture high-resolution imagery, detect traffic congestion, and identify areas prone to bottlenecks or accidents [245]. As shown in Figure 17 (b), traffic flow at a busy crossroad is monitored with the help of an aerial platform. This information enables traffic management authorities to implement timely interventions, such as adjusting signal timings, rerouting traffic, or deploying emergency services. By assessing traffic patterns over time, urban planners can make data-driven decisions to design and expand roadways, enhance public transportation systems, and implement intelligent transportation solutions. This proactive approach helps in preventing future congestion and improving overall urban mobility [245]. By providing a bird's-eye view of traffic dynamics, analyzing aerial data empowers authorities to implement responsive strategies, reduce congestion, and create sustainable and well-managed urban transportation networks.

The pNEUMA experiment, discussed in [247], presents a large-scale field experiment employing a swarm of 10 UAVs to collect extensive data on traffic congestion in a bustling urban area. Additionally, [174] introduced AU-AIR, a versatile multi-modal aerial dataset tailored for low-altitude traffic surveillance. In [248], a real-time UAV-based traffic monitoring system is presented, aimed at detecting violations such as over speeding and wrong-lane driving. The authors in [249] contribute with DL-based algorithms for traffic monitoring and surveillance using UAV imagery, leveraging popular models like Faster-RCNN and YOLOv3 on the AU-AIR dataset. In [250], conducted a systematic literature review on UAVs' role in urban traffic monitoring, providing a comprehensive overview. The authors in [251], introduced an innovative framework for high-resolution vehicle trajectory extraction from aerial videos, enhancing the precision and automation of this critical task. In [252], the authors introduced an innovative four-stage real-time framework for estimating traffic flow parameters from UAV videos. Additionally, they have made the dataset used in the study publicly available for further use.

### 4.7. Urban Planning

Urban planning involves the systematic design and organization of urban areas to ensure sustainable development, efficient resource utilization, and creation of livable and functional spaces. Analyzing aerial data plays a crucial role in urban planning, providing valuable insights for informed decision-making and efficient city management [87]. Aerial data analysis in urban planning involves the use of high-resolution imagery captured from UAVs or other sources to understand and monitor various aspects of the urban landscape. This includes assessing land use patterns, infrastructure development, transportation networks, green spaces, and the overall spatial organization of the city, as shown in Figure 18 (a). The role of analyzing aerial data in urban planning includes the identification of changes in the urban environment over time, monitoring the expansion of infrastructure, and assessing the impact of new developments [77]. This information is crucial for making decisions that promote sustainable urban growth, reduce environmental impact, and enhance the overall well-being of urban residents. For example, analyzing aerial data can help identify areas prone to flooding, informed transportation and parking planning, and guide the allocation of resources for public services. By leveraging advanced technologies and analytical tools, urban planners can anticipate future challenges, respond to dynamic urbanization trends, and develop strategies to create inclusive, accessible, and environmentally friendly urban spaces [99]. In essence, the analysis of aerial data empowers urban planners with the insights needed to build cities that are not only aesthetically pleasing but also functional, sustainable, and responsive to the evolving needs of their inhabitants. A few studies are carried out in this regard. In [253], 3D city models for creating development plans and city maps are generated using UAV imagery. A dataset named Semantic Riverscapes is proposed in [254], which is then used for analysis of river landscapes to identify vegetations, grass, water, buildings, under-construction structures that can aid in formulating riverside development policies. In [255], an algorithm to automatically detect available parking from drone captured images of parking lots is proposed. A Hough transform segmentation-based building detection method from aerial images in dense urban settlements is proposed in [256]. Many studies using Google SVI are carried out in this domain as [257], [258], [259]. These studies can be expanded to include aerial images, leveraging their advantages in coverage area and contextual information and this extension can enhance the efficiency of the system by accelerating coverage over a larger area.

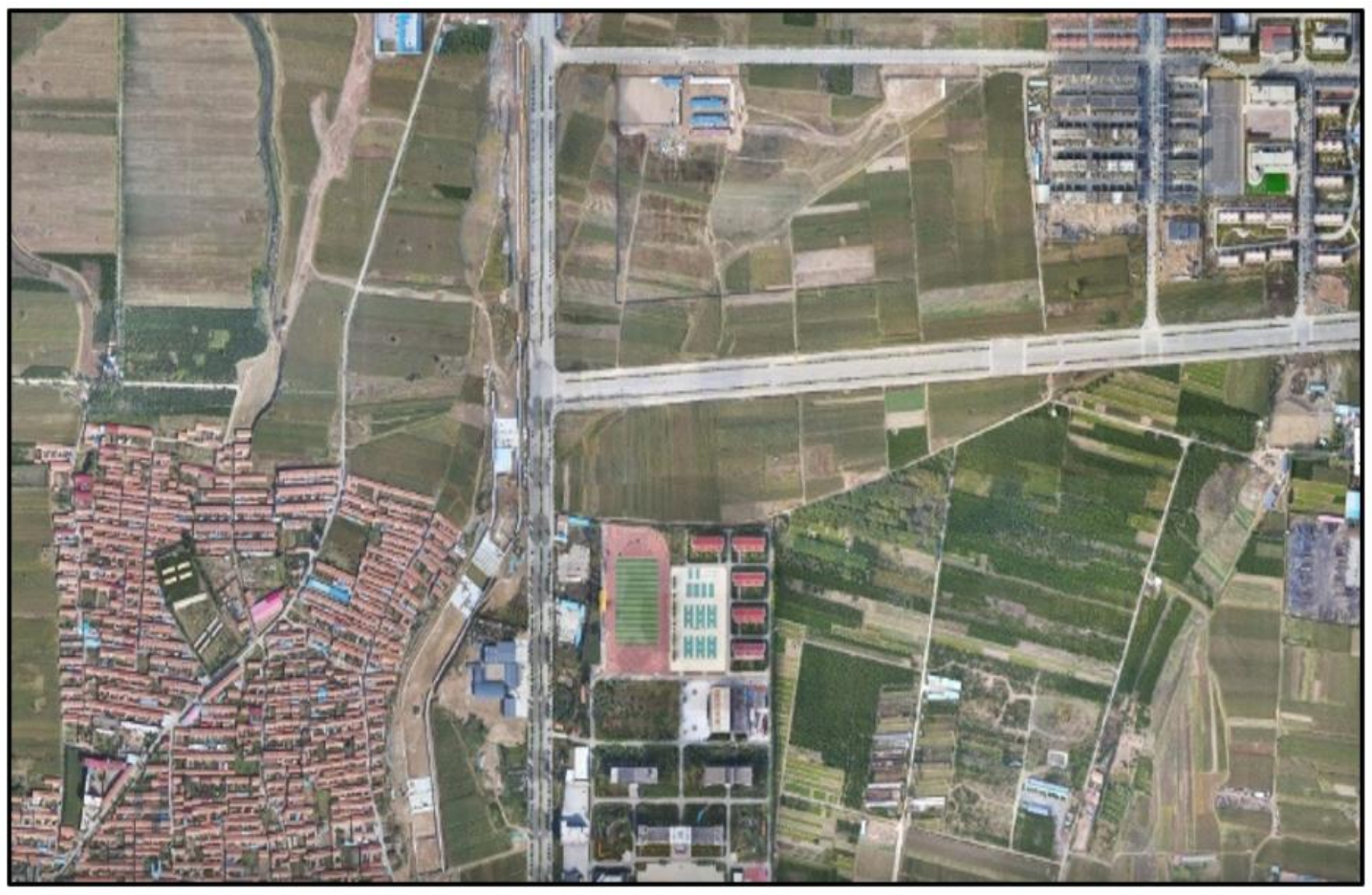
(a)

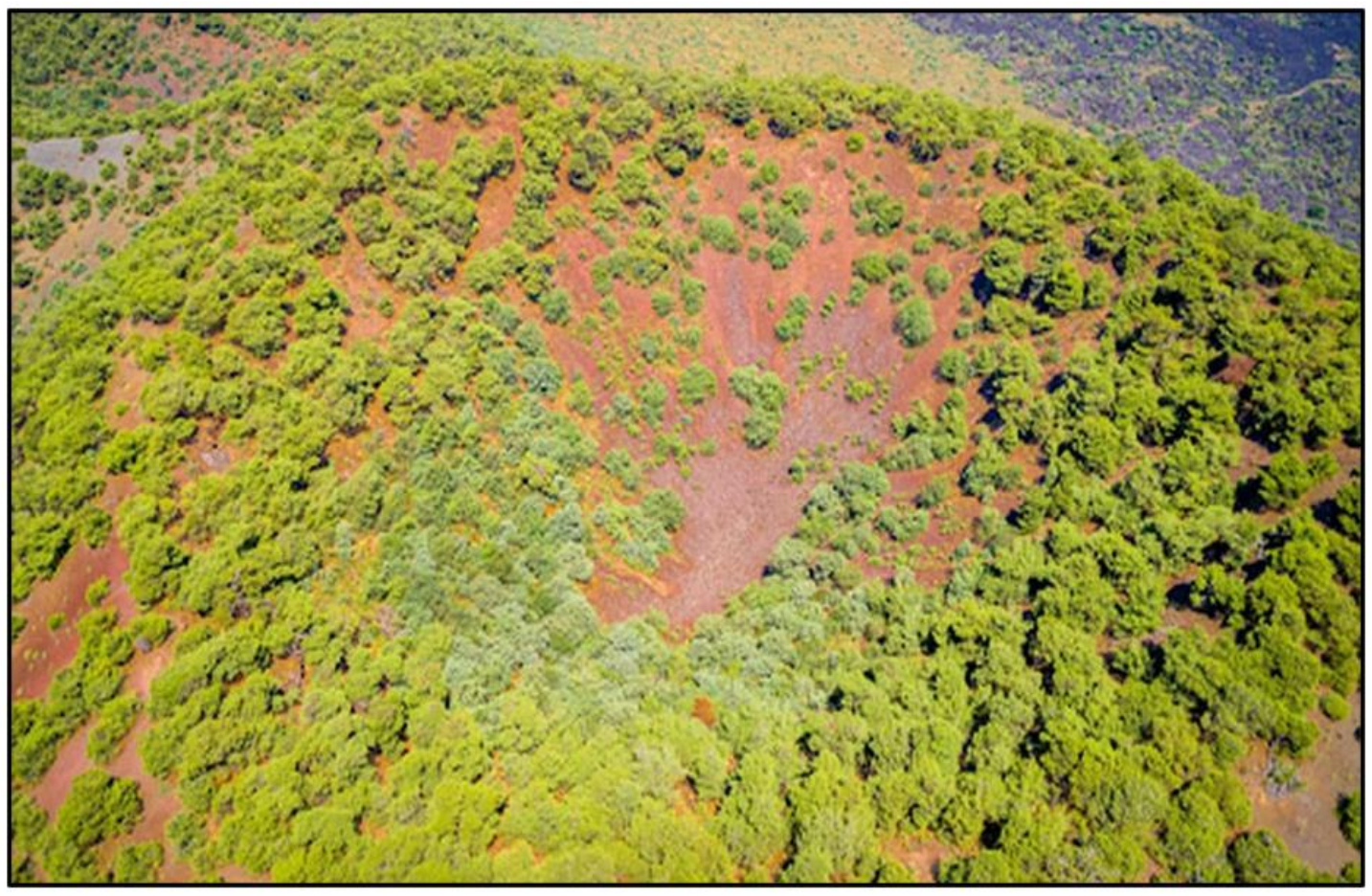
(b)

Figure 18: (a) Usage of aerial data analysis in urban planning [260] (b) Surveying of landforms using drone for environmental monitoring [264]

## 4.8. Environmental Monitoring

Environmental monitoring is a crucial aspect of assessing and managing the impact of human activities and natural processes on ecosystems, biodiversity, and the overall health of the planet, helping scientists, policymakers, and conservationists make informed decisions to protect and sustain natural resources [261]. The role of analyzing aerial data in environmental monitoring encompasses a wide range of applications, from tracking deforestation and land-use changes to monitoring wildlife habitats and assessing the impact of climate change. It allows for the identification of patterns, trends, and anomalies that might be challenging to observe from the ground [112]. For instance, the analysis of aerial data can help track the expansion or contraction of forests, and monitor changes in ice cover in polar regions. As shown in Figure 18 (b), a landform is monitored and surveyed using an aerial platform to assess the forest cover. This information is crucial for understanding the state of the environment and formulating effective conservation and management strategies. By providing real-time and historical information, it enables rapid response and decision-making to mitigate the impact of these events on ecosystems and human communities [86]. The analysis of aerial data in environmental monitoring is an invaluable tool that enhances our understanding of the Earth's dynamic systems, supporting efforts to conserve biodiversity, protect natural resources, and promote sustainable practices for the benefit of current and future generations. In [262] and [263], the authors provided a CNN based approach to detect forest fire based on aerial datasets.

## 4.9. Defense and Security

Defense and security are paramount concerns for nations worldwide, and the analysis of aerial data plays a critical role in fortifying these domains. Its significance lies in the ability to detect, monitor, and respond to security challenges effectively, ensuring the safety and sovereignty of a nation [265]. Analyzing aerial data is instrumental in strengthening defense and security efforts by providing valuable ISR capabilities. Aerial data analysis enables the identification and tracking of military movements, infrastructure development, and potential threats to national security [266]. The high spatial resolution and real-time capabilities of aerial data analysis contribute to effective border surveillance, aerial reconnaissance of sensitive areas before convoy movements, allowing authorities to detect unauthorized border crossings, monitor maritime activities, and respond promptly to security breaches [265]. As shown in Figure 19 (a), real-time border patrolling can be done with the help of a drone. It provides decision-makers with timely and accurate information to formulate strategies, respond to emerging threats, and safeguard the well-being of citizens. In [265], a CNN based real time technique is proposed for detection of small boats that are suspected to be in use for irregular border crossings. In [266], a real time DL model FDLID is presented for recognition of suspicious or enemy's flying objects using a UAV. In [267], the use of multiple UAVs along with unattended ground sensors-based alert stations for border surveillance is proposed.

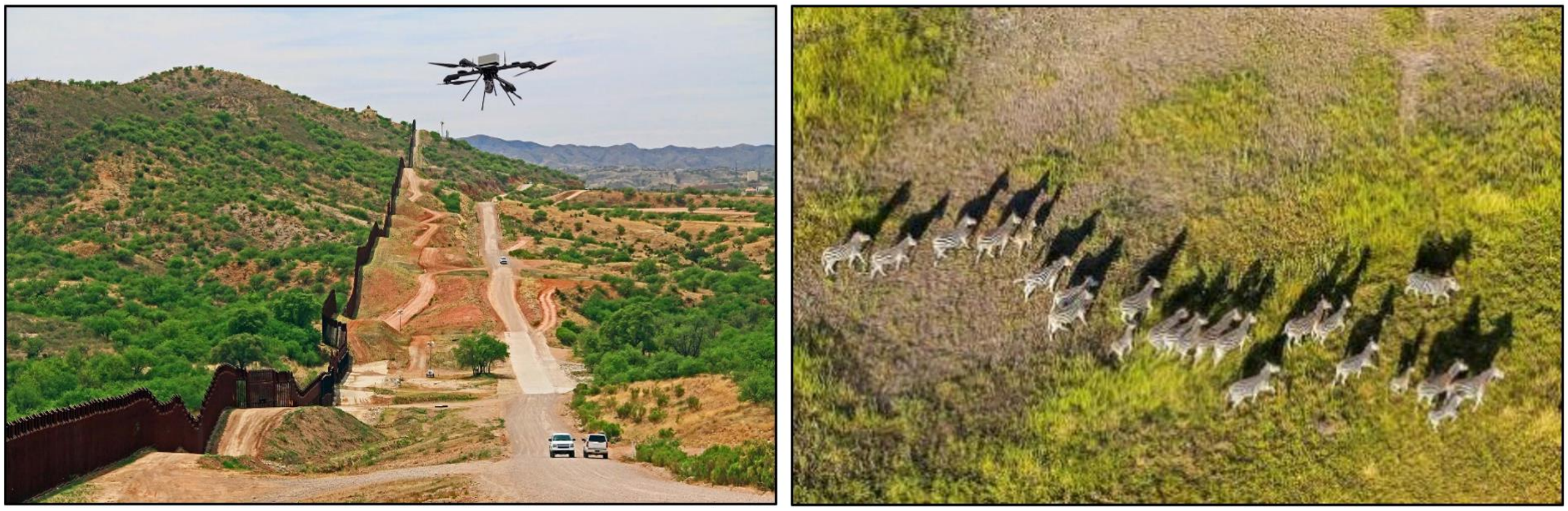

(a) (b)

Figure 19: (a) Border patrolling using a drone [268] (b) Monitoring of wildlife in a reserved forest using drone [275]

4.10. Wildlife Conservation

Wildlife conservation encompasses a comprehensive effort to safeguard and sustain the rich variety of flora and fauna present on our planet. It includes efforts to safeguard both flora and fauna, promote sustainable practices, and address the threats posed by climate change, and habitat loss. The significance of wildlife conservation extends beyond the intrinsic value of preserving biodiversity; it is integral to ensuring the health and resilience of ecosystems, which, in turn, sustains the overall well-being of the planet [64]. The ability to survey large areas efficiently and non-invasively allows to gather valuable data on wildlife populations, identification of endangered species, migration patterns, and habitat conditions [107]. As shown in Figure 19 (b), wildlife in a forest can be monitored using an aerial platform. The role of analyzing aerial data in wildlife conservation involves employing advanced technologies such as GIS, that enables the collection of spatial and temporal information, helping track changes in land cover, detect deforestation, assess the impact of climate change, and monitor the movements and behaviors of wildlife populations [269]. Furthermore, aerial data analysis facilitates the identification and documentation of endangered species, aiding in conservation planning and management strategies. By providing a bird's-eye view of ecosystems and enabling data-driven decision-making, aerial data analysis contributes to the preservation of biodiversity, the protection of endangered species, and the sustainable coexistence of wildlife and human communities. Some notable studies carried out in this domain are discussed here. In [270], the RetinaNet multi-class CNN demonstrated effectiveness in identifying elephants, giraffes, and zebras in aerial images, surpassing human annotators in certain aspects. Another approach proposed by [53], utilized UNet and EfficientNet models with density maps for counting sea lions and elephants in aerial photos. The datasets, Aerial Elephant and WAID, were introduced by [167] and [168], respectively, providing benchmarks for wildlife detection from aerial imagery. The authors in [271] conducted a population census of large wild herbivores using UAVs and computer vision techniques, demonstrating advantages in speed and cost-effectiveness. The authors in [272], employed UAV surveys to estimate the density of threatened deer species in a forested area. The authors in [273], reported increased accuracy and precision in counting replica seabird colonies with drone-based monitoring compared to traditional human methods. In [274], the authors investigated the use of CNNs for detecting African mammal species in aerial imagery. They employed three CNN-based detection algorithms namely Faster R-CNN, Libra-RCNN, and RetinaNet. The comparative analysis showed that the Libra-CNN algorithm outperformed the rest of the discussed algorithms. Subsequently, the authors applied the Libra-CNN algorithm in a case study using an independent dataset, highlighting its efficacy in the accurate detection and identification of African mammals in aerial imagery.

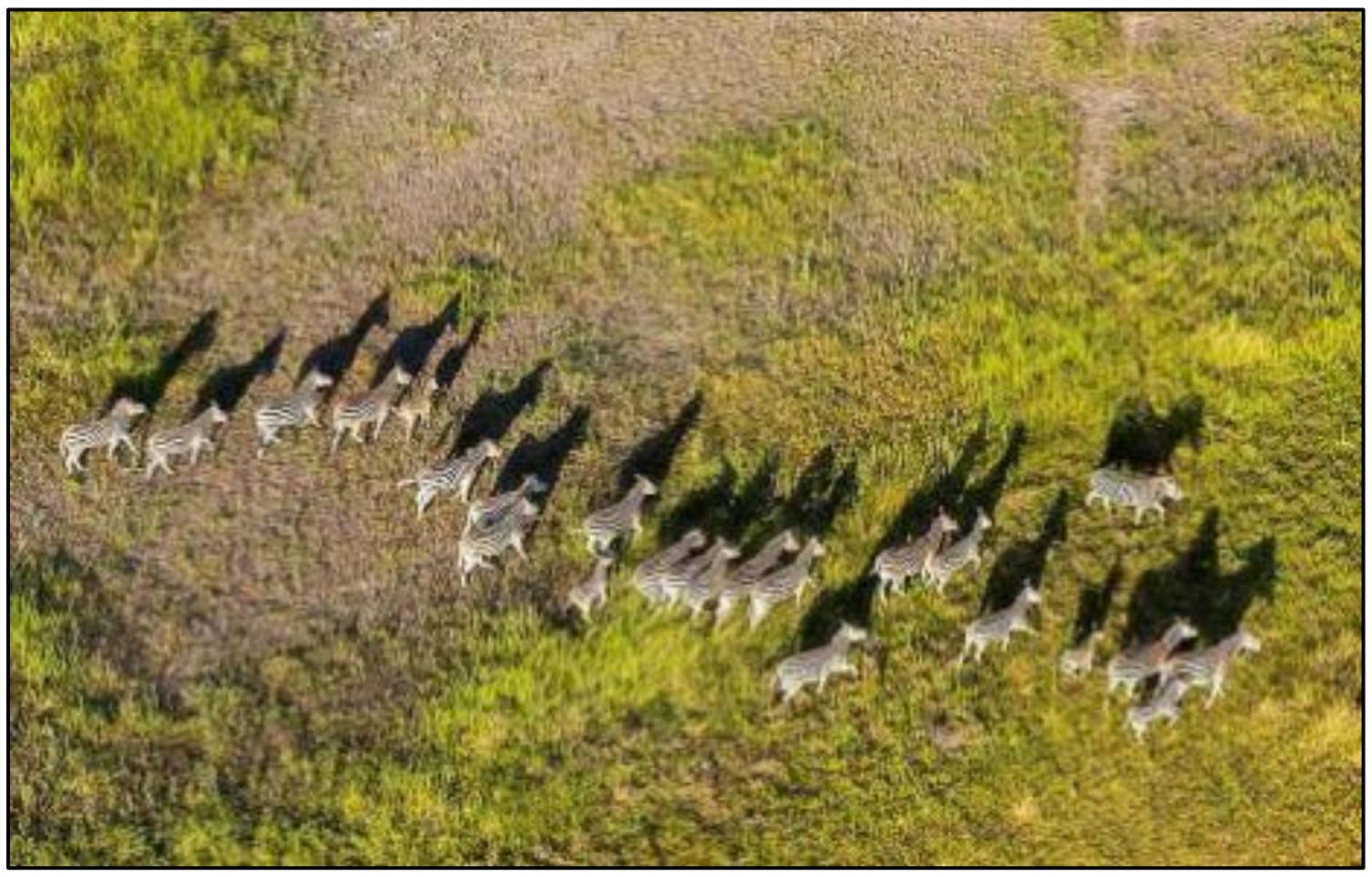

Figure 20:

Case studies conducted across diverse application domains serve as instrumental tools in demonstrating the practical value and efficacy of aerial data analysis compared to traditional methods. These case studies provide tangible examples and real-world scenarios where the application of aerial data analysis has shown significant advantages, validating its utility and impact. Case studies in various application domains have been carried out and some of the studies are listed in Table 7.

Table 7: Case studies in aerial data analysis for various application scenarios

| S. No. | Reference | Application Domain | Contribution | Study performed at | Time and Event | Experiment details | Findings |
|---|---|---|---|---|---|---|---|
| 1 | [245] | Traffic Monitoring | Analysing the traffic streams at an urban roundabout in terms of traffic flow | Sint-Truiden, Belgium | Early-evening rush hour on a Friday afternoon (15:00 to 16:30 hours) | The assessment of traffic volume involves the utilization of origin-destination matrices, specifically tracking the number of vehicles entering or exiting a roundabout. | The analysis revealed that the predominant traffic originated from the city, flowing towards shopping and residential areas.<br><br>Automated traffic counts were cross-validated with manual counts, revealing minimal discrepancies. The only discernible errors occurred in instances of vehicles partially obscured by nearby trees. |
| 2 | [261] | Environmental Monitoring | Assessing the impacts of wildfire to plan and execute the restoration work | Bugac Juniper Forest, Kiskunság National Park, Hungary | Fire: 29 Apr 2012 to 5 May 2012<br><br>Aerial survey: 7 June 2012; 1 July 2013 | The study involved categorizing the extent of damage, the condition and type of vegetation, and the presence of invasive species by examining aerial photographs.<br><br>Manual, semi-automatic and automatic approaches were compared. | On the basis of the forest maps, using manual approach 24% of the area was classified as intact, 38% as partly damaged, and 38% as totally damaged.<br><br>Using semi- automatic approach, 14% of the area was classified as intact, 28% as partly damaged, 30% as totally |
| | | | | | | | damaged and 21% as shadowed.<br><br>Using supervised classification or automatic approach, 12% of the area was classified as intact, 30% as partly damaged, 50% as totally damaged and 8% as shadowed or open sand land. |
| 3 | [99] | Urban Planning | 3D Urban modelling | Loulé Castle, District of Faro, Southern Portugal | 2016 | The generation of 3D model for urban surroundings using aerial photos. | 51 aerial photos were used in this work, each with 90% longitudinal overlapping and 40% transversal overlapping; resulting in point cloud comprising of 23,116,106 points, 2,311,901 vertices, and 4,616,659 faces. |
| 4 | [214] | Disaster Management | Developing a pre-disaster 3D map of a city using UAV imagery. | Downtown Victoria, British Columbia, Canada | 14 June 2018 morning hours | The study compared PPK-corrected drone DSMs to a lidar DSM in terms of error.<br><br>Assess the results in terms of geospatial accuracy and building representation. | The study revealed vertical errors ranging from 0.03 to 0.13 m for the slow PPK-corrected drone DSM. The average error was 0.08 m, and the rapid PPK-corrected DSMs showed reduced errors with longer processing times. The lidar DSM exhibited a much lower RMSE of 0.04 m |
| | | | | | | | compared to the drone DSMs. |
| 5 | [269] | Wildlife Monitoring | Assessing the likelihood of encountering marine megafauna through real-time video analysis | Surf Beach, Kiama, South coast, New South Wales, Australia | December 2017 to January 2018 | The study observed common marine megafauna, including grey nurse sharks, Australian fur seals, and smooth Stingray, in a nearshore area of approximately 18,500 $m^2$.<br><br>This work provides insights into the habitat usage and behavior of these marine species. | The encounter rate varied by species, with stingrays being more frequently observed than sharks and seals. Seals showed a higher likelihood of presence when baitfish were present, while the occurrence of sharks and stingrays did not seem significantly influenced by baitfish presence in the bay. |
| 6 | [95] | Precision Agriculture | Analyzing field trials of fertilizer, damage from armyworm and lodging, tile drainage field mapping. | West Nipissing, northeastern Ontario, Canada. | 2013 | The study compared different fertilizer treatments in a soybean field using aerial imagery. The field is categorized in to three sections.<br><br>The area which is treated with organic fertilized, mix of organic and chemical fertilized and chemical fertilized is termed as section A, B and C respectively. | Section A showed lower vegetation vigor compared to section C.<br><br>In the initial phase of growth, section B did not significantly differ from section while the difference increases between the two in the final yield. |
| 7 | [238] | Search and Rescue | The SARUAV system was used for | Cergowa, Beskid | 28 and 29 | Search and Rescue Operation carried out | Located the missing person within 4 hours |
| | | Operations | Search and Rescue Operation. | Niski SE, Poland | June 2021 | by Bieszczady Mountain Rescue Service to find a missing 65-year-old man using UAV images and computer vision techniques. | and 31 minutes of its deployment. The person was identified during third flight, covering 10 hectares, and SARUAV processed the data in just 1 minute and 50 seconds. Validation by an analyst took an additional 2 minutes and 15 seconds. In comparison, manual rescue teams typically require 2 to 2.5 hours for searching 10 hectares area. |

## 5. Challenges and Mitigations

The challenges in aerial data analysis stem from the unique advantages offered by aerial data. Aerial platforms like UAVs or drones, while capable of capturing data over expansive geographic areas, introduce complexities. Objects may appear small and scattered or, when too close, merge into a single entity [276]. The angle of data capture further influences the analysis, altering object perspectives from the top compared to the side. Relative positions are affected by the camera's angle, impacting the interpretation of features of interest [47]. Shadows become pronounced due to variations in light and camera angles, especially during continuous UAV or drone motion, and when both the capturing device and objects are in motion. These diverse factors contribute to the array of challenges encountered in aerial data analysis, ranging from scale variations to occlusions caused by various structures [42], as illustrated in Figure 25. In this section, we will provide the details of various challenges related to aerial data analysis and the related studies, which work on their mitigation. An overview of the work contributing to navigating through these challenges in the analysis of aerial data, specifically object detection is as shown in Table 8.

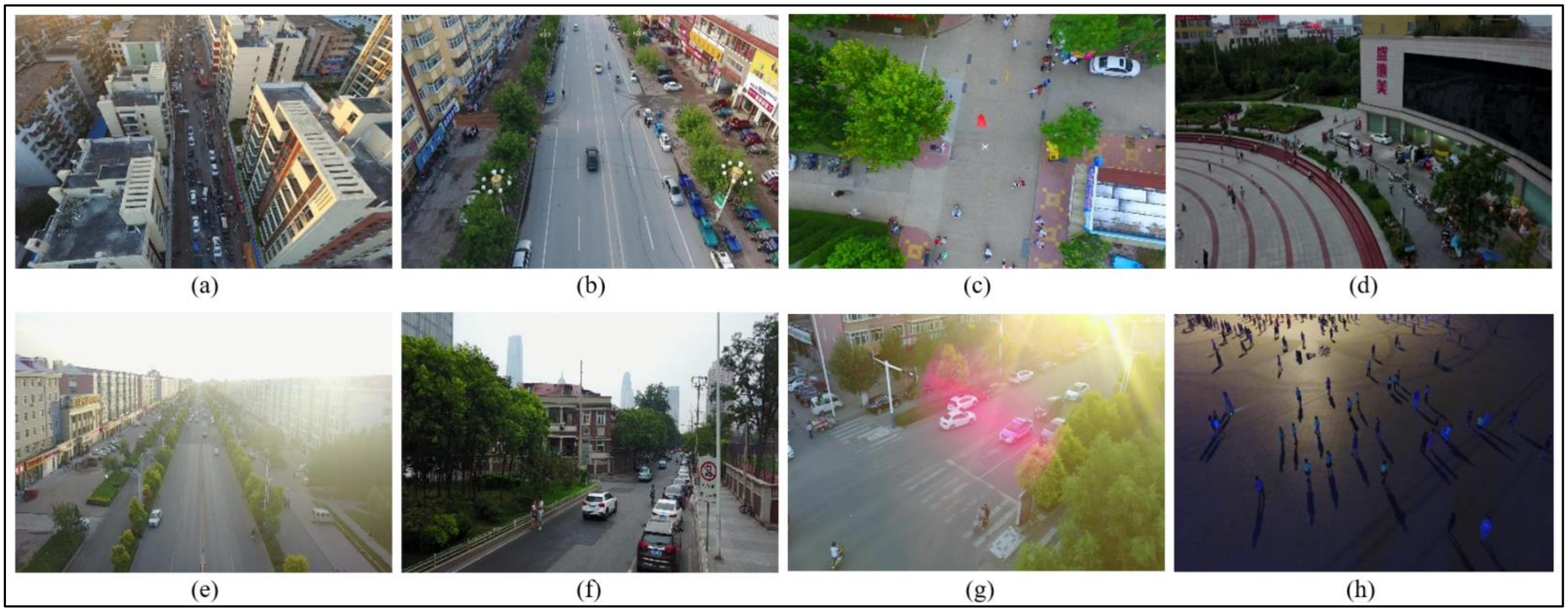

Figure 21: Images from VisDrone 2019 [59] dataset depicting challenges in aerial data analysis: (a) & (b) Cluttered backgrounds; (c) & (d) Occluded objects; (e) & (f) Unpredictable orientations and geometric distortions; scale variation; (g) Uneven object intensity; (h) Night timing

#### *5.1.1. Cluttered Backgrounds*

The complexity of backgrounds in aerial data, characterized by irregular object distribution, poses significant challenges for object detection and recognition. The diverse and often cluttered environments make it difficult for computer vision systems to distinguish between target objects and their surroundings [42]. This complexity is heightened by factors such as varying scales, orientations, and occlusions, making it challenging to develop algorithms that can accurately identify and classify objects of interest as the features for same object may vary depending on these factors. As shown in Figure 25 (a) & (b), the objects exhibit varying density, with less density in Figure 25 (b) and higher density in Figure 25 (a). This irregular distribution of objects poses a challenge for analyzing aerial data. This problem is addressed in [277] by means of a supervised MDA-Net consisting of pixel attention network and channel attention network to suppress the noise and highlight foreground. In [278], the authors proposed a data augmentation based end-to-end GLSAN with a SARSA, to dynamically crop local regions from the crowded area.

#### *5.1.2. Occlusion*

Aerial data is often obscured or occluded due to atmospheric conditions like cloud cover, shadows, vegetation, thus hindering the visibility of certain objects and hiding crucial details in imagery [42], [47]. Cloud cover, for instance, introduces a significant barrier, limiting the penetration of light and distorting the captured information. Shadows cast by terrain irregularities or structures further complicate the interpretation of aerial data, creating areas of darkness that conceal important features as shown in Figure 25 (d). Vegetation, with its varying density and height, can also obscure the view, making it challenging to identify or analyze objects beneath the canopy as shown in Figure 25 (c). The authors in [279], [280] addressed the problem of detecting occluded objects by means of random covering data augmentation technique and mix-up and mosaic data augmentation techniques respectively.

#### *5.1.3. Unpredictable Orientations*

Aerial data is generally collected by devices in continuous motion, from varying distances and angles, thus introduces motion blur, unpredictable orientations, extreme views and geometric distortions in the collected data. The dynamic nature of data capture, coupled with varying distances and angles, contributes to motion blur, affecting the clarity and details of the imagery [42]. Unpredictable orientations of the capturing device introduce complexities in interpreting the spatial relationships between objects as shown in Figure 25 (e) & (f). In Figure 25 (f), objects appear closer than they actually are due to variations in capturing angle and distance. Additionally, in Figure 25 (e), where the capture distance is relatively larger than in Figure 25 (f), objects of the same class exhibit varying shapes and orientations. The extreme viewing angles may result in distorted perspectives, making it challenging to accurately assess features [45]. Geometric distortions further compound the issues, impacting the precision of measurements and thus, the process of aerial data analysis. This issue is addressed in [281] by means of using UAV-specific metadata for training an adversarial learning framework. In [277], an angle sensitive network is devised by introducing an angle related parameter (IoU smooth L1 loss) for estimation. In [282], a domain sensitive model considering the altitude and view angle is proposed to overcome the problem of unpredictable orientations.

#### *5.1.4. Varying Scale and Resolution*

Aerial data collection introduces scale and resolution challenges as the imaging devices operate from varying distances above the ground. This variation in altitude leads to differences in the scale and resolution of objects within the data [42]. This difference in scale and resolution is illustrated in Figure 25 (e) & (f), where objects captured from higher altitudes, as in Figure 25 (e), appear considerably smaller in the imagery. This discrepancy affects the perception of fine details and adds complexity to the process of aerial data analysis. The issue is addressed in [277] by means of feature fusion and finer anchor sampling. In [278], the authors proposed a data augmentation based end-to-end GLSAN, including a GLDN for different scales of object detection in drone-view images. A preprocessing strategy named Adaptive Resizer was proposed by [283] to rescale images eliminating scale variance. A RFEB to extend the size of the receptive field for high-level semantic features is proposed in [284] to improve spatial resolution of multi-scale data. An approach named Scale Match to align object scale is proposed by [285]. In [280], the authors proposed a modified YOLOv5s model that uses k-means++ clustering and EIoU_Soft_NMS integrated with an attention mechanism to detect small objects. In [279], a deformable Faster-CNN using a deformable convolutional layer, with skip connections to extract feature map of high resolution is proposed.

### *5.1.5. Uneven Intensity*

Uneven illumination in aerial data, caused due to obstructions, terrain variations, and the changing time of day, introduces significant challenges in the extraction of meaningful features. Shadows cast by structures or elevated terrain can create darkened regions, hiding crucial details in the imagery [42]. The dynamic changes in sunlight throughout the day intensify variations in illumination, as depicted in Figure 25 (g). These factors add complexity to the process of aerial data analysis. The issue is addressed in [281] using UAV-specific metadata for training an adversarial learning framework. In [284], the authors proposed a RFEB to extend the size of the receptive field for high-level semantic features from data with varying light conditions.

### *5.1.6. Adverse Weather Conditions*

Adverse weather conditions, including night time, fog, haze, rain, and storms, significantly impact the visibility and subsequently degrade the quality of aerial data [45]. Night time conditions result in reduced natural illumination thus, requiring artificial light sources, introducing shadows and limitations in capturing accurate details as shown in Figure 25 (h). Fog and haze introduce a diffusion of light, reducing contrast and making it challenging to discern objects and details within the imagery [42]. Rain can distort the captured information, leading to blurred and obscured visuals. Storms, with their dynamic atmospheric effects, further complicate data quality by introducing turbulence and unpredictable variations in lighting. This issue can be addressed by training a model using features that doesn’t vary depending on the weather conditions or time of day as in [281] or a domain sensitive model as in [282].

### *5.1.7. Limited Training Data*

The process of collecting and annotating aerial data is resource-intensive, leading to a scarcity of labeled training data [100]. Aerial data acquisition involves deploying equipment such as drones which entails substantial costs and logistical complexities. Moreover, the manual annotation of datasets, a critical step for training machine learning models, is a time-consuming process requiring expertise and meticulous attention [286]. The scarcity of labelled training data poses a challenge for developing robust algorithms, especially in computer vision applications like object detection and image classification. This issue can be addressed by employing transfer learning as in [281] where it leads to an improvement in AP by 4.23%.

Table 8: Contribution of different works to overcome the challenges in the analysis of aerial data

| S. NO. | Reference | Challenge Addressed | Proposed Solutions | Achievements | Datasets used in the experiments | Remarks |
|---|---|---|---|---|---|---|
| 1. | [281] | Unpredictable Orientations, Adverse weather conditions, Uneven intensity, Limited training data | New Architectures (NDFT-DE-FPN, NDFT-Faster-CNN) proposed | The AP score of NDFT-DE-FPN boost by 4.36% over DE-FPN (VisDrone2018).<br>The AP score of NDFT-Faster-CNN boost by 2.27% over Faster-RCNN with ResNet-101 backbone (UAVDT).<br>The transfer learning of NDFT from UAVDT to VisDrone2018 leads to the increment of 4.23% in terms of AP score. | UAVDT, VisDrone2018 | Utilized the UAV-specific metadata (e.g., flying altitude, camera view or weather condition) along with training data to learn task specific and domain (altitude/ scale/ angle/ weather) invariant features. |
| 2. | [277] | Unpredictable Orientations, Varying scale and Resolution, Cluttered Backgrounds | New Architectures (SCRDet) proposed | SCRDet gives the best performance in terms of mAP score of 75.35% (DOTA)<br>SCRDet gives the best performance in terms of mAP score of 91.75% (NWPU VHR-10) | DOTA, NWPU VHR-10 | For small objects: The SF-Net incorporating feature fusion and finer anchor sampling is devised.<br>For cluttered objects: A supervised MDA-Net consisting of pixel attention network and channel attention network to suppress the noise and highlight foreground.<br>For rotated objects: an angle sensitive network is devised by introducing an angle related parameter (IoU smooth L1 loss) for estimation. |
| 3. | [282] | Unpredictable Orientations, Time of Day | The domain sensitive model was proposed, Created a dataset for aerial object detection namely POG | The AP70 score of the proposed model is boost by 21% over DE-FPN (UAVDT).<br>The $mAP_{50}$ score of the proposed model is boost by 1.3% over EfficientDet-D0 (VisDrone).<br>The $AP_{50}$ score of the proposed model is boost by 5.9% over EfficientDet-D0 (POG). | UAVDT, VisDrone, POG | Proposed a domain sensitive model (with domains namely altitude, view angle and time of day) for object detection.<br>Created a dataset POG containing 2.9k images (3840 × 2160), at angles varying from 0 to 90 and altitudes varying from 4m to 103m; along with the metadata. |
| 4. | [278] | Varying Scale and Resolution (Small), Cluttered Background | New Architectures (GLSAN) proposed | GLSAN significantly improved the accuracy of different models at a range of 9.4%-14.9%, and the model based on FPN achieved the largest improvement and the best performance (VisDrone2019-DET).<br>The $AP_{50}$ score of the proposed network with ResNet50 is improved by 30.5% over CascadeRCNN (UAVDT)<br>The $AP_{50}$ of the proposed network has 8.8% and 6.0% improvements over FPN and SNIPER (DroneBolts) | UAVDT, VisDrone2019-DET, DroneBolts | Proposed a data augmentation based end-to-end GLSAN model that includes:<br>• GLDN for different scales of object detection in drone-view images<br>• SARSA is used to dynamically crop local regions from the crowded (higher-resolution) area<br>• LSRN is used to obtain two components pertinent to TDA. |
| 5. | [283] | Varying Scale and Resolution | New pre-processing strategy (Adaptive Resizer) proposed | The $AP_{50}$ score of the proposed solution with ResNet50 is improved by 5% over DE-FPN (VisDrone)<br>The AP70 score of the proposed solution with Resnet-101-FPN backbone is improved by 13% over Faster R-CNN (UAVDT) | UAVDT, VisDrone | Proposed a pre-processing strategy Adaptive Resizer to rescale images to eliminate the scale variance problem. |
| 6. | [284] | Varying Scale and Resolution (data sets of small distant targets), Uneven Intensity (light variation), Occlusion (dense targets with occlusion) | New Architectures (SAMFR-Cascade RCNN) proposed | The proposed model (SAMFR-Cascade RCNN) achieves 4.09% improvement over the baseline method of Cascade RCNN that obtains a mAP of 16.09% with [0.5:0.95] IoU. | VisDrone2019 | To improve spatial resolution of the multi-scale data, a RFEB is proposed that extends the size of the receptive field for high-level semantic features, which are then passed through an SRM. |
| 7. | [285] | Varying Scale and Resolution (tiny persons less than 20 pixels), | New pre-training (Scale Match) approach | SM COCO for pre-training improves $AP_{50}$ score by 3.22% while MSM COCO for pre-training improves $AP_{50}$ score by 3.74%. | New benchmark dataset TinyPerson | Proposed an approach named Scale Match to align the scales of objects for pre-training and detector training datasets. |
| | | Cluttered Backgrounds | proposed, Created a dataset for aerial object detection namely TinyPerson | | | |
| 8. | [280] | Varying Scale and Resolution, Occlusion | New Architectures (YOLOv5s_2E) proposed | The proposed YOLOv5s model outperforms the same series of m and l models by 8.2%, 5.7% with mAP@0.5 score and 7.6%, 6.0% in mAP@0.5:0.95 score respectively.<br>Compared with YOLOv7-tiny, the mAP@0.5 score and mAP@0.5:0.95 score of the proposed model is increased by 10.2% and 10.1%, respectively. | VisDrone | Proposed YOLOv5s_2E method for detection of occluded and small objects along with Mix-up and Mosaic techniques of data augmentation. |
| 9. | [279] | Occlusion, Unpredicted Orientations (geometric variations), Varying Scale and Resolution (small objects) | Faster-CNN with Random Covering data augmentation technique was proposed | The proposed model achieves the mAP score of 84.4% (NWPU VHR-10), and 88.3% (SORSI). The AP value significantly increased for small objects in both the datasets. The AP value increased from 0.758 to 0.901 when Random Covering is used to address Occlusion. | NWPU VHR-10, SORSI, HRRS | Random Covering data augmentation technique is used with deformable Faster-CNN using a deformable convolutional layer, with skip connections to extract feature map of high resolution. |

## 6. Future Research Directions

We have discussed the different challenges associated with the aerial dataset in the previous section. Although the researchers have made attempts to address those challenges still there is a lot of scope for improvements. Along with that there are different future research directions related to this domain. Following the standard approach, we have divided the future research prospects into two categories namely data driven and model driven as shown in Figure 26. In data-driven approach, the emphasis is given to address the data related problems like limited data availability, high costs of data labelling etc., while in model-driven approach, the emphasis is given to address model related problems like adversarial attacks, context-ignorance, high computation costs etc.

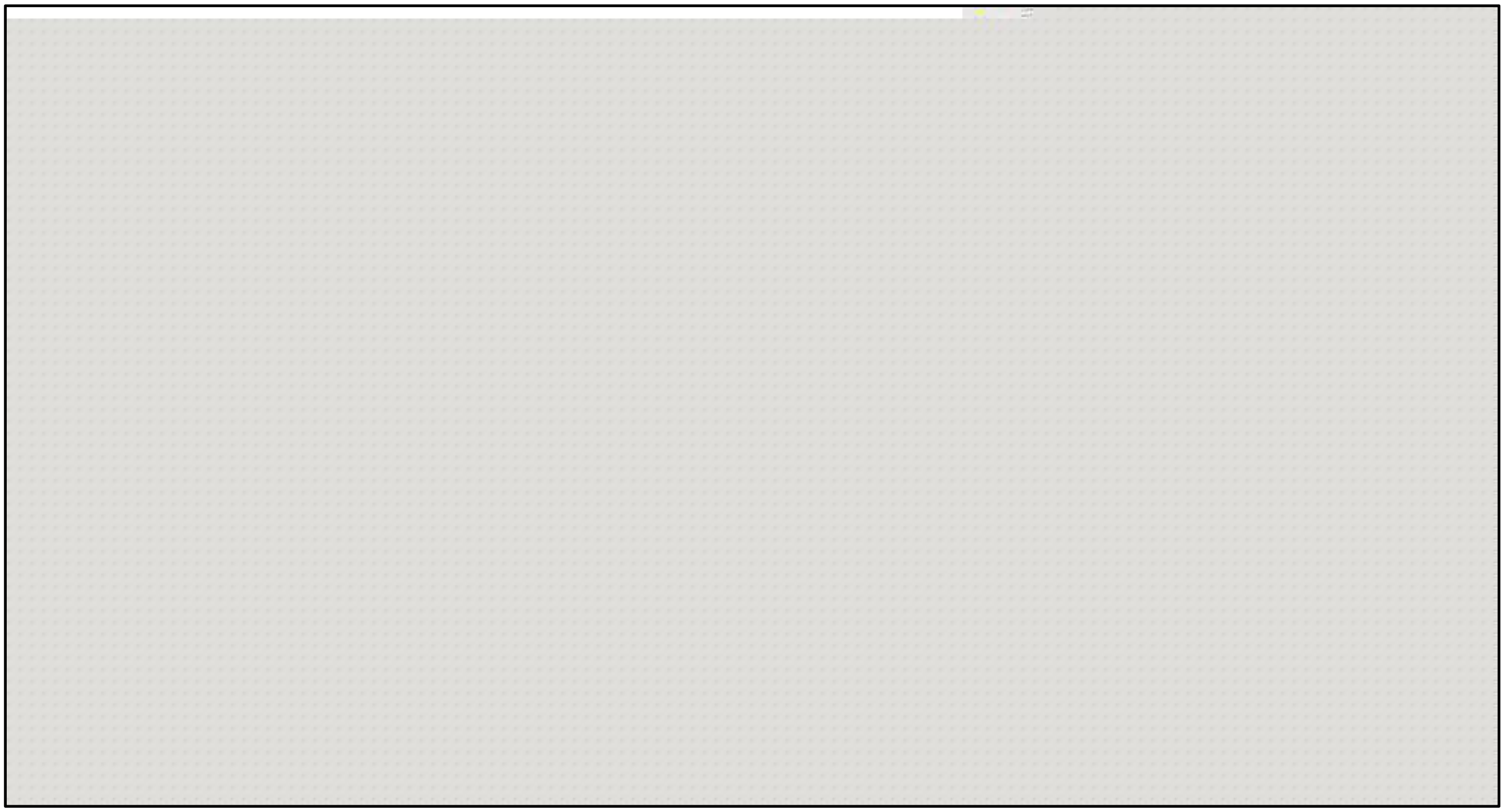

Figure 22: Top-level categorisation of future research directions in aerial data analysis

### 6.1. Data Driven

Aerial computer vision encounters various data-related challenges that impact its efficiency and performance. One significant issue is the limited availability of diverse aerial datasets for training vision models [100]. The scarcity of well-annotated aerial data, especially in diverse environmental conditions, hinders the robust training of vision models [161]. Additionally, the dependency on weather conditions poses a challenge, as adverse weather can affect data collection, resulting in gaps in availability. Furthermore, the manual annotation of aerial data is a time-consuming and labor-intensive task, leading to potential inconsistencies and errors in labeled datasets [286]. To address these challenges, efforts are needed to create comprehensive benchmark datasets that are diverse, rich, and scalable, comprising of multiple modalities to ensure availability in diverse weather conditions and labelled autonomously, facilitating the development and evaluation of intelligent algorithms for aerial data analysis. These approaches are detailed in the following sub-sections.

#### *6.1.1. Availability of large training datasets*

The impact of insufficient data availability is substantial on the performance of computer vision models, as it can lead to biased models with limited capability to handle real-world variations [100]. Models trained on sparse datasets may struggle to recognize objects or features in unfamiliar environments, resulting in decreased accuracy and reliability. In certain tasks like object detection, segmentation, etc., the model's performance may suffer when faced with diverse landscapes or unique scenarios that are not adequately represented in the training data. Also, while data is collected, due to various financial and legal constraints, it is not feasible to collect data for every possible scenario [161]. Use of synthetic data or data augmentation, can contribute to overcoming these data-related obstacles in aerial computer vision. Also, the process of data collection can be streamlined by establishing standardized protocols to create efficient benchmark datasets. In [161], [287], the authors presented synthetic aerial datasets with images synthesized in a virtual environment. The authors in [288] presented general guidelines for creating benchmark datasets efficiently, emphasizing DiRS properties including diversity, richness, and scalability, desired for effective interpretation of aerial data. As an example, a large-scale benchmark dataset named Million Aerial Image Dataset with a million instances for RS image scene classification is introduced in [288]. Additionally, leveraging advanced techniques such as synthetic data generation can also help augment limited datasets, providing models with a broader understanding of various environmental conditions [161]. Although some researchers have worked in the direction of availability of large datasets, yet some more exploration in this direction is needed.

#### *6.1.2. Heterogenous aerial data/Multi-modal data fusion*

Multi-modal data fusion refers to the integration of information captured using different sensors in different modalities, resulting in an enhanced overall understanding of a given area [289]. The aerial data can be captured by aerial platforms mounted with different sensors like optical, thermal, NIR, hyper-spectral or multi-spectral cameras, or satellites, LiDAR etc. Each of these sensors can capture distinct aspects such as optical cameras can provide better spatial resolution than hyper-spectral cameras whereas multi-spectral and hyper-spectral cameras can provide better spectral resolution [39], [40], [41]. Thermal, NIR cameras can be used for data collection during night time or in adverse weather conditions. Integrating these data can represent the observed environment more comprehensively and accurately. Also, multi-modal fusion ensures adaptability across different domains by providing insights from various perspectives. In [290], the authors employed multimodal data fusion, integrating RGB, multispectral, and thermal images, to enhance the accuracy of soybean yield predictions. The authors in [291] proposed MDL-RS framework with a fusion module namely cross fusion that can efficiently transfer information across modalities. A S2FL model is proposed in [289] that introduces a mechanism to separate multimodal data into modality-shared and modality-specific components, enhancing the effective blending of information from diverse sources. To evaluate this approach, three benchmark datasets—Houston2013 (hyperspectral and multispectral data), Berlin (hyperspectral and SAR data), and Augsburg (hyperspectral, SAR, and DSM data) are released for landcover classification. The proposed methods have improved the efficiency and performance of models using multi-modal data, yet there is great scope of improvement in this area.

#### *6.1.3. Autonomous labelling of data*

Annotating visual data, especially in the context of aerial imagery, poses challenges due to its time-consuming nature and resource-intensive requirements. With advantages like extensive coverage and the ability to capture challenging terrains, aerial data introduces additional complexities like vast size of data, complex structures, along with challenges such as varying scales, resolutions, and occlusions [286]. Specialized considerations are needed to ensure annotations are applicable across diverse environments. Solutions like autonomous labeling and crowdsourcing annotations by volunteers are explored, yet the latter may suffer from issues of inconsistent annotations. In [292], the authors, introduced a crowdsourcing partial annotation approach in which instead of annotating a specific number of images, volunteers are tasked with annotating a set annotation budget i.e. total number of labels, resulting in the creation of partially annotated datasets with salient labels and demonstrated that, under the same annotation budget, the classification model trained on partially annotated images exhibited superior performance compared to the model trained on fully labeled images. The authors in [286] suggested multi-label learning in image analysis with limited annotation, introducing innovative approaches such as partial annotations in multi-label image classification to reduce errors and enhance efficiency. An autonomous data annotation technique to label data on-the-fly using reverse engineering is proposed in [293]. The above studies have tried to work in the direction of annotating large aerial datasets, yet some more exploration in this direction is needed.

### 6.2. Model Driven

Aerial computer vision faces several challenges related to its models, influencing both efficiency and performance. These challenges may stem from factors such as the saturation of performance in current models, training solely on datasets from controlled environments, susceptibility to adversarial attacks, insufficient contextual information in datasets, demanding computational requirements for models, and the associated high costs and energy consumption in implementing these algorithms [294], [67]. Overcoming these model-related limitations in aerial data analysis is pivotal for advancing the effectiveness of these systems. Diversifying models by exploring architectures successful in other domains can bring new insights and potentially enhance performance. Designing models robust to adverse conditions, such as rain or fog, involves training on datasets that replicate real-world challenges. Adversarial defense mechanisms, like adversarial training, can fortify models against intentional attacks [295]. Integrating rich contextual information into training datasets aids models in understanding complex surroundings [296]. Prioritizing lightweight and real-time models ensures efficient processing, especially in resource-constrained environments, while energy-efficient solutions align with sustainability goals [297], [298]. Additionally, employing transfer learning from pre-trained models on diverse datasets contributes to overcoming limitations associated with training solely in controlled environments [299]. Embracing these strategies collectively enhances the resilience and versatility of aerial computer vision applications in diverse and dynamic scenarios. These approaches are detailed in the following sub-sections.

#### *6.2.1. Applying successful models from other fields*

Leveraging successful models from other domains can provide valuable solutions to similar challenges in the field of aerial computer vision. Various issues like adversarial attacks, adverse weather conditions etc. are addressed well in other domains. These approaches can be applied in the analysis of aerial data. For instance, the transformer architecture, that was originally designed for natural language processing tasks, is adapted in computer vision for classification tasks as in ViT [147] for better performance with lesser resources. In [148], a transformer-based model for semantic segmentation is proposed which is relatively lightweight. In [300] a fast Anomaly Generative Adversarial Network architecture for anomaly detection in aerial images is used which was used on medical images originally. In [301], deep reinforcement learning, initially applied in game theory, is employed in enhancing aerial data quality by dehazing. This inter-domain application of models and techniques showcases the need and potential for innovative solutions in aerial data analysis.

#### *6.2.2. Robustness to adverse conditions*

The datasets used for training models are typically obtained in controlled environments, mirroring ideal conditions. However, real-world application scenarios often differ due to factors such as adverse weather conditions, including rain, fog, storms, or haze; night-time etc. [281]. Models trained on datasets might exhibit good performance during testing, but their efficacy may falter when confronted with the dynamic and unpredictable conditions present in actual real-time scenarios. This underscores the need for models that can generalize well beyond the controlled settings of the training data, ensuring robust performance in the face of the unpredictable and varied conditions encountered in real-world applications. For instance, in [302], the authors proposed a domain adaptive unsupervised approach for object tracking that can effectively track objects during night time. A dehazing technique based on deep reinforcement learning is proposed in [301] to reduce the effects of hazing of images on model’s performance. In [281], the researchers suggested leveraging freely available UAV meta-data, like weather conditions, altitude, and angles, in conjunction with associated images through an adversarial training framework for learning domain-robust features. These proposed solutions exemplify the potential of models to address diverse and dynamic conditions encountered in real-world applications. Although some researchers have worked in this direction, still more attention is needed in this area.

### *6.2.3. Robustness to adversarial attacks*

An adversarial attack refers to manipulation of input data with the intent of causing a ML or DL model to make incorrect predictions or classifications. The input is crafted in such a way that it seems innocuous to a human observer, but can mislead the model [294]. As adversarial attacks are not perceptible to humans, it is very difficult to identify if an attack has happened. The impact of these attacks on models can be significant, leading to misclassifications, decreased accuracy, and compromised model reliability. Addressing adversarial attacks is crucial to enhance the security and reliability of the models. In [303], the authors presented both proactive (building resilient models) and reactive (detecting adversarial attacks) defense approaches based on ensemble models using CNN and transformer architectures. In [295], the authors investigated the effectiveness of adversarial training and defensive distillation methods for enhancing robustness to adversarial attacks against DL-based UAVs. A method using attribution maps created by model visualization techniques is introduced in [294] for detection of adversarial examples. These proposed solutions illustrate the potential of models to address adversarial challenges encountered in real-world applications, thus, emphasizing the need of future work in this direction.

### *6.2.4. Transfer learning for aerial data*

It is the process of utilizing a pre-trained model designed for one task as a foundation or starting point for creating a new model tailored to a different task. This technique becomes particularly beneficial when there is limited data available for a specific task [304]. The premise of transfer learning lies in the idea that during the training of a model, it learns a multitude of low-level features for detecting edges, color, corners, variations of intensities, textures etc. These low-level features are generally applicable across various datasets and tasks within the realm of computer vision. Due to the limited availability of aerial datasets, the models established using ground level imagery, can be effectively applied to enhance performance in aerial data analysis. A U-Net model pretrained on a large dataset is used to employ transfer learning for change detection in aerial images in [299]. In [305] also, transfer learning is employed for change detection in urban scenarios from aerial imagery. In [304], transfer learning is leveraged using pretrained VGG16 with ImageNet dataset for classification of crops using UAV images. These studies emphasize the importance of transfer learning in aerial computer vision, thus, advocating the need of future work in this direction.

### *6.2.5. Context-awareness*

Context-awareness is a crucial aspect of computer vision, mirroring the way human vision incorporates context for effective decision-making. In these systems, understanding the surrounding environment and considering the associations and relationships between objects within an image is essential for accurate interpretation [67].

Context provides the necessary background information, supporting the system to infer meaning and make more informed decisions. Incorporating context-awareness enables these systems to recognize objects in their environmental context, improving overall performance and reliability of models. A context-aware building change detection model considering local as well as global features and using an aggregation module with a Siamese structure is proposed in [306] which eliminates detection of pseudo-changes. In [296], a relational context-aware FCN using a spatial relation module to consider global relationships between any two spatial positions, and a channel relation module to consider global relationships between any two feature maps is proposed. In [67], the authors used CRAM and CCFM to analyze the relationships between classes and patch regions and between classes and local or global pixels to increase model's perception capability. These studies emphasize the importance of context-awareness in aerial computer vision, thus, advocating the need of future work in this direction.

#### *6.2.6. Lightweight models for real-time processing of aerial data directly on edge devices*

DL-based models, while delivering high performance, often pose challenges due to their complexity and high computational requirements. This complexity makes their deployment on edge devices or those with limited computational resources impractical [307]. In real-time applications such as surveillance, monitoring, and search and rescue, on-board UAVs require models capable of swift decision-making. Lightweight models, designed with reduced computational complexity and memory requirements, are well-suited for efficient on-edge processing of aerial data. Such models facilitate rapid analysis directly at the source, reducing the need for data transfer to centralized servers. The authors in [297] have proposed a lightweight detection network for detection tasks on remote sensing images using network weight reduction and optimization methods. A lightweight CNN architecture based on depth wise separable convolutions, namely LW-AerialSegNet is proposed in [308] for segmentation of aerial images. The authors in [309] proposed a light weight detection model based on self-attention mechanism of ViT using YOLOv5. The authors in [307] proposed a light weight detection model based on YOLOv4 using sparsity training, channel and layer pruning for reducing model size and knowledge distillation techniques to maintain the performance. As discussed above, many lightweight models for edge devices have been proposed, yet there is great scope of future work in this area.

#### *6.2.7. Investigate energy-efficient, cost effectiveness and complexity of the algorithms*

State-of-the-art models in DL typically require substantial computational resources, resulting in impractical power consumption, especially in real-world applications. This results in a trade-off between achieving high model performance, and minimizing computational resources and power consumption during real-world applications [297]. Striking a balance in this trade-off is crucial for making DL models viable and practical in various scenarios, especially when faced with constraints such as limited computational resources or the need for real-time processing. In [297], the authors proposed a FPGA based energy efficient CNN accelerator using a reconfigurable and efficient convolutional processing engine. Additionally, a flexible FPGA architecture is introduced in [298] to reduce power consumption, demonstrating its efficiency with AlexNet, for which the power consumption reduced by approximately 40%. Another study in [310] conducted a comparative analysis on the energy efficiency of CPU, GPU, and FPGA implementations for embedded vision tasks, revealing that, for simple tasks, GPUs perform better, while FPGAs outperform others in more complex tasks. These efforts underscore the importance of developing energy-efficient hardware implementations to enhance the practicality of DL models in real-world applications. These studies emphasize the need of energy-efficient, cost effective and simple algorithms for aerial computer vision, thus, advocating the need of future work in this direction.

## 7. Conclusion

Due to various advantages provided by aerial data in comparison to land data, aerial data analysis provides more holistic view of the situation. Due to the presence of various challenges associated with aerial data analysis, the research in this area is very daunting. This paper presented a comprehensive review of various computer vision tasks for aerial data analysis. Encompassing aerial datasets, adopted architectural, and evaluation metrics for all the computer vision tasks in aerial data analysis, the paper delved into applications across various domains, offering insights through case studies. A thorough discussion on various hyper parameters used in diverse architectures and tasks, along with an in-depth exploration of libraries, their categorization, and their relevance to different domains of expertise was provided. Thorough examination of challenges in aerial data analysis accompanied by their practical solutions was discussed. The paper identified unresolved issues, laying the groundwork for future research directions in the field of aerial data analysis.